\documentclass[twoside,11pt]{article}

\usepackage{blindtext}

\usepackage[abbrvbib, preprint]{jmlr2e}

\usepackage{jmlr2e}
\usepackage{xspace}
\usepackage{booktabs}
\usepackage{multicol}
\usepackage{multirow}
\usepackage{xcolor}
\usepackage{caption}
\usepackage{subcaption}

\newcommand{\ie}{\emph{i.e.,}\xspace}

\newcommand{\ignore}[1]{}

\usepackage{tcolorbox}
\newtcolorbox{promptbox}[2][Prompt]{
colback=black!5!white,
arc=5pt, 
boxrule=0.5pt,
fonttitle=\bfseries,
title=#1, 
before upper={\scriptsize}, fontupper=\fontfamily{ptm}\selectfont,
colframe=#2, 
}

\ShortHeadings{Causal-Copilot: An Autonomous Causal Analysis Agent}{Causal-Copilot: An Autonomous Causal Analysis Agent}
\firstpageno{1}

\begin{document}

\title{Causal-Copilot: An Autonomous Causal Analysis Agent}
\author{\name Xinyue Wang$^*$ \email xiw159@ucsd.edu \\
       \name Kun Zhou$^*$ \email franciskunzhou@gmail.com \\
       \name Wenyi Wu$^*$ \email wew058@ucsd.edu \\
       \name Har Simrat Singh \email h6singh@ucsd.edu \\
       \name Fang Nan \email fnan@ucsd.edu \\
       \name Songyao Jin \email soj007@ucsd.edu \\
       \name Aryan Philip \email arphilip@ucsd.edu \\
       \name Saloni Patnaik \email sapatnaik@ucsd.edu \\
       \name Hou Zhu \email hoz020@ucsd.edu \\
       \name Shivam Singh \email shs046@ucsd.edu \\
       \name Parjanya Prashant \email pprashant@ucsd.edu \\
       \name Qian Shen \email q1shen@ucsd.edu \\
       \name Biwei Huang \email bih007@ucsd.edu \\
       \addr University of California San Diego, CA 92093, USA}

\editor{My editor}

\maketitle

\def\thefootnote{*}\footnotetext{Equal contribution.}\def\thefootnote{\arabic{footnote}}

\begin{abstract}

Causal analysis plays a foundational role in scientific discovery and reliable decision-making, yet it remains largely inaccessible to domain experts due to its conceptual and algorithmic complexity. This disconnect between causal methodology and practical usability presents a dual challenge: domain experts are unable to leverage recent advances in causal learning, while causal researchers lack broad, real-world deployment to test and refine their methods. To address this, we introduce Causal-Copilot, an autonomous agent that operationalizes expert-level causal analysis within a large language model framework. Causal-Copilot automates the full pipeline of causal analysis for both tabular and time-series data—including causal discovery, causal inference, algorithm selection, hyperparameter optimization, result interpretation, and generation of actionable insights. It supports interactive refinement through natural language, lowering the barrier for non-specialists while preserving methodological rigor. By integrating over 20 state-of-the-art causal analysis techniques, our system fosters a virtuous cycle - expanding access to advanced causal methods for domain experts while generating rich, real-world applications that inform and advance causal theory. Empirical evaluations demonstrate that Causal-Copilot achieves superior performance compared to existing baselines, offering a reliable, scalable, and extensible solution that bridges the gap between theoretical sophistication and real-world applicability in causal analysis.
\end{abstract}


\begin{center}
\textbf{Code \& Data: \url{https://github.com/Lancelot39/Causal-Copilot}}

\textbf{Video Demo: \url{https://www.youtube.com/watch?v=U9-b0ZqqM24}}

\textbf{Online Demo: \url{https://causalcopilot.com/}}
\end{center}

\section{Introduction}


Causal learning and reasoning lie at the foundation of scientific understanding and principled decision-making. They enable researchers to move beyond associational patterns to uncover the underlying mechanisms that govern observed phenomena~\citep{spirtes2000causation}. By revealing how system variables interact and influence outcomes, causal analysis provides a foundation for actionable insights across critical domains, including healthcare~\citep{prosperi2020causal}, economics~\citep{imbens2020potential}, and engineering~\citep{kleinberg2011review}. As data-driven systems increasingly underpin high-stakes applications, integrating causal reasoning into analytical pipelines has become essential for building models that are not only accurate but also robust, interpretable, and trustworthy.

Over the past decades, the field has seen a rapid development of methods for causal discovery, treatment effect estimation, and counterfactual inference~\citep{nogueira2022methods}. These methods are grounded in diverse theoretical frameworks and designed to accommodate real-world challenges such as latent confounding, selection bias, nonstationarity, or feedback loops. However, despite their growing sophistication, these methods remain largely underutilized in practice due to their steep learning curves, algorithmic diversity, and the expert knowledge required to navigate them effectively~\citep{machlanski2024understanding}. This creates a paradoxical situation where increasingly powerful causal tools are developed but rarely deployed at scale—domain experts cannot access the methodological advances they need, while causal researchers lack the broad real-world testing grounds necessary to refine their approaches, perpetuating the disconnect between theoretical sophistication and practical applicability.

A complete causal analysis pipeline typically involves multiple interconnected steps: understanding the task, selecting appropriate algorithms based on data characteristics and assumptions, configuring hyperparameters, executing causal discovery or inference procedures, and interpreting results~\citep{machlanski2024understanding}. Each step demands familiarity with the underlying assumptions, limitations of various approaches, and domain-specific constraints. In practice, such expertise is rarely available across application domains, and organizations are often forced to rely on oversimplified heuristics or manual trial-and-error processes. This limits the scalability and reliability of causal analysis in applied settings and highlights the need for more intelligent and accessible tools.

To address these challenges, we propose \textbf{Causal-Copilot}, a modular system powered by large language models (LLMs) that automates the entire causal analysis workflow. The system integrates components for task understanding, algorithm selection, data preprocessing, inference execution, and result interpretation, all coordinated through a centralized reasoning layer~\citep{Brown2020LanguageMA,zhao2023survey}. Given a tabular or time-series dataset and a natural language query, Causal-Copilot identifies the analytical intent, configures appropriate methods and hyperparameters, and executes the analysis. It then generates a structured PDF report containing visualizations of descriptive statistics, inferred causal structures, treatment effects, and accompanying natural language explanations. The interface supports optional user interaction, allowing users to monitor and refine the process without programming expertise.

Causal-Copilot currently integrates over 20 state-of-the-art causal learning methods, including constraint-based, score-based, and hybrid approaches for structure learning, as well as inference algorithms based on double machine learning and counterfactual estimation. This design allows for multi-perspective analysis and robustness checks without requiring users to manually implement or configure individual tools. Empirical results demonstrate that Causal-Copilot outperforms standard baselines across multiple tasks and datasets, validating the effectiveness of its automated algorithm selection and parameter-tuning strategies. More broadly, this work contributes to the development of intelligent, interpretable, and extensible systems for causal learning and reasoning - bringing advanced causal analysis within reach of a wider community of researchers and practitioners, and facilitating its integration into practical workflows across disciplines.

This paper is organized as follows. Section 2 presents the overall architecture of Causal-Copilot, highlighting its modular design. Section 3 details the supported causal discovery and inference algorithms, along with additional analytical tools. Section 4 describes the workflow and usage of Causal-Copilot in practical scenarios. Section 5 reports extensive experimental results across diverse datasets, providing comprehensive evaluations and comparisons under various settings.

\section{Architecture}
\begin{figure*}
  \centering
  \includegraphics[scale=0.55]{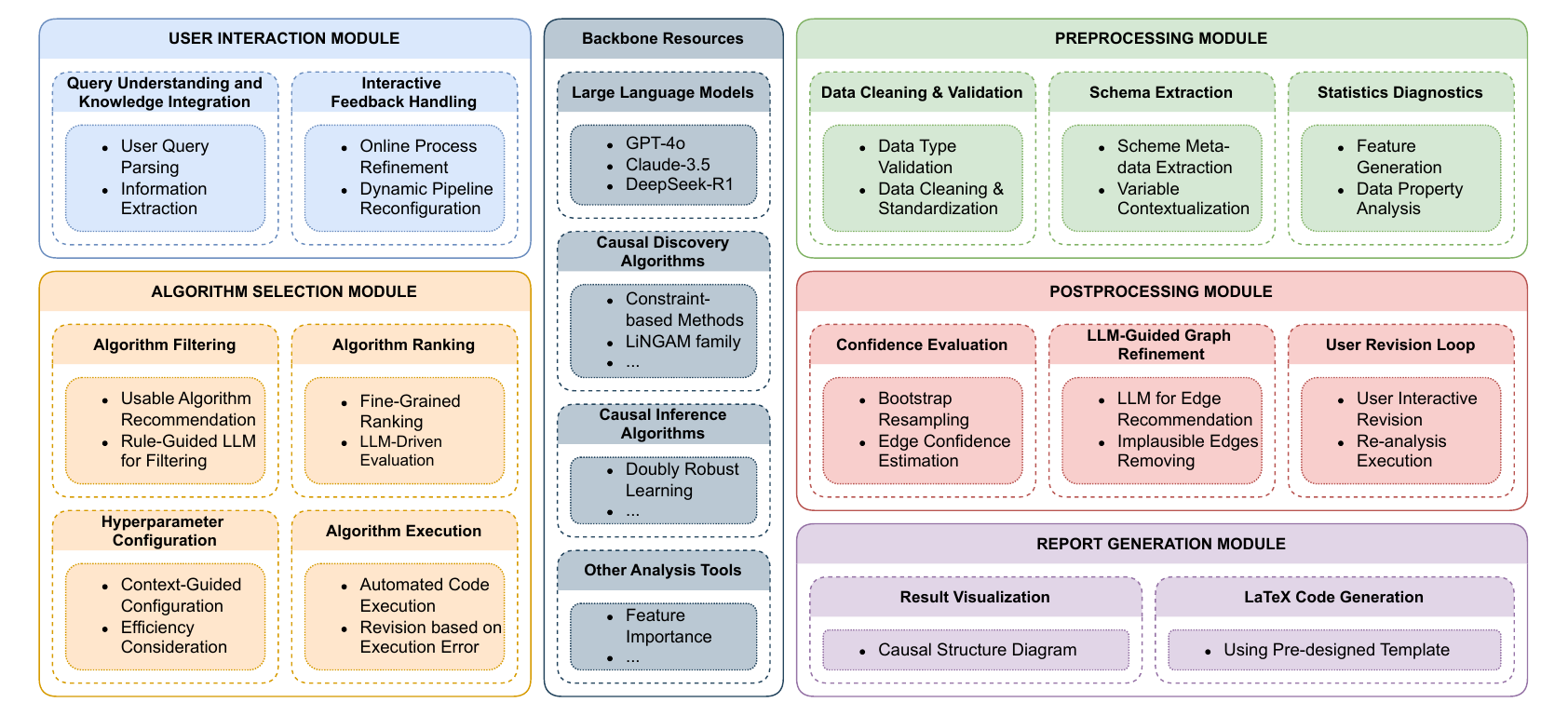}
  \caption{The overall architecture of our Causal-Copilot.}
  \label{approach}
\end{figure*}


This section provides an overview of the architecture of Causal-Copilot, which is designed as a modular framework comprising five core components: Simulation, User Interaction, Preprocessing, Algorithm Selection, Postprocessing. These modules form a cohesive workflow for end-to-end causal analysis. At the center of the system, an LLM orchestrates the overall workflow by directing information flow between modules and leveraging external resources, such as a causality-specific knowledge memory and local algorithm libraries. Communication between modules is facilitated through standardized interfaces that transmit structured metadata and intermediate outputs. This design enables the LLM to manage the execution process effectively while preserving modularity, extensibility, and ease of integration across the system.



\subsection{User Interaction Module}
The user interaction module is mainly used to collect and process queries and user feedback. With the module, users can ask any causal questions in natural language, then our Causal-Copilot will propose a plan, execute the analysis, and return results. In the process, users can engage in follow-up conversations to refine the analysis, or ask for alternative methods, without requiring specialized technical vocabulary or formal query syntax. Concretely, this module serves two main functions:

\begin{itemize}

    \item \emph{Query Understanding and Knowledge Integration:} Given the input dataset (tabular, time series, or panel data), users can provide natural language queries specifying analysis objectives, domain constraints, prior knowledge, and algorithmic preferences. For example, users might request "Find causal relationships between smoking and cancer while controlling for socioeconomic factors," specify algorithm preferences like "Use the PC algorithm," or impose constraints such as "Variable 'age' cannot be an effect of any other variable." The system employs an LLM to parse these queries and extract structured information that guides the subsequent analysis pipeline while respecting domain-specific knowledge.

    \item \emph{Interactive Feedback Loop:} Throughout the causal analysis workflow, users can intervene at strategic checkpoints to refine the process. The system implements a comprehensive feedback mechanism with predefined interaction points where users may modify algorithm selections, adjust parameter settings, incorporate additional constraints, or revise variable relationships. At each checkpoint, the LLM processes natural language feedback, translates it into actionable modifications, and dynamically reconfigures the analysis pipeline while maintaining consistency with previous steps. This bidirectional communication ensures that domain expertise can be seamlessly integrated at any stage, enhancing both the accuracy and relevance of the results.

\end{itemize}

This module is designed to be both user-friendly and robust. Depending on the application, it can be realized via a command-line interface, a REST API, or a graphical user interface that displays interactive forms and visualizations. 

\subsection{Preprocessing Module}
The Preprocessing Module transforms user-provided raw data into a standardized format suitable for downstream causal analysis. It performs the following key functions:

\begin{itemize}

    \item \emph{Data Cleaning and Validation:} The module implements a comprehensive suite of data preparation functions including missing value detection and imputation for both tabular and time-series data, handles constant features, and ensures data consistency across observations. The system automatically determines appropriate transformations based on data characteristics, such as scaling continuous variables, encoding categorical features, and segmenting time-series data into meaningful windows.

    \item \emph{Schema Extraction:} The system extracts schema-level metadata-such as variable names and types-and stores them in memory for downstream use. An LLM is further employed to generate contextual insights about the variables and collect relevant domain knowledge, offering high-level guidance for subsequent analysis.
    
    \item \emph{Statistics Diagnostics:} The module performs extensive statistical analysis critical for causal analysis through specialized diagnostic functions. For tabular data, it conducts correlation analysis, linearity checks, Gaussian distribution tests, and heterogeneity assessments to characterize variable relationships and distributions. For time-series data, it implements stationarity checks and lag estimation to identify temporal dependencies. These diagnostics serve dual purposes: they guide algorithm selection by determining which causal methods are statistically appropriate for the specific data structure, and they identify potential confounding factors or statistical anomalies that might influence causal conclusions. For example, detecting non-Gaussian distributions might indicate the need for nonparametric methods, while non-linearity results would suggest more flexible causal discovery approaches. Additionally, the module incorporates exploratory data analysis capabilities to help users gain deeper insights into their data including variable distributions, correlation analysis and etc.

\end{itemize}

Given the sensitivity of causal analysis to data characteristics and variable semantics, careful preprocessing is essential. To ensure reliability and transparency, the module logs each transformation step for traceability, allowing users to audit or revert changes when necessary.

\subsection{Algorithm Selection Module}

The Algorithm Selection Module serves as the central decision-making component of Causal-Copilot, responsible for selecting appropriate causal analysis algorithms and configuring their hyperparameters to align with the user’s objectives. Its core logic includes the following steps:

\begin{itemize}

    \item \emph{Algorithm Filtering:} This step systematically narrows down the vast algorithm space by leveraging user requirements, metadata, and diagnostics from preprocessing. The filtering logic is encoded in the causality-specific knowledge memory, which contains detailed algorithm tags with conditional performance ratings. These tags provide the LLM with concise and informative perspectives to reason about algorithm selection. The system first identifies critical compatibility requirements—such as time-series structure, variable count, or heterogeneity patterns—and applies these as primary filters. This structured knowledge enables the LLM to make informed algorithm recommendations that respect both user constraints (such as runtime limitations) and dataset characteristics, while maintaining algorithmic diversity by selecting methods from different methodological families when appropriate. This approach ensures that only algorithms fundamentally compatible with the data characteristics remain in the candidate pool, preventing the selection of methods that would produce misleading results regardless of their general performance ratings.
  

    \item \emph{Algorithm Ranking:} After filtering, the module implements a dual-criterion evaluation framework that integrates encoded causal expert knowledge with empirical evidence. The LLM leverages corresponding algorithm profiles containing domain-specific applications, theoretical guarantees, and known limitations to assess theoretical properties. Simultaneously, it retrieves and analyzes benchmarking results across diverse scenarios to evaluate empirical performance. This knowledge-enhanced decision process identifies algorithms that not only satisfy formal statistical requirements but also demonstrate proven effectiveness in comparable real-world contexts. By systematically balancing theoretical foundations with practical performance metrics, the system selects methods that optimize both statistical validity and computational efficiency for the specific causal discovery task at hand.

    
    \item \emph{Hyperparameter Configuration:} After algorithm selection, the module optimizes hyperparameters by first honoring explicit user constraints, then applying a context-sensitive configuration strategy. The LLM leverages dataset-specific insights and algorithm profiles from the knowledge base to determine appropriate parameter values. This process balances statistical rigor with computational efficiency by evaluating critical trade-offs for each parameter setting. The resulting configuration is tailored to the specific causal discovery task while respecting both technical requirements and the user's analytical objectives, ensuring the selected algorithm performs optimally within the given constraints.


    \item \emph{Algorithm Execution:} The module invokes the selected algorithm with the configured hyperparameters through an integrated codebase. If execution errors occur, the LLM interprets the error messages and iteratively revises the algorithm or hyperparameter settings to ensure successful completion.

\end{itemize}

By automating algorithm and hyperparameter selection, this module reduces reliance on expert intuition, enabling more consistent and accessible causal analysis for users with varying levels of domain knowledge. Furthermore, multiple causal inference methods and auxiliary analysis tools such as feature importance attribution and abnormal detection can be applied to the discovered graphs for deeper causal insights.

\subsection{Postprocessing Module}

To enhance the accuracy and robustness of the learned causal graph, Causal-Copilot incorporates a postprocessing module that performs graph tuning by combining statistical confidence estimates, LLM reasoning, and user input together. 

\begin{itemize}
\item \emph{Confidence Evaluation:} To enhance the reliability of the causal graph, we implement a statistical validation approach that quantifies the confidence level of each potential causal relationship. Using bootstrap resampling techniques, we systematically evaluate edge confidence, automatically strengthening the graph by adding highly confident missing edges, removing statistically weak connections, and identifying uncertain relationships that require further assessment.

\item \emph{LLM-Guided Graph Refinement:} Moderate-confidence edges are passed to the LLM, which assesses their conceptual plausibility and directionality using domain-relevant knowledge. To ensure both local consistency and global plausibility, the LLM operates in a structured, node-centric manner. The LLM may reintroduce plausible but statistically weak edges, or remove or redirect statistically present but conceptually implausible edges.
To preserve statistical integrity, LLM recommendations are treated as soft suggestions and are not permitted to override high-confidence decisions made by the bootstrap procedure.


\item \emph{User Revision Loop:} The module supports interactive refinement, allowing users to modify assumptions or exclude implausible edges in the causal graph. Following user input, the system can automatically re-execute relevant portions of the pipeline to incorporate the updated constraints.

\end{itemize}

By combining statistical confidence, LLM reasoning, and user input, this module enables flexible yet principled refinement of causal graphs beyond purely algorithmic outputs. Support for both automated validation and manual control ensures that the final output is accurate, interpretable, and aligned with domain knowledge.

\subsection{Report Generation Module}

Upon completion of the main causal analysis, the Report Generation Module compiles results into a structured and user-friendly output, \ie a PDF report. It includes the following key components:

\begin{itemize}
\item \emph{Result Visualization:} Outputs from causal analysis algorithms—such as causal graphs and effect estimates—often require expert knowledge to interpret. This module provides intuitive visualizations to help users understand and explore the results. For instance, it generates graphical representations of the discovered causal structure, where nodes denote variables and directed edges indicate causal relationships. Moreover, other elements, such as estimated causal effects, will be highlighted to enhance interpretability. Additional examples are provided in the case study section.

    \item \emph{LaTeX Code Generation:} All intermediate and final results are organized into a comprehensive PDF report compiled based on latex code, which includes dataset summaries, the rationale for algorithm selection, and visual representations of results (e.g., descriptive statistics and estimated causal graphs). An LLM populates a pre-designed LaTeX template to ensure clarity, coherence, and contextual relevance.

    
\end{itemize}

By prioritizing clarity, interpretability, and user control, the Report Generation Module ensures that complex analytical results are accessible to a broad range of users and that key assumptions and uncertainties are explicitly communicated.

\section{Supported Causal Analysis Algorithms}


Causal-Copilot integrates over twenty state-of-the-art causal analysis algorithms, broadly categorized into \emph{causal discovery}, \emph{causal inference}, and \emph{auxiliary analysis tools}. This section provides an overview of the supported methods, organized by methodological families and assumptions.

\subsection{Causal Discovery}


Causal discovery algorithms aim to uncover the underlying causal structure among variables from observational data. Causal-Copilot supports a diverse set of discovery methods selected for their theoretical soundness, empirical robustness, and practical applicability. These methods are organized into categories based on their underlying assumptions and algorithmic paradigms.

\begin{itemize}

    \item \textbf{Constraint-based Methods}: These methods infer causal structures by systematically performing conditional independence (CI) tests to identify the skeleton and orient edges of a causal graph. By using various CI tests (e.g., partial correlation, kernel-based, or mutual information tests), these methods can flexibly adapt to different data scenarios including linear, nonlinear, continuous, and discrete variables. The constraint-based family includes numerous algorithms with varying capabilities: PC \citep{spirtes2000causation}, which identifies a Markov equivalence class under causal sufficiency; FCI \citep{spirtes1995causal} and its efficient variant RFCI \citep{colombo2012learning}, which handle latent confounders; conservative approaches like CPC and BCCD \citep{ramsey2012adjacency, claassen2012bayesian} that reduce orientation errors. For heterogeneous or nonstationary data, CD-NOD \citep{huang2020causal} leverages changes in causal mechanisms to improve identifiability. Time series data benefit from specialized algorithms like PCMCI and its variants,  \citep{runge2019detecting, runge2020discovering,gerhardus2020high,gunther2023causal}, which efficiently captures both contemporaneous and lagged causal relationships. 
    
    In Causal-Copilot, among these, we integrate PC for its foundational role, consistent stability and computational efficiency, FCI for robust handling of hidden confounders, CD-NOD for its unique ability with heterogeneous data, and PCMCI for time series analysis—collectively providing a practical balance of theoretical soundness, computational feasibility, and applicability across diverse data characteristics.


    \item \textbf{Score-based Methods}: These methods identify causal structures by optimizing predefined scoring functions that evaluate how well candidate graphs explain observed data. Classical algorithms include Greedy Equivalence Search (GES)~\citep{chickering2002optimal}, which performs a two-phase search over Markov equivalence classes, and its scalable extension Fast GES (FGES)~\citep{ramsey2017million} that efficiently handles high-dimensional data. Recent advancements include eXtremely Greedy Equivalence Search (XGES)~\citep{nazaret2025extremely}, which introduces an improved heuristic to reduce convergence to local minima, and Greedy Relaxation of the Sparsest Permutation (GRaSP)~\citep{lam2022greedy}, which uses a permutation-based approach with progressive relaxation of sparsity assumptions, showing superior performance on dense graphs. Beyond these integrated algorithms, the field has expanded with methods addressing specific challenges, such as Generalized Score Functions using Reproducing Kernel Hilbert Space for handling nonlinear relationships, Bayesian approaches ~\citep{heckerman2006bayesian} and Dynamic Programming~\citep{koivisto2004exact} for improved accuracy in smaller networks, and Score-and-Search algorithms like hill-climbing~\citep{tsamardinos2006max} for trading off optimality for computational efficiency. 
    
    We incorporate a diverse set of these algorithms (GES, XGES, GRaSP, FGES) to provide comprehensive coverage across different data characteristics and computational constraints, allowing Causal-Copilot to maximally support a diverse range of downstream application scenarios.

    \item \textbf{Continuous Optimization-based Methods}: These methods reformulate causal discovery as a differentiable optimization problem with smooth or relaxed acyclicity constraints, enabling efficient, gradient-based learning of causal graphs. Specifically, NOTEARS (Linear)\citep{zheng2018dags} models linear structural equation models and introduces a continuous characterization of acyclicity for scalable structure learning. NOTEARS (Nonlinear)\citep{zheng2020learning} extends this framework to nonlinear settings by incorporating nonparametric sparsity through partial derivatives. DYNOTEARS\citep{pamfil2020dynotears} adapts NOTEARS for time-series data, learning dynamic Bayesian networks that capture both contemporaneous and lagged dependencies. NTS-NOTEARS\citep{sun2021nts} further enhances temporal modeling by integrating convolutional networks and prior knowledge. GOLEM\citep{ng2020role} simplifies the objective by directly optimizing a linear SEM loss with soft sparsity and acyclicity constraints, improving scalability. CALM\citep{jin2024revisiting} addresses non-convexity by combining L0-penalized likelihood with hard acyclicity enforcement in a hybrid differentiable framework. Moreover, CORL~\citep{wang2021ordering} formulates causal discovery as a sequential decision-making process, using a Markov Decision Process and an encoder-decoder architecture trained via reinforcement learning to infer causal orderings. 

    \item \textbf{MB-based Methods}: Markov Blanket (MB)-based methods aim to identify the minimal set of variables (i.e., parents, children, and spouses) that renders a target variable conditionally independent from all others, facilitating efficient causal discovery and feature selection. InterIAMB\citep{tsamardinos2003algorithms} combines forward selection and backward elimination to iteratively refine the MB with fewer conditional independence tests. IAMBnPC\citep{tsamardinos2003algorithms} augments IAMB with a correction phase to reduce false negatives in parent-child identification. HITON-MB\citep{aliferis2003hiton} adopts a divide-and-conquer approach by separately identifying the parent-child (PC) and spouse (SP) sets, improving robustness in complex domains. MBOR\citep{de2008novel} incorporates the "OR" rule and limits the size of conditioning sets to enhance reliability, particularly in small-sample or low-signal settings. BAMB~\citep{ling2019bamb} simultaneously identifies all MB components, offering improved efficiency in both computation and data usage. These methods are motivated by the Causal Markov condition, which states that a node is conditionally independent of all variables which are not effects or direct causes of that node, given its direct causes. 
    
    Since MB-based methods focus on finding the Markov blanket for each target variable individually, they present a natural tradeoff between efficiency (as each target variable's MB discovery can be parallelized) and precision (as each iteration focuses only on a local structure). In Causal-Copilot, we extend these methods to enable full causal graph discovery by using constraint-based methods (e.g., PC) to distinguish between parents, children, and spouses within each identified MB, and then aggregating these local structures into a complete CPDAG representation of the causal network. This approach leverages the screening property of the Causal Markov condition, where parents of a variable screen it from other indirect causes, while maintaining computational efficiency through parallelization.

    \item \textbf{Granger Causality Methods}: Granger causality methods evaluate whether past values of one time series help predict another, providing a statistical framework for inferring temporal causal relationships. Linear Granger causality employs vector autoregressive (VAR) models that compare prediction error variances between full and reduced models using F-tests or likelihood ratio tests \citep{granger1969investigating}. Non-linear extensions include kernel-based methods \citep{marinazzo2008kernel}, transfer entropy \citep{schreiber2000measuring}, and neural network implementations \citep{tank2021neural, nauta2019causal}. 
    
    Causal-Copilot incorporates both pairwise and multivariate Granger causality tests \citep{barrett2010mvgc} with configurable lag selection and significance thresholds, enabling flexible temporal causal discovery for time series data.

    \item \textbf{Functional Causal Model-based Methods}: Functional causal models (FCMs) specify the data-generating process using functions and noise distributions, enabling identifiability of causal structures under various assumptions. The Linear Non-gaussian Additive Model (LiNGAM) family represents a classical approach, assuming linear relationships with non-Gaussian independent noise. ICA-LiNGAM \citep{shimizu2006linear} applies Independent Component Analysis to recover causal structures from observational data, while DirectLiNGAM \citep{shimizu2011directlingam} improves stability in small-sample or noisy settings by recursively identifying exogenous variables through independence tests. Beyond linear models, Additive Noise Models (ANMs) \citep{hoyer2008nonlinear} extend to nonlinear functions with additive noise, maintaining identifiability under mild conditions. Post-Nonlinear Models (PNL) \citep{zhang2012identifiability} further generalize this framework by incorporating invertible transformations after the additive noise process, addressing scenarios with measurement distortions. For time-series data, VAR-LiNGAM \citep{hyvarinen2010estimation} integrates Vector Autoregressive modeling with causal discovery. 
    
    We strategically deploy these algorithms based on their specific strengths: LiNGAM variants are primarily used for linear systems with non-Gaussian noise, where they provide strong statistical guarantees and complete identifiability; ANMs and PNLs serve as refinement tools within our hybrid framework to determine directions of undirected edges produced by constraint-based or score-based methods; and specialized time-series variants address temporal dependencies.
    
    \item \textbf{Hybrid Methods}: Causal-Copilot also incorporates hybrid approaches that combine the strengths of global structure discovery and pairwise causal direction identification. Specifically, it integrates constraint-based methods and score-based methods (e.g., PC~\citep{spirtes1991algorithm}, GES~\citep{chickering2002optimal}) with bivariate functional causal model-based approaches (e.g., ANM~\citep{hoyer2008nonlinear}, PNL~\citep{zhang2012identifiability}) to determine edge orientations. Starting from a skeleton or partially oriented graph obtained via global methods, a second-stage orientation procedure is applied to infer causal directions between variable pairs. The final output is a maximally oriented DAG that combines statistical consistency with functional identifiability.

    \item \textbf{Algorithm Acceleration}: 
    Recent advances in causal discovery have prioritized computational efficiency to handle large-scale datasets. Hardware acceleration approaches include FPGA-based methods that shift the computational bottleneck from executing conditional independence (CI) tests to generating them, achieving up to 8.8x speedup over GPU implementations \citep{guo2023fpga}. GPU-accelerated techniques like GPUCMIknn-Parallel and cuPC parallelize CI tests, demonstrating speedups of up to 1000x compared to single-threaded CPU implementations \citep{hagedorn2022gpu,zarebavani2019cupc}. Parallelized versions of constraint-based algorithms such as Parallel-PC distribute computations across multiple cores, reducing runtime exponentially \citep{le2016fast}. Divide-and-conquer approaches partition large graphs into smaller subgraphs for parallel processing while preserving accuracy via Markov blanket inclusion \citep{dong2024dcdilp}. Continuous optimization methods like NOTEARS use smooth acyclicity constraints for efficient gradient-based learning with batch-wise optimization on GPU \citep{zheng2018dags,ng2020role,zheng2020learning}, while pruning strategies such as the Dsep-CP algorithm reduce CI test requirements significantly. These acceleration techniques collectively make causal discovery feasible for high-dimensional datasets. 
    
    To address scalability challenges in Causal-Copilot, we implemented several optimizations: all conditional independence tests are modified to batch-wise operation for CPU parallelization; GPU-accelerated skeleton discovery using Fisher-Z, Chi-squared, and CMIknn tests is integrated for both linear and non-linear data in homogeneous (PC) and heterogeneous (CD-NOD) scenarios; FGES is incorporated for efficient discovery on large-scale sparse graphs; GPU-optimized DirectLiNGAM and VAR-LiNGAM \citep{akinwande2024acceleratedlingam} handle non-Gaussian data in both tabular and time-series formats; and GPU-trainable continuous optimization methods (e.g., GOLEM, NOTEARS) are integrated to leverage modern hardware acceleration for complex causal structures.

\end{itemize}

\subsection{Causal Inference}
Causal inference aims to estimate the effects of interventions under a known or assumed causal structure. Causal-Copilot supports a range of both classical and modern techniques, enabling flexible estimation of treatment effects across diverse settings.

\begin{itemize}
    \item \textbf{Double Machine Learning}: Double Machine Learning (DML) \citep{chernozhukov2018double} is a powerful framework for causal effect estimation that leverages machine learning to control for high-dimensional confounders while maintaining valid statistical inference. \citet{econml} developed variants of DML to suit different data structures and modeling needs, including LinearDML, SparseLinearDML, and CausalForestDML. LinearDML assumes a linear relationship between treatment and outcome, enabling interpretable effect estimates while using machine learning to flexibly model nuisance functions. SparseLinearDML extends this by incorporating regularization techniques, making it suitable for high-dimensional settings where only a subset of covariates are relevant. CausalForestDML replaces linear models with random forests to capture complex, non-linear, and heterogeneous treatment effects across subpopulations. 

    \item \textbf{Doubly Robust Learning}: Doubly Robust Learning (DRL) \citep{hlynsson2024tutorial} is a powerful framework in causal inference that combines outcome modeling and treatment modeling to estimate treatment effects with increased robustness. Its key advantage lies in the doubly robust property: the estimator remains consistent if either the outcome model or the propensity score model is correctly specified, offering protection against model misspecification. \citet{econml} developed several specialized versions of DRL to suit different modeling needs. LinearDRL assumes a linear relationship between covariates and outcomes, providing a straightforward and interpretable approach for estimating treatment effects. SparseLinearDRL extends this by incorporating regularization techniques to handle high-dimensional data, enabling variable selection and mitigating overfitting when only a subset of covariates is truly influential. On the other hand, ForestDRL leverages random forests to capture complex, nonlinear relationships and heterogeneous treatment effects across subpopulations. 
    
    \item \textbf{Instrumental variable (IV) methods}: The DRIV (Double Robust Instrumental Variables) family of estimators, introduced in the EconML library \citep{econml}, enables robust estimation of heterogeneous treatment effects in the presence of endogeneity using instrumental variables. The base DRIV method combines outcome and treatment models for double robustness. LinearDRIV assumes a linear treatment effect model for interpretability, while SparseLinearDRIV extends this to high-dimensional settings using techniques like the Debiased Lasso for variable selection. ForestDRIV replaces the linear model with a non-parametric regression forest, capturing complex, nonlinear relationships between covariates and treatment effects. 

    \item \textbf{PSM/CEM}: Propensity Score Matching (PSM) \citep{rosenbaum1983central} and Coarsened Exact Matching (CEM) \citep{iacus2012causal} are widely used methods for reducing confounding in observational causal inference. PSM estimates the probability of treatment assignment (the propensity score) given observed covariates, then matches treated and control units with similar scores to mimic a randomized experiment. While it is flexible and widely adopted, PSM does not guarantee covariate balance and may suffer from poor overlap between groups. In contrast, CEM performs matching by temporarily coarsening covariates into discrete bins and then finding exact matches across treatment groups. This approach guarantees covariate balance by design, though it may exclude units without matches, potentially reducing the sample size. In Causal-Copilot, CEM is used when the number of discrete variables is greater than that of the continuous variable. Otherwise, PSM is used for analysis.

     \item \textbf{Counterfactual Estimation}: Causal-Copilot enables counterfactual estimation by assessing potential outcomes under different interventions via DoWhy Framework \citep{dowhy}. It constructs and utilizes causal models based on provided or recovered causal graph, to estimate the effects of hypothetical changes in treatment on outcomes of interest. This integration allows for rigorous analysis of counterfactual queries, facilitating informed decision-making and policy evaluation. For instance, users can estimate how modifying a specific variable might have altered an observed outcome, providing valuable insights into causal relationships within the data. 
\end{itemize}

\subsection{Auxiliary Analysis Tools}

Beyond core causal discovery and inference, Causal-Copilot includes complementary tools that enhance interpretability and leverage causal structures, including \textit{Model Explanation} and \textit{Causal-Structure-Based Anomaly Detection}.

\begin{itemize}
    \item \textbf{Feature Importance}: To interpret model predictions and assess the contribution of individual covariates, Causal-Copilot incorporates feature importance analysis using SHAP (SHapley Additive exPlanations) values \citep{lundberg2017unified}. SHAP values are grounded in cooperative game theory and quantify each feature’s marginal contribution to a model’s prediction by considering all possible combinations of feature subsets. This approach provides consistent and locally accurate attributions, enabling users to understand how each variable influences estimated outcomes or treatment effects. By aggregating SHAP values across samples, Causal-Copilot generates global feature importance rankings, facilitating transparent and explainable causal analyses, especially in complex, non-linear models such as tree ensembles or neural networks.

    \item \textbf{Abnormality Attribution}: To enhance interpretability and support diagnostic tasks, Causal-Copilot incorporates abnormality attribution through causal structure-based root cause analysis of outliers \citep{budhathoki2022causal}. This method leverages the inferred causal graph to trace the propagation of anomalous behavior across variables. When an outlier is detected in a downstream variable, the system traverses upstream causal paths to identify potential root causes—variables whose unusual values or perturbations are most likely responsible for the observed abnormality.
\end{itemize}



\begin{table*}[t]
  \centering
  \small
  \renewcommand\arraystretch{1.1}
  \scalebox{0.75}{
  \begin{tabular}{cllll}
    \toprule
    & \textbf{Algorithm} & \textbf{Data Type} & \textbf{Family} & \textbf{Acceleration} \\
    \midrule
    \multirow{24}*{\textbf{Causal Discovery}} &
    PC & Tabular (Flexible) & Constraint-based & CPU, GPU \\
    & FCI & Tabular (Flexible) & Constraint-based & CPU \\
    & CD-NOD & Tabular (Flexible) & Constraint-based & CPU, GPU \\
    & GES & Tabular (Flexible) & Score-based & - \\
    & FGES & Tabular (Linear) & Score-based & - \\
    & XGES & Tabular (Linear) & Score-based & - \\
    & GRaSP & Tabular (Flexible) & Score-based & - \\
    & ICA-LiNGAM & Tabular (Linear) & LiNGAM & - \\
    & DirectLiNGAM & Tabular (Linear) & LiNGAM & GPU \\
    & NOTEARS (Linear) & Tabular (Linear) & Continuous-opt & GPU \\
    & NOTEARS (Nonlinear) & Tabular (Nonlinear) & Continuous-opt & GPU \\
    & GOLEM & Tabular (Linear) & Continuous-opt & GPU \\
    & CALM & Tabular (Linear) & Continuous-opt & GPU \\
    & CORL & Tabular (Linear) & Continuous-opt & GPU \\
    & InterIAMB & Tabular (Flexible) & MB-based & CPU \\
    & IAMBnPC & Tabular (Flexible) & MB-based & CPU \\
    & HITON-MB & Tabular (Flexible) & MB-based & CPU \\
    & MBOR & Tabular (Flexible) & MB-based & CPU \\
    & BAMB & Tabular (Flexible) & MB-based & CPU \\
    & Hybrid & Tabular (Flexible) & Hybrid & CPU \\
    & PCMCI & Time Series (Flexible) & Constraint-based & CPU \\
    & VAR-LiNGAM & Time Series (Linear) & LiNGAM & GPU \\
    & DYNOTEARS & Time Series (Linear) & Continuous-opt & GPU \\
    & NTS-NOTEARS & Time Series (Nonlinear) & Continuous-opt & GPU \\
    \midrule
    \multirow{9}*{\textbf{Causal Inference}} & LinearDML & Tabular (Linear) & Double ML & - \\
    & SparseLinearDML & Tabular (Linear) & Double ML & - \\
    & CausalForestDML & Tabular (Nonlinear) & Double ML & - \\
    & LinearDRL & Tabular (Linear) & Doubly Robust & - \\
    & SparseLinearDRL & Tabular (Linear) & Doubly Robust & - \\
    & ForestDRL & Tabular (Nonlinear) & Doubly Robust & - \\
    & DRIV Family & Tabular (Flexible) & Instrumental Var & - \\
    & PSM & Tabular (Flexible) & Matching & - \\
    & CEM & Tabular (Flexible) & Matching & - \\
    & Counterfactual Estimation & Tabular (Flexible) & Counterfactual & - \\
    \midrule
    \multirow{3}*{\textbf{Auxiliary Analysis}} & Feature Importance & Mixed (Flexible) & Model Explanation & - \\
    & Abnormal Detection & Mixed (Flexible) & Root Cause Analysis & - \\
    \bottomrule
  \end{tabular}
  }
  \caption{A comprehensive list of all the supported causal analysis algorithms in Causal-Copilot. The "Data Type" column indicates the supported data formats, where "Flexible" means the algorithm can handle various data distributions and relationships (linear, nonlinear, categorical, etc.). The "Acceleration" column shows hardware optimization support, with "CPU" indicating multi-core parallelization and "GPU" denoting graphics processing unit acceleration for significantly faster computation on large datasets.}
  \label{tab:causal_algorithms}
\end{table*}

\section{Usage}
The primary goal of Causal-Copilot is to automate the entire pipeline of rigorous causal analysis. In practice, users can deploy Causal-Copilot on a local server via simple code snippets for automated processing, or interact with it directly through our web-based interface.

\begin{figure*}[htbp]
  \centering
  \includegraphics[scale=0.6]{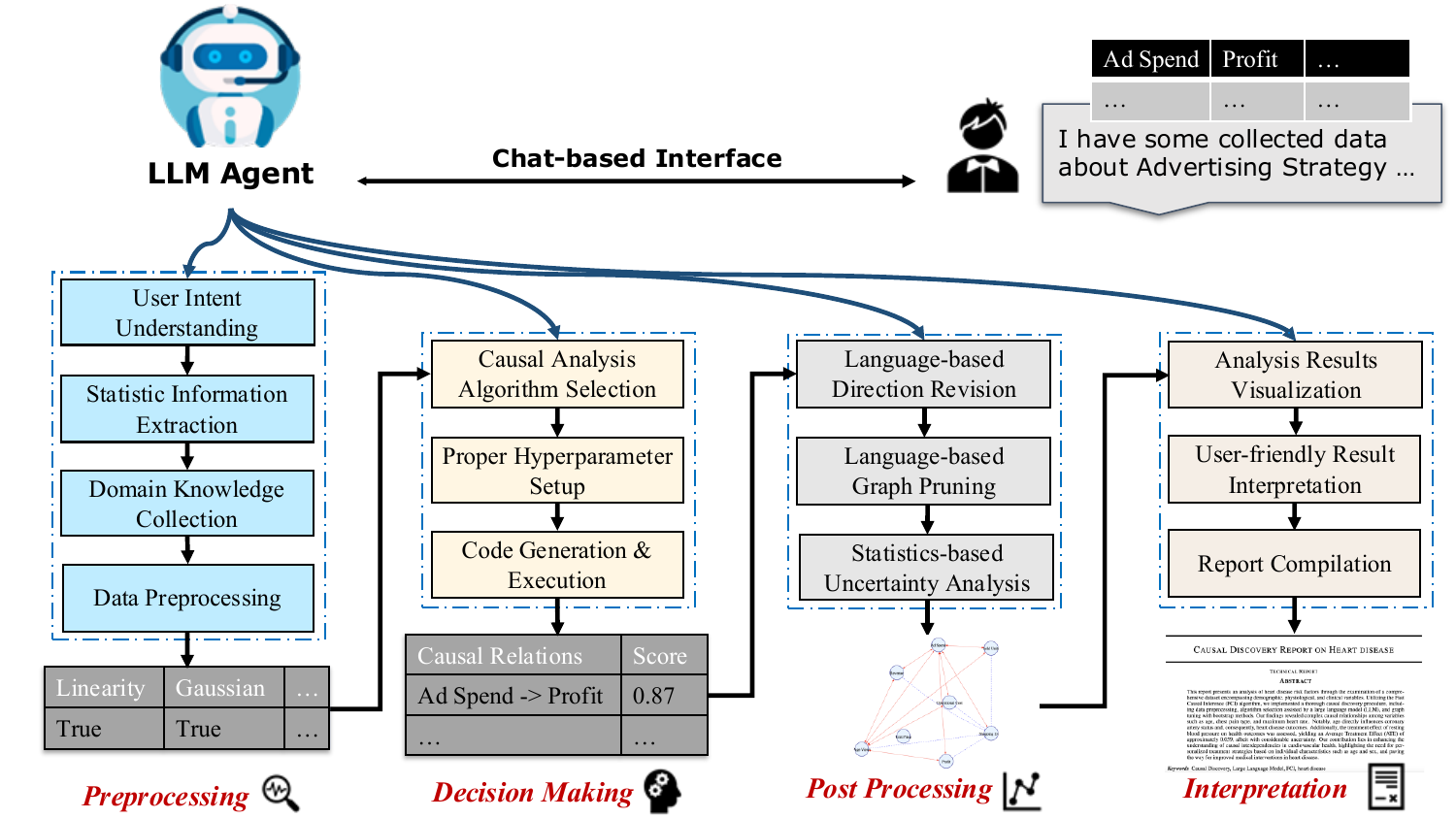}
  \caption{The overall workflow of our Causal-Copilot, with an example of discovering the causal structure in collected advertising related data.}
  \label{workflow}
\end{figure*}

\subsection{Autonomous Workflow}
By deeply integrating an LLM, our Causal-Copilot implements an autonomous workflow for end-to-end causal analysis, as shown in Figure~\ref{workflow}. Concretely, a user begins by uploading a dataset, \ie in tabular form (with rows and columns representing observations and variables, respectively), along with a natural language query specifying the analysis intent. 
Given these inputs, the LLM, supported by rule-based methods, interprets the query, extracts relevant statistical information, incorporates available domain knowledge, and preprocesses the dataset by addressing data types and handling missing values. Based on the collected metadata and inferred intent, the system autonomously selects an appropriate causal analysis algorithm, configures its hyperparameters, and generates executable code for analysis.

After executing the selected method (e.g., producing an inferred causal graph), the LLM evaluates the results, identifies potential inconsistencies, and refines the output using both internal knowledge and contextual cues. The system may also interact with the user for clarification or refinement before finalizing the results.
Note that as our Causal-Copilot has encapsulated both causal discovery and other analysis algorithms, it also supports autonomously first performing causal discovery to identify the causal structure, then selecting another causal analysis algorithm to conduct further analysis based on it, such as causal effect estimation and counterfactual estimation.


Finally, Causal-Copilot generates a user-friendly analysis report that includes dataset summaries, intermediate decisions, and visualized causal structures, accompanied by natural language explanations—making the results interpretable even for non-expert users.

\subsection{Easy Deployment and Usage}
Causal-Copilot is designed for ease of use, providing both command-line and web-based interfaces to accommodate diverse user preferences and deployment environments.

\subsubsection{Deployment via Command Line}
The command-line interface offers a transparent, scriptable environment for users who prefer direct control over the analysis process. Once initiated, Causal-Copilot autonomously performs data preprocessing, algorithm selection, and postprocessing—including visualizations such as histograms and causal graphs. Throughout the workflow, it provides real-time progress updates, displays the methods under evaluation, and concludes by generating a comprehensive PDF report summarizing key results.
Below are examples of simple commands for running and deploying our Causal-Copilot locally:

\begin{verbatim}
# Launch Causal-Copilot, support autonomous analysis through command line
python main.py --data_file data --apikey openai_key --initial_query query

# Deploy Causal-Copilot, support user-interaction in a localhost website
python Gradio/demo.py
\end{verbatim}


\subsubsection{User-Friendly Interactive Web Interface}
\begin{figure*}[htbp]
  \centering
  \includegraphics[scale=0.28]{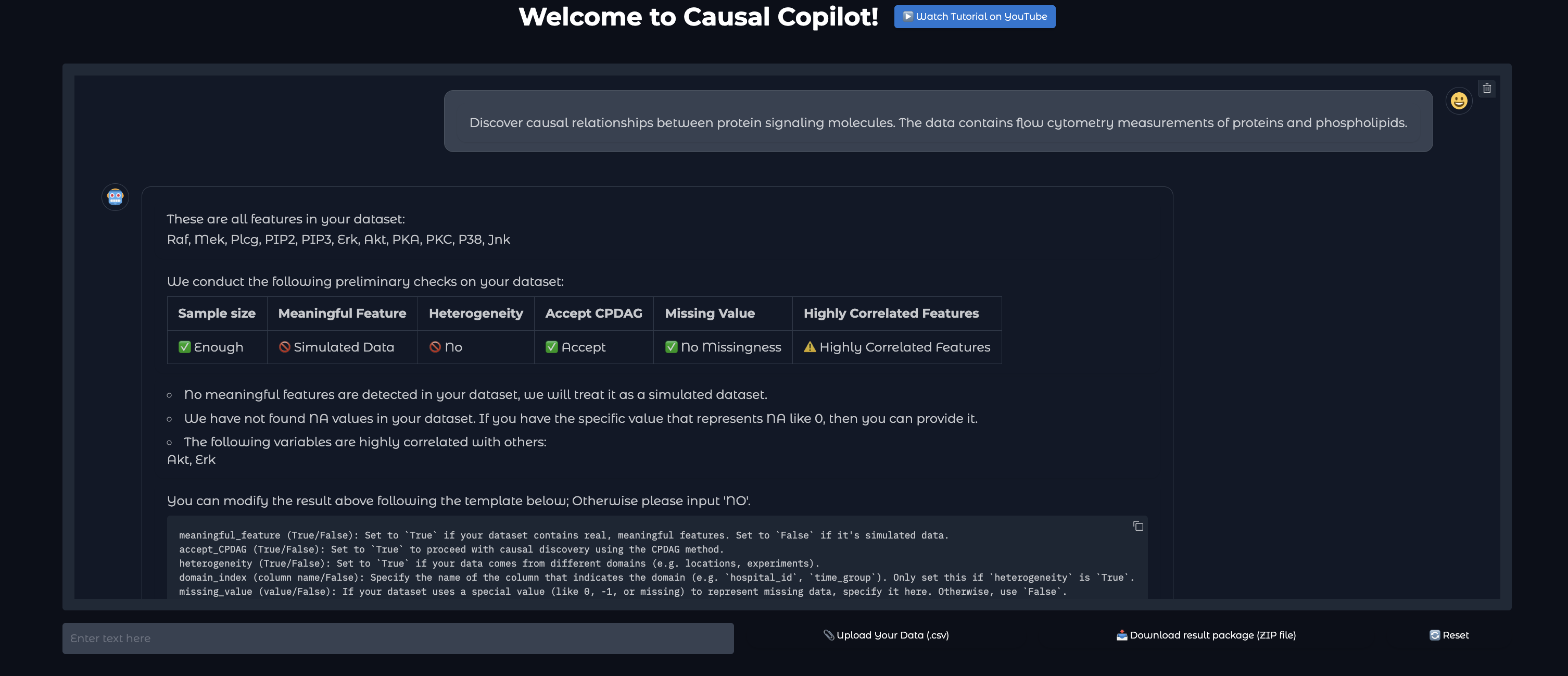}
  \caption{The website demo of our Causal-Copilot.}
  \label{demo}
\end{figure*}

To enhance accessibility and usability, Causal-Copilot is also available through a dedicated web interface. Users can upload datasets in standard formats and submit a natural language query to initiate the analysis. As shown in Figure~\ref{demo}, users may either specify key parameters manually or allow the system to automatically determine optimal settings.

As the analysis progresses, intermediate results are presented incrementally on the website. This stepwise visualization provides users with early insights and enables real-time adjustments-such as modifying assumptions or changing algorithms-before finalizing the output.


\section{Experimental Results}

In this section, we present experimental results that validate Causal-Copilot's effectiveness through two complementary approaches: (1) a comprehensive preliminary benchmarking of causal discovery algorithms to gather first-hand information about their practical characteristics beyond theoretical assumptions, and (2) a systematic evaluation of Causal-Copilot's algorithm decision-making capabilities across diverse causal discovery scenarios. These experiments demonstrate how our system leverages empirical performance data to make informed algorithm selections that outperform individual methods across challenging real-world conditions.

\subsection{Preliminary Benchmarking of Causal Discovery Algorithms}

To maximize the effectiveness of Causal-Copilot’s algorithm selection capabilities, we conducted a comprehensive preliminary benchmarking study across a wide range of causal discovery algorithms. This benchmarking aims to help address a key challenge in the field: despite the rapid proliferation of causal discovery methods, selecting the most suitable algorithm for a given application remains difficult due to overlapping assumptions and the lack of unified empirical validation—particularly for newer approaches. Our study serves two primary objectives:
\begin{itemize}
 \item Reduce the pool of candidate algorithms by filtering out those less suited for practical use in Causal-Copilot.
  \item Gather rich, first-hand performance data to support algorithm recommendations that complement theoretical assumptions and reflect how methods actually perform in similar data settings.
\end{itemize}

We developed a multi-dimensional benchmarking framework motivated by the practical challenges encountered in real-world causal discovery applications. Our approach systematically evaluates algorithm performance across dimensions that directly impact causal discovery quality and computational feasibility. 

For tabular data, we identified critical factors affecting data quality and structural complexity: variable size (dimensionality), sample size (statistical power), graph density (system complexity), function type (relationship complexity), noise distribution (robustness to non-Gaussianity), discrete variable ratio (mixed data types), measurement error (data reliability), and missing rate (data completeness). For time series data, we focused on temporal-specific challenges, including variable size (dimensionality), maximum lag (long-term dependencies), and inter/intra-temporal connection densities (temporal complexity). Recognizing the computational constraints and search complexity in large-scale applications, we implemented runtime caps to assess algorithmic efficiency—a crucial consideration when applying these methods to large and complex systems. We generate data of both linear and nonlinear function types to investigate each data property in tabular form. Specifically for time series experiments, we generate data with underlying linear relationships while ensuring stationarity. Additionally, we test different values for the important hyperparameters for most of the algorithms. For constraint-based methods and MB-based methods, we evaluate various conditional independence tests, including both linear and nonlinear approaches such as Fisher-z, RCIT, KCI, FastKCI, and CMIknn . For score-based methods like FGES and XGES, we test different sparsity regularization strengths to understand their impact on structure learning performance. We use default parameters values for continuous-optimization-based methods since it is not intuitive to choose. 

This comprehensive evaluation framework enables us to precisely characterize each algorithm's performance profile under various conditions, providing an empirical foundation for algorithm selection in Causal-Copilot. The following figures (see Figure 4-21) illustrate key findings from our preliminary benchmarking on both tabular and time-series causal discovery algorithms; complete benchmarking setup and detailed comparison are given in Appendix \ref{app:pre_exp_setup} and \ref{app:pre_exp_res}. We exclude algorithms that often exceed the time limit from our analysis, even if they perform well, as they may not be scalable for practical use. 

\subsubsection{Tabular data}
\paragraph{Overall Performance Trends.}
Continuous optimization-based methods like GOLEM and NOTEARS (Linear) excel in scenarios with linear relationships due to their mathematical foundation optimized for global score functions. This advantage enables them to effectively recover causal structures in well-behaved data. However, their performance significantly decreases when confronted with non-linear relationships, revealing a fundamental limitation. In contrast, constraint-based methods paired with appropriate independence tests, such as PC, FCI and IAMBnPC with Randomized Conditional Independence Test (RCIT), demonstrate remarkable versatility across both linear and non-linear relationships. While they may not reach the peak performance of score-based methods in purely linear settings, their consistent performance across diverse data types makes them more reliable choices for real-world applications where relationship types are often unknown or mixed.

\begin{figure}[htbp]
  \centering
  \includegraphics[width=\textwidth]{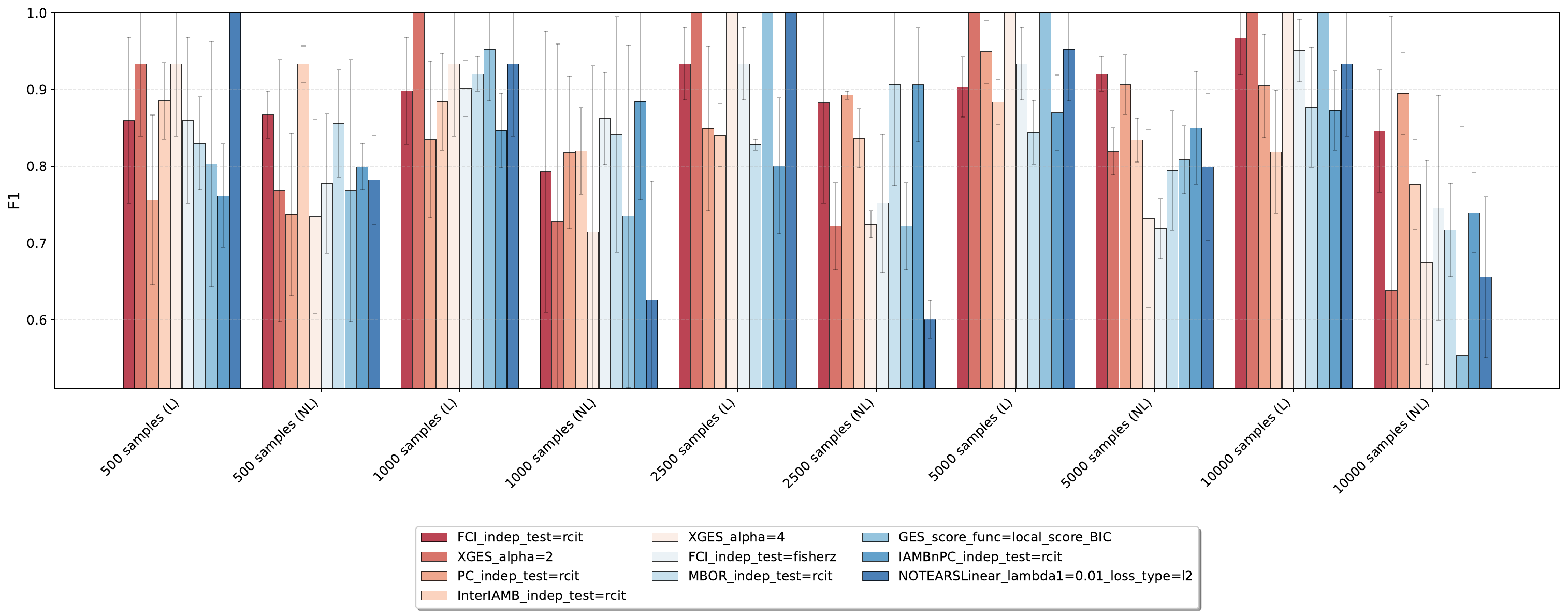}
  \caption{Performance vs. sample size for tabular causal discovery algorithms. Only the top 10 algorithms with the highest average F1 scores are shown. 'L' and 'NL' indicate results on linear and non-linear data, respectively.}
  \label{fig:perf_vs_sample_size}
\end{figure}

\paragraph{Linear vs. Non-Linear Performance.}
A distinct performance divide emerges between linear and non-linear scenarios in causal discovery (see Figure \ref{fig:perf_vs_sample_size}). Score-based algorithms including GOLEM, NOTEARS (Linear), and XGES dominate in linear settings, achieving superior performance scores. This dominance stems from their optimization approaches being particularly well-suited for linear structural equation models. Conversely, for non-linear relationships, independence-test based methods such as InterIAMB, BAMB, and PC with RCIT and KCI (Kernel Independence Test) perform substantially better. This pattern underscores the critical importance of algorithm selection based on the expected relationship type in a given domain. When relationship types remain unknown, constraint-based methods with appropriate independence tests offer more reliable performance across varied data characteristics.

\begin{figure}[htbp]
  \centering
  \includegraphics[width=\textwidth]{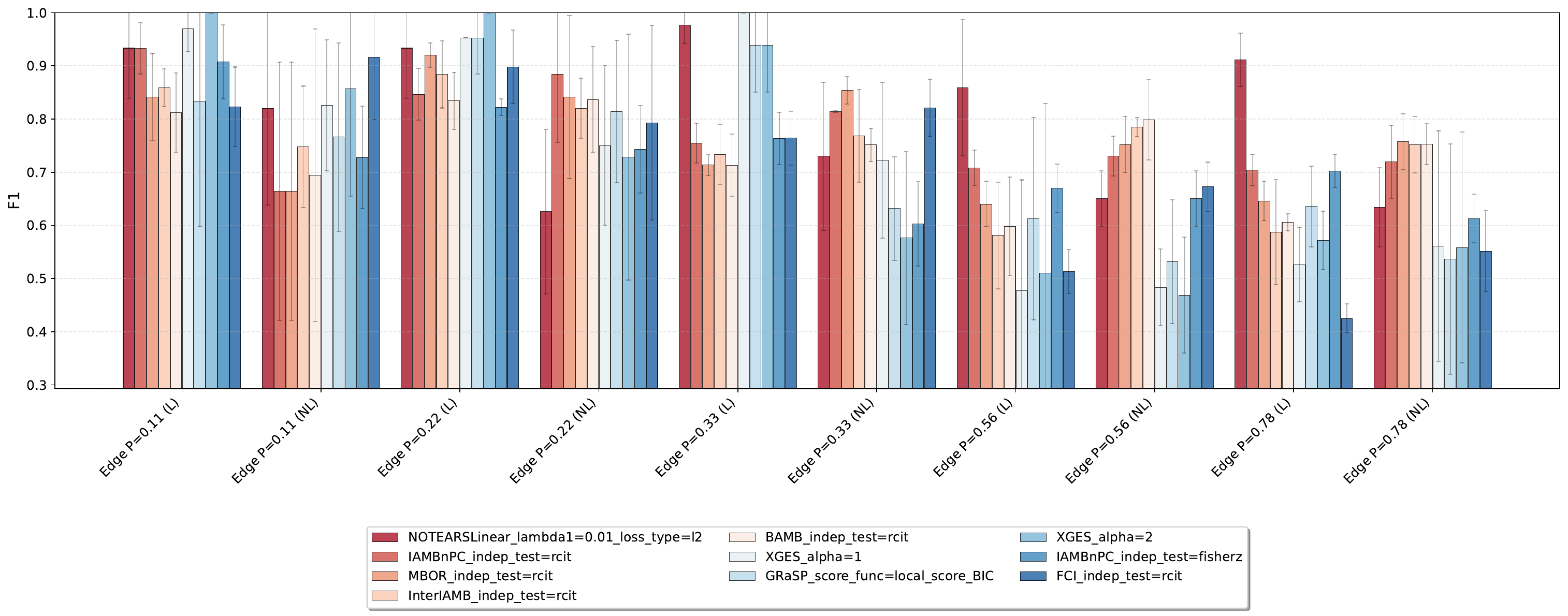}
  \caption{Performance vs. edge probability for tabular causal discovery algorithms. Only the top 10 algorithms with the highest average F1 scores are shown. 'L' and 'NL' indicate results on linear and non-linear data, respectively.}
  \label{fig:perf_vs_edge_prob}
\end{figure}

\begin{figure}[htbp]
  \centering
  \includegraphics[width=\textwidth]{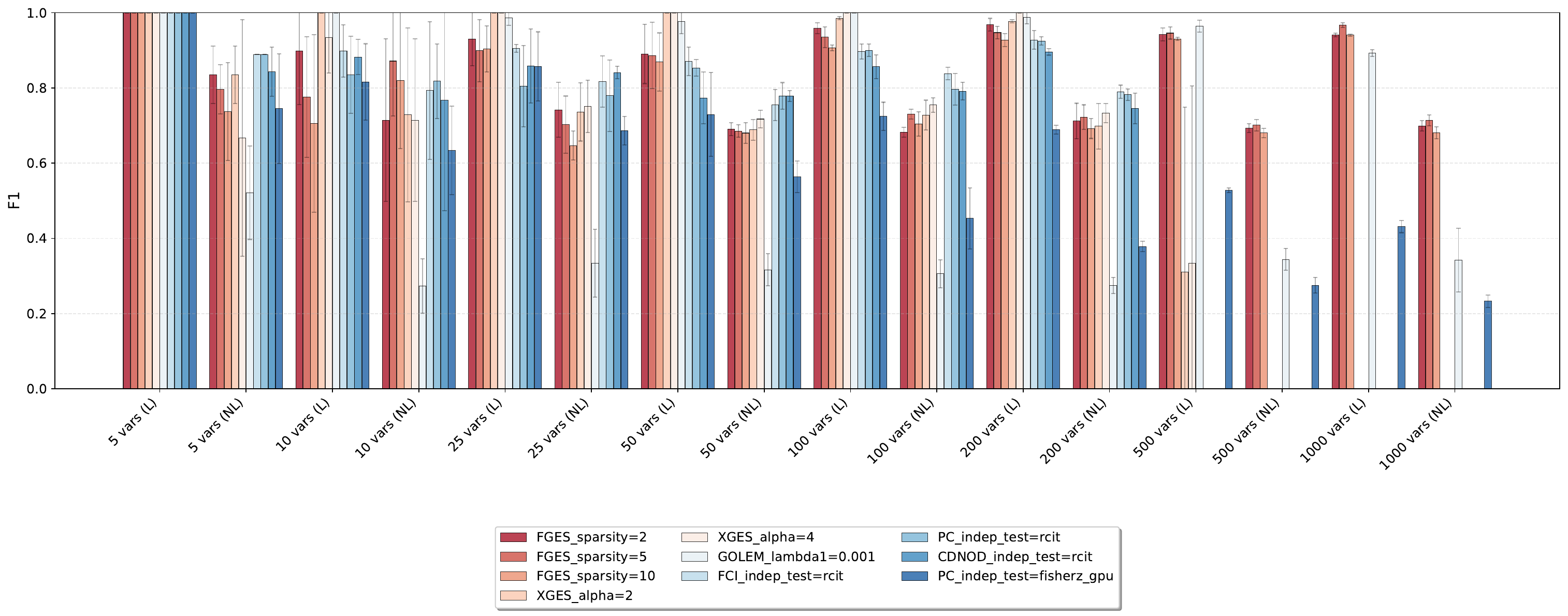}
  \caption{Performance vs. number of variables for tabular causal discovery algorithms. Only the top 10 algorithms with the highest average F1 scores are shown. 'L' and 'NL' indicate results on linear and non-linear data, respectively. An F1 score of 0 means the algorithm did not complete within the 20-minute time limit.}
  \label{fig:perf_vs_variables}
\end{figure}

\paragraph{Scalability to Variable Size.} 

Figure~\ref{fig:perf_vs_variables} presents the F1 performance of leading tabular causal discovery algorithms under different sizes of graphs. In linear relationships, FGES, XGES and GOLEM outperform constraint-based methods by incorporating data likelihood through their score functions and optimization objective. This advantage enables them to better capture the underlying causal structure in linear settings. Conversely, for nonlinear relationships, constraint-based methods like FCI and PC equipped with non-parametric conditional independence tests (particularly RCIT) demonstrate better performance.

\paragraph{Sensitivity to Graph Density.}

Figure~\ref{fig:perf_vs_edge_prob} illustrates how algorithm performance varies with graph density. For datasets with linear relationships, NOTEARS (Linear) demonstrates superior robustness to both sparse and dense graphs, which could be attributed to its soft sparsity regularization technique. MB-based methods such as IAMBnPC and MBOR with RCIT exhibit greater robustness to increasing edge probability, especially in non-linear scenarios. Their performance remains relatively stable across a wide range of densities, suggesting that local structure learning and robust independence tests can mitigate the challenges posed by dense connectivity. Score-based methods such as XES and GRaSP maintain high F1 scores in sparse to moderately dense linear graphs, but their performance declines as edge probability increases. This drop is more pronounced in non-linear settings, suggesting their sensitivity to both graph density and relationship type. 

\begin{figure}[htbp]
  \centering
  \includegraphics[width=\textwidth]{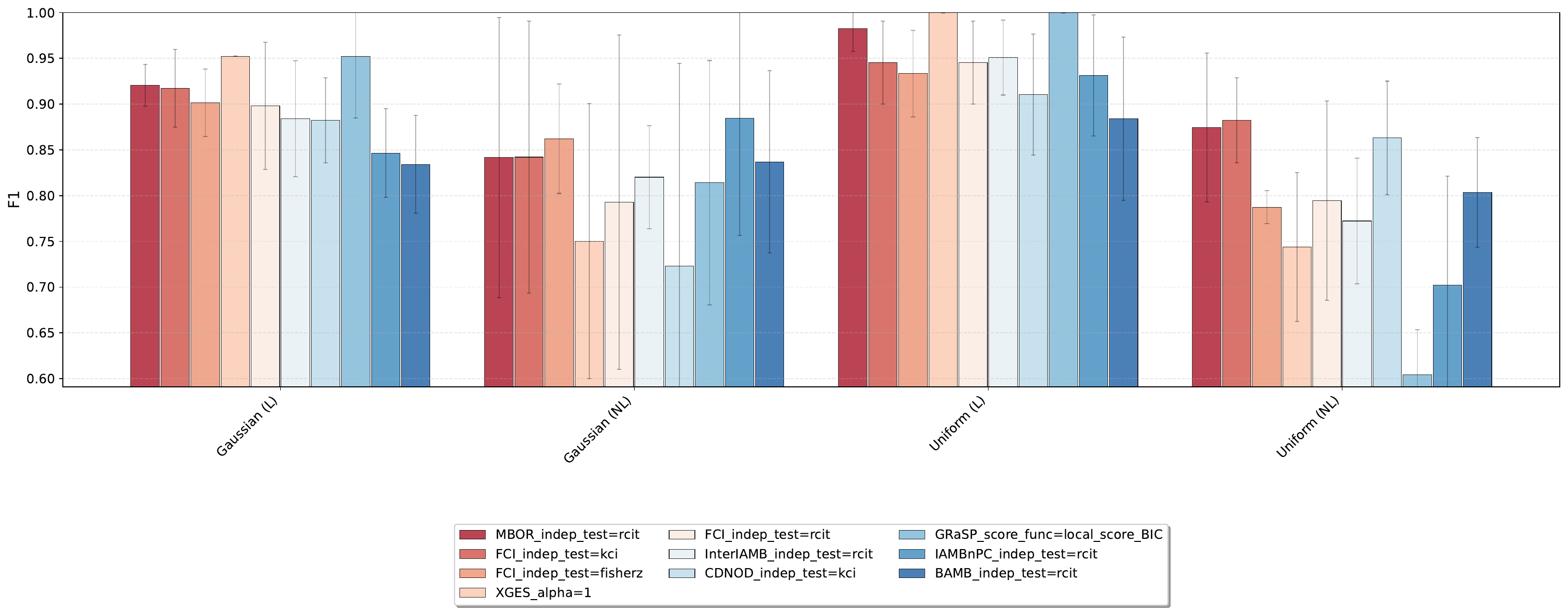}
  \caption{Performance across noise types for tabular causal discovery algorithms. Only the top 10 algorithms with the highest average F1 scores are shown. 'L' and 'NL' indicate results on linear and non-linear data, respectively.}
  \label{fig:perf_vs_noise}
\end{figure}

\paragraph{Sensitivity to Noise Type.} Our analysis of noise distribution effects reveals distinct algorithm behaviors as shown in Figure~\ref{fig:perf_vs_noise}. As a representative method of functional-causal-model based methods, DirectLiNGAM shows perfect identification in the linear non-Gaussian setting, but performance degrades significantly in the presence of Gaussian noise and non-linear settings, which makes it out of the general top performers. Interestingly, we see that although score-based methods like XGES and GRaSP implicitly assume Gaussian distribution in the common formulations of Bayesian Information Criterion (BIC), empirically, they are still robust to the uniform noise type data. Independence-test based methods (e.g., MBOR, FCI, CDNOD) show remarkable stability across both Gaussian and non-Gaussian noise distributions, demonstrating their flexibility and adaptability to different noise types.

\begin{figure}[htbp]
  \centering
  \includegraphics[width=\textwidth]{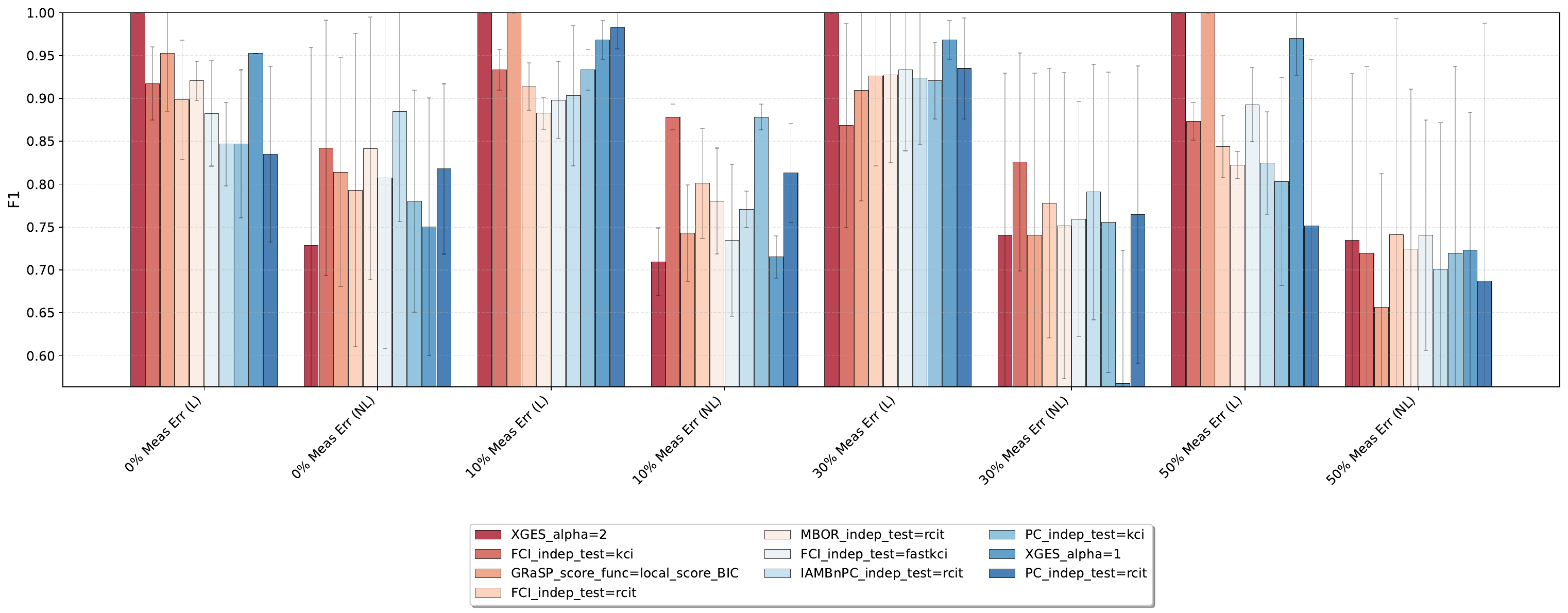}
  \caption{Performance vs. measurement error for tabular causal discovery algorithms. Only the top 10 algorithms with the highest average F1 scores are shown. 'L' and 'NL' indicate results on linear and non-linear data, respectively.}
  \label{fig:perf_vs_measurement_error}
\end{figure}

\paragraph{Robustness to Measurement error.}
Different algorithms exhibit varying degrees of resilience when confronted with common data challenges. When dealing with measurement error, as shown in Figure~\ref{fig:perf_vs_measurement_error}, XGES and GRaSP algorithms demonstrate exceptional robustness in linear settings, maintaining high accuracy even under significant noise. However, when the underlying variable interactions are non-linear, independence-test based methods struggle, exhibiting poor performance due to their limited ability to capture complex dependencies in the presence of measurement error. Similarly, score-based methods like XGES and GRaSP, while robust in linear cases, are fundamentally limited in non-linear scenarios unless equipped with non-parametric score functions—an area where current implementations fall short. This highlights a critical need for developing more scalable and flexible score functions capable of handling non-linear and noisy data.

\begin{figure}[htbp]
  \centering
  \includegraphics[width=\textwidth]{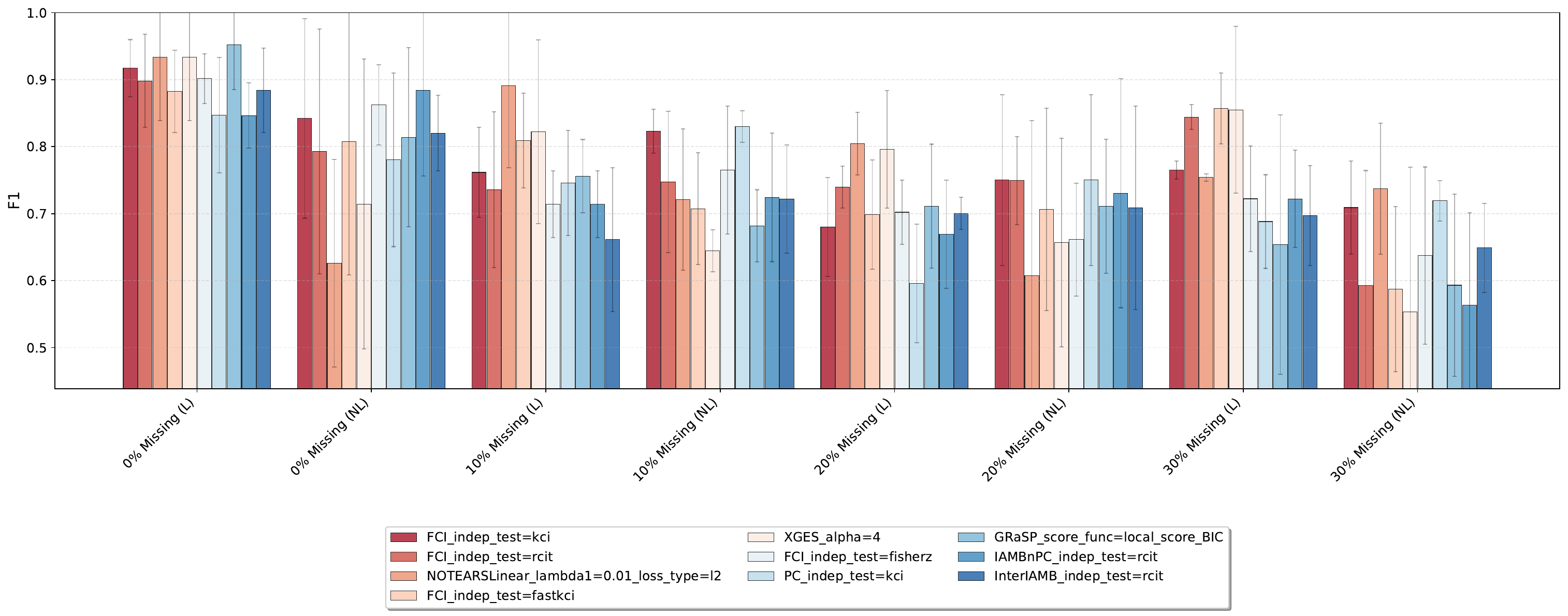}
  \caption{Performance vs. missing data rate for tabular causal discovery algorithms. Only the top 10 algorithms with the highest average F1 scores are shown. 'L' and 'NL' indicate results on linear and non-linear data, respectively.}
  \label{fig:perf_vs_missing}
\end{figure}

\paragraph{Robustness to Missing Data.} For missing data challenges illustrated in Figure~\ref{fig:perf_vs_missing}, constraint-based methods like PC with KCI, FCI with KCI and RCIT show stronger resilience than most alternatives, suggesting their underlying mechanisms better handle incomplete information. This could benefit from the flexibility of non-parametric conditional independence tests, as well as the ability to handle hidden confounders in FCI. 

\begin{figure}[htbp]
  \centering
  \includegraphics[width=\textwidth]{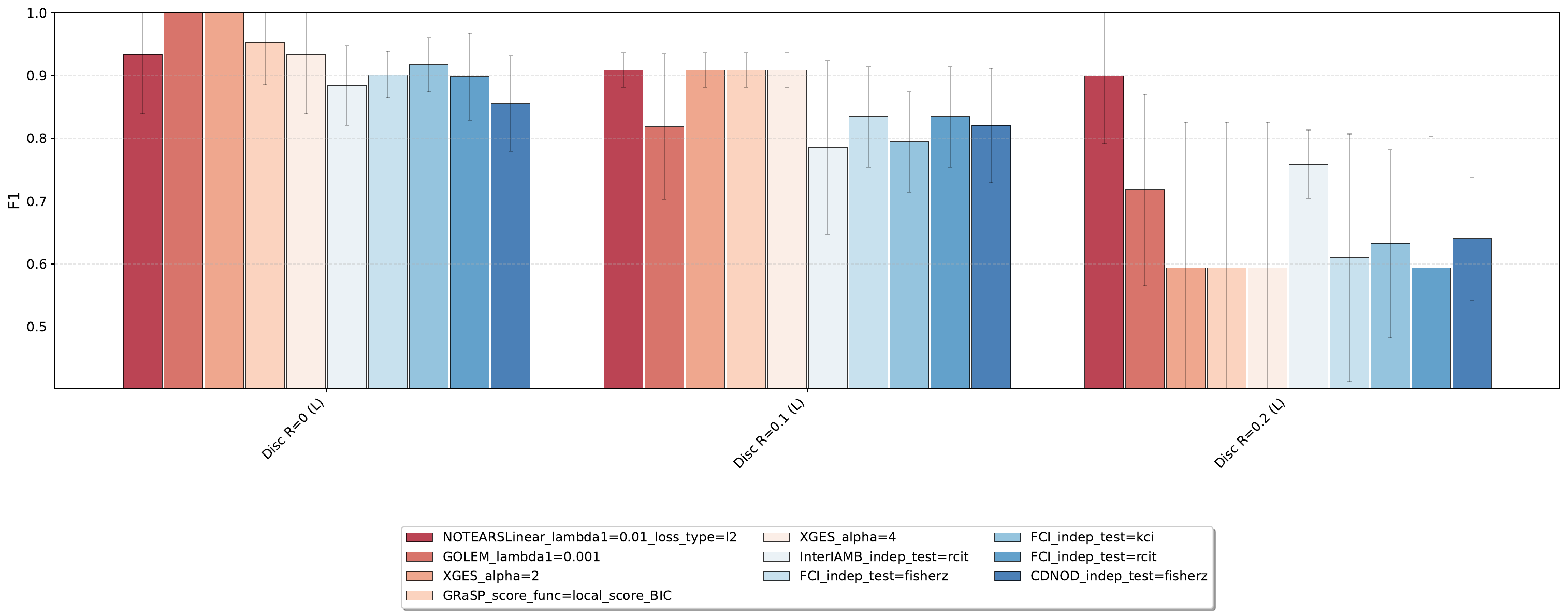}
  \caption{Performance vs. discrete variable ratio for tabular causal discovery algorithms. Only the top 10 algorithms with the highest average F1 scores are shown. 'L' indicates results on linear data.}
  \label{fig:perf_vs_discrete}
\end{figure}

\paragraph{Robustness to Mixed Data.} In mixed data scenarios where discrete and linear continuous variables co-exist, score-based methods and continuous-optimization-based methods like XGES and NOTEARS (Linear) consistently outperform alternatives in the low discrete data scenario, as shown in Figure \ref{fig:perf_vs_discrete}. However, as the discrete data ratio increases, the performance of score-based methods shows a pronounced drop, and only NOTEARS (Linear), GOLEM, and InterIAMB with RCIT maintain reasonable performance. NOTEARS (Linear) excels by optimizing for a global structure using a differentiable score function, while InterIAMB, as a Markov Blanket-based method, recovers local graphs independently, reducing error propagation in discrete settings.

\begin{figure}[htbp]
  \centering
  \includegraphics[width=\textwidth]{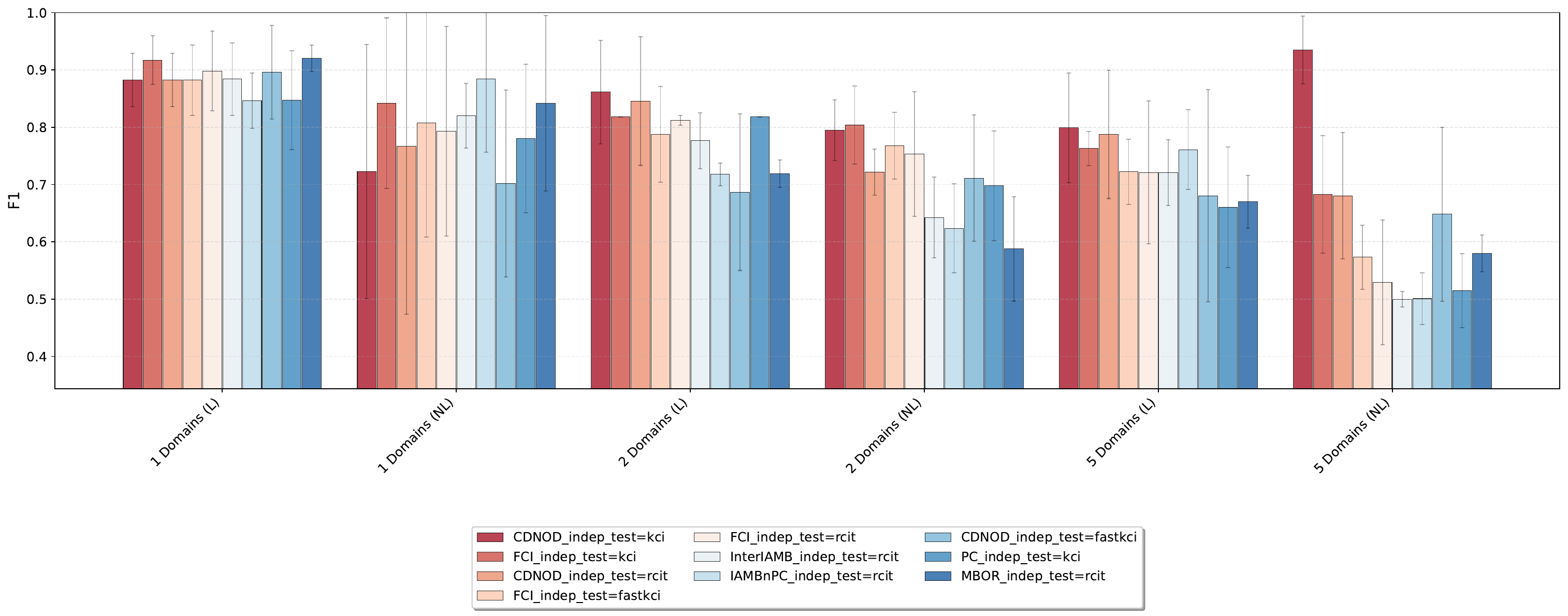}
  \caption{Performance across strength of heterogeneity for tabular causal discovery algorithms. Only the top 10 algorithms with the highest average F1 scores are shown. 'L' and 'NL' indicate results on linear and non-linear data, respectively.}
  \label{fig:perf_vs_domain}
\end{figure}

\paragraph{Robustness to Heterogeneous Data.} Figure~\ref{fig:perf_vs_domain} shows algorithm performance under heterogeneous data conditions. The results highlight the superior performance of CDNOD, which consistently outperforms other methods across varying levels of heterogeneity, especially in the nonlinear case with strong heterogeneity (5 domains). This exceptional resilience is due to CDNOD's ability to leverage the domain index as a surrogate to model the heterogeneity. In comparison, score-based methods and continuous-optimization-based methods suffer from the hidden confounding effects of domain changes.

\begin{figure}[htbp]
  \centering
  \includegraphics[width=\textwidth]{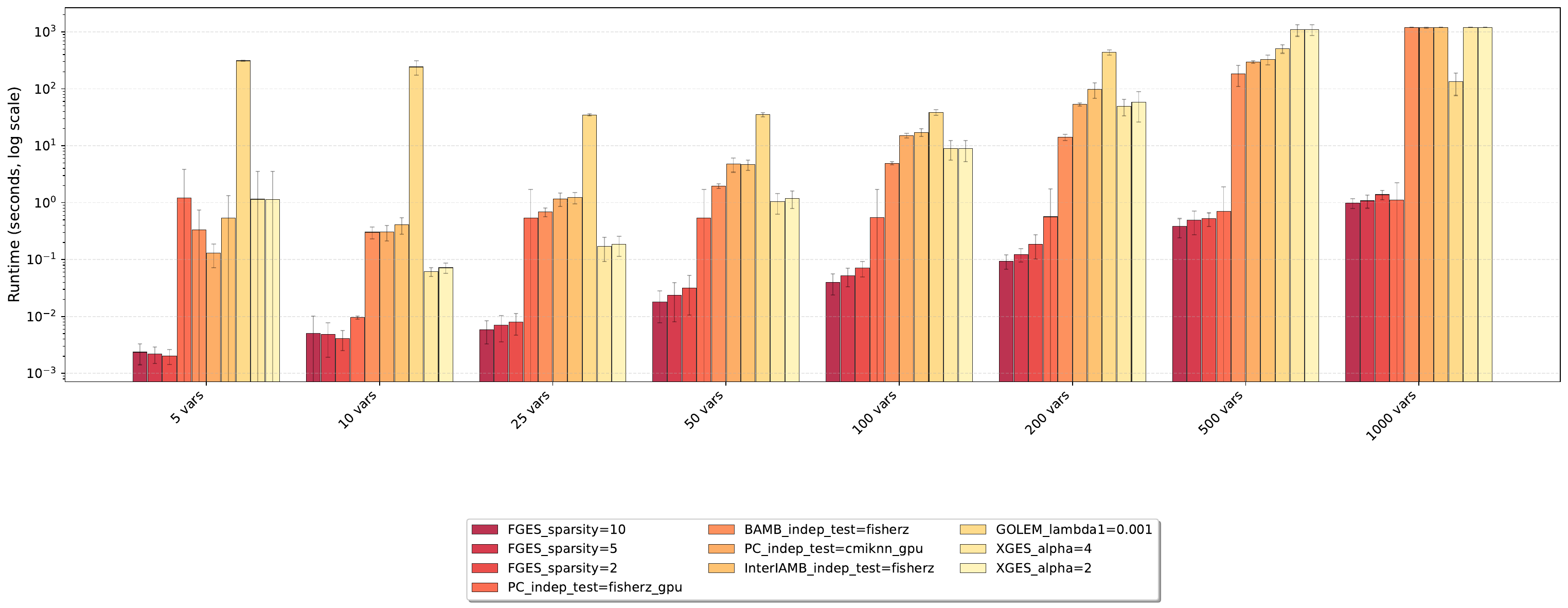}
  \caption{Runtime vs. number of variables for tabular causal discovery algorithms. Only the top 10 algorithms with the lowest runtime are shown. The runtime is plotted in the logarithmic scale, and the y-axis indicates the regular scale.}
  \label{fig:runtime_vs_variables}
\end{figure}

\begin{figure}[htbp]
  \centering
  \includegraphics[width=\textwidth]{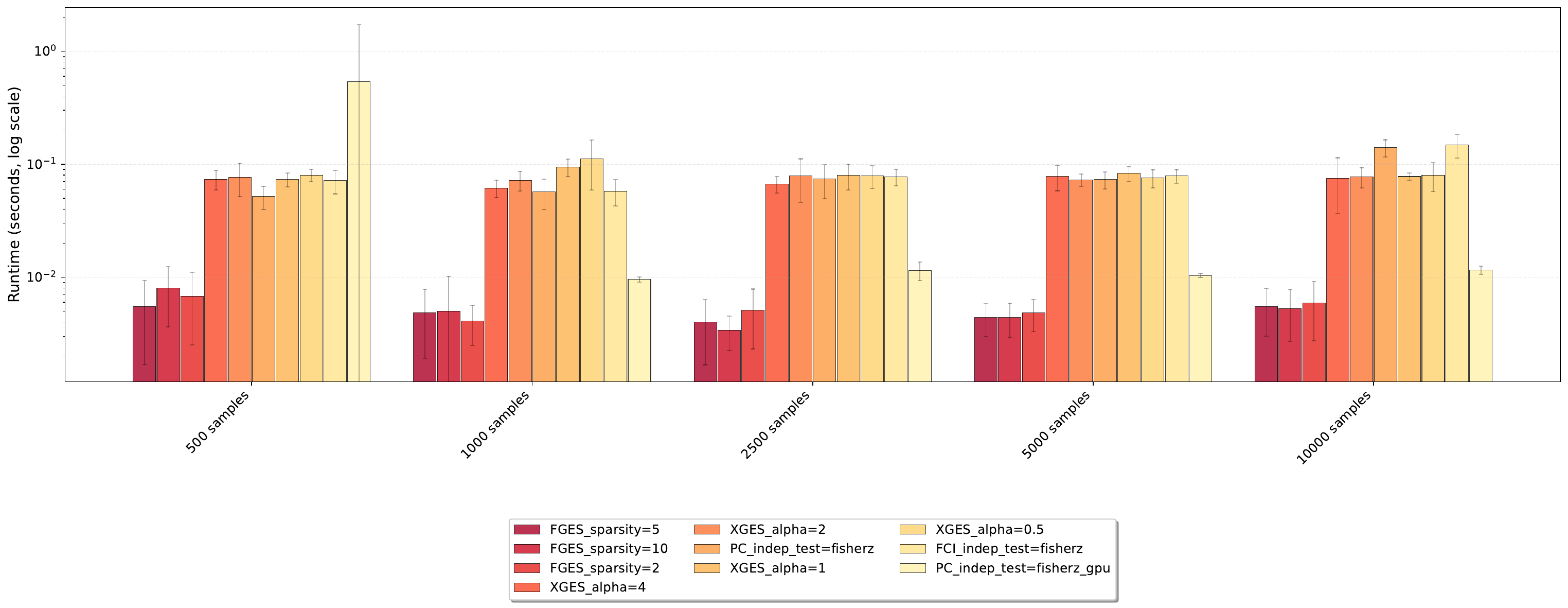}
  \caption{Runtime vs. sample size for tabular causal discovery algorithms. Only the top 10 algorithms with the lowest runtime are shown, and the y-axis indicates the regular scale.}
  \label{fig:runtime_vs_sample_size}
\end{figure}

\begin{figure}[htbp]
  \centering
  \includegraphics[width=\textwidth]{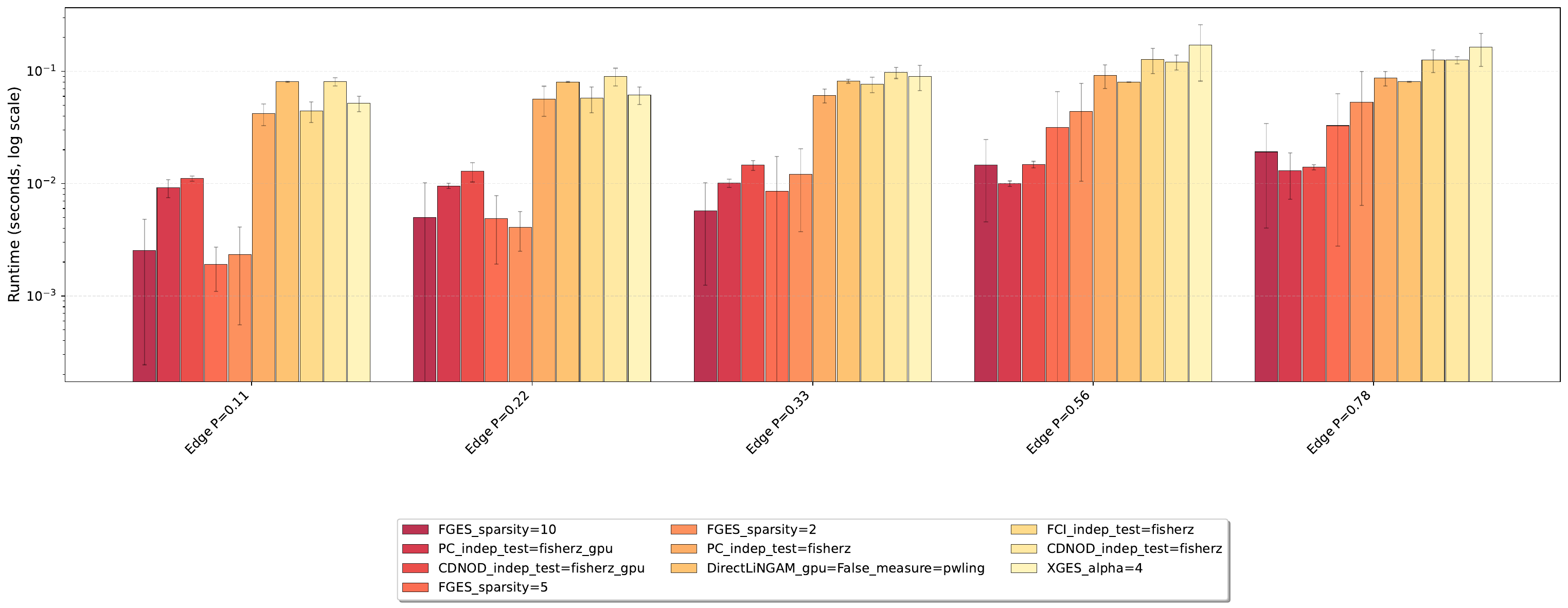}
  \caption{Runtime vs. edge probability for tabular causal discovery algorithms. Only the top 10 algorithms with the lowest runtime are shown, and the y-axis indicates the regular scale.}
  \label{fig:runtime_vs_edge_prob}
\end{figure}

\paragraph{Efficiency Considerations.}
Algorithm efficiency varies dramatically across causal discovery methods as demonstrated in Figures~\ref{fig:runtime_vs_variables}, \ref{fig:runtime_vs_sample_size}, and \ref{fig:runtime_vs_edge_prob}, with particularly notable differences emerging in large-scale applications. FGES demonstrates exceptional efficiency across variable scaling scenarios, making it suitable for large networks with many variables, especially when the network is linear and sparse. GPU-accelerated implementations, particularly for PC with Fisher-Z and CMIknn , show substantial efficiency improvements for large networks by leveraging parallel computation to handle both linear and non-linear data. The efficiency of continuous-based methods like GOLEM is mainly limited by the optimization iterations. Since these methods rely on optimization convergence rather than network size, they can achieve comparable runtime on larger graphs as on smaller ones when the functional complexity remains similar. Markov-blanket based methods like BAMB and InterIAMB offer another efficiency advantage by decomposing the global structure learning problem into local subproblems, allowing them to scale more gracefully with network size while maintaining accuracy in identifying causal relationships.

\paragraph{Algorithm Selection Recommendations.}
Based on our comprehensive benchmarking results, algorithm selection should be guided by specific data characteristics and scenario conditions. Firstly, consider the scale of the problem: for large-scale applications (variables > 100), computational efficiency becomes critical, making FGES and GPU-accelerated PC with Fisher-Z test ideal for linear data, while PC/FCI with RCIT or GPU-accelerated CMIknn  are recommended for non-linear data. For smaller-scale problems with clean linear relationships, continuous-optimization-based methods like NOTEARS (Linear) and GOLEM excel, particularly with dense graphs due to their global optimization approach. In linear non-Gaussian settings, DirectLiNGAM offers perfect identification capabilities. For non-linear relationships or mixed data types, constraint-based methods paired with kernel-based independence tests (PC, FCI, IAMBnPC, or InterIAMB with RCIT) demonstrate robust performance across diverse conditions. When facing heterogeneous data, CDNOD with either RCIT (non-linear) or Fisher-Z (linear) provides superior performance by leveraging domain indices to model heterogeneity. For scenarios with significant measurement errors or noisy observations, score-based algorithms like XGES and GRaSP show notable resilience. When dealing with missing data, FCI-based methods are particularly effective due to their inherent mechanisms for handling incomplete information. For dense and sparse graphs, continuous-optimization-based methods and score-based methods (e.g., NOTEARS (Linear) and XGES) with tuned sparsity parameters offer the best balance of accuracy and computational efficiency. For new practitioners without specific domain knowledge, we recommend starting with XGES and PC with KCI/RCIT for general-purpose linear and non-linear applications, FGES for efficiency with linear sparse data, and NOTEARS (Linear) or GOLEM for smaller datasets with unknown graph density, as these methods provide a good balance of performance, interpretability, and computational efficiency across a wide range of scenarios.

\subsubsection{Time-series data}

\begin{figure}[htbp]
  \centering
  \includegraphics[width=\textwidth, height=0.4\textwidth, keepaspectratio]{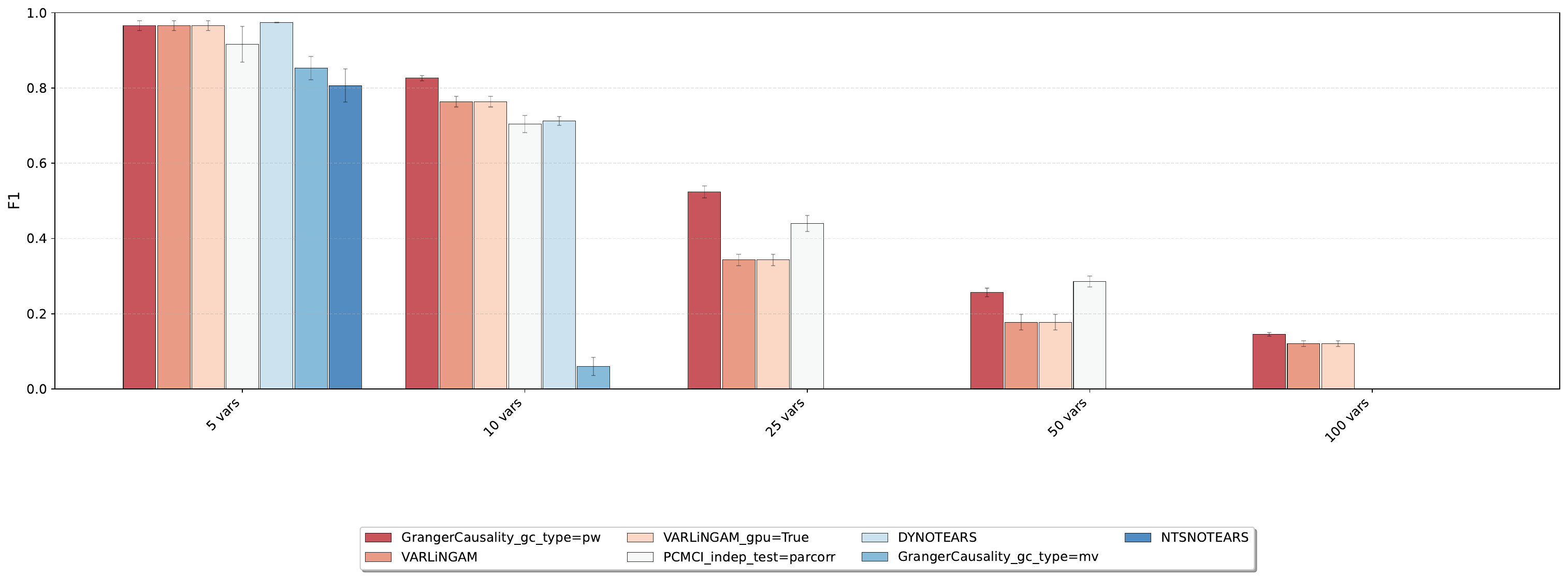}
  \caption{Performance vs. Number of variables for time-series causal discovery algorithms}
  \label{fig:perf_vs_vars_ts}
\end{figure}

\begin{figure}[htbp]
  \centering
  \includegraphics[width=\textwidth, height=0.4\textwidth, keepaspectratio]{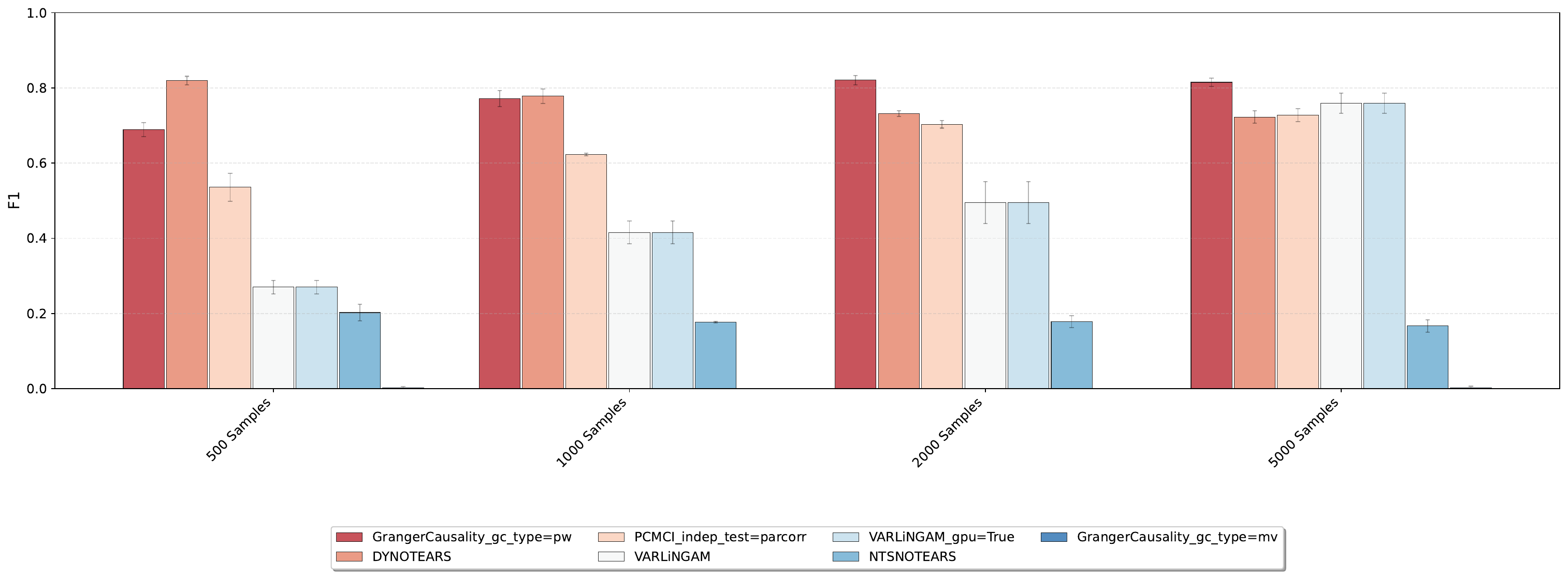}
  \caption{Performance vs. sample size for time-series causal discovery algorithms}
  \label{fig:perf_vs_samples_ts}
\end{figure}

\begin{figure}[htbp]
  \centering
  \includegraphics[width=\textwidth, height=0.4\textwidth, keepaspectratio]{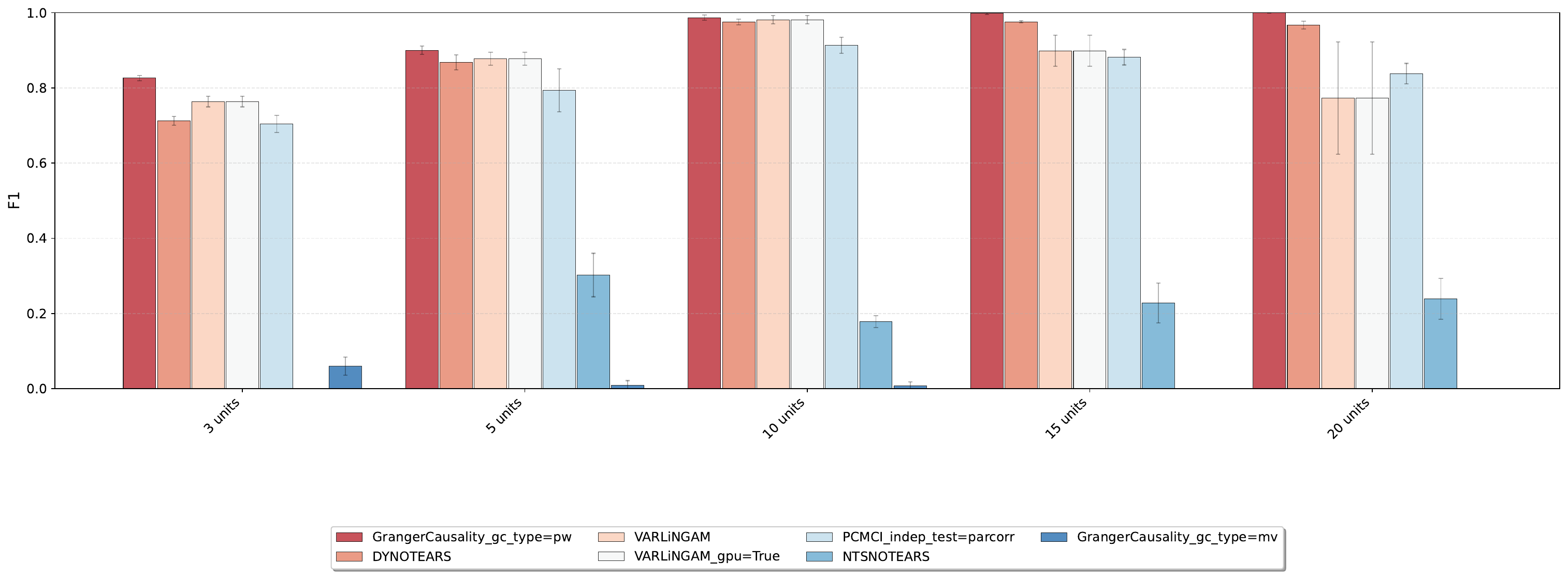}
  \caption{Performance vs. time lag for time-series causal discovery algorithms}
  \label{fig:perf_vs_lag_ts}
\end{figure}

\begin{figure}[htbp]
  \centering
  \includegraphics[width=\textwidth, height=0.4\textwidth, keepaspectratio]{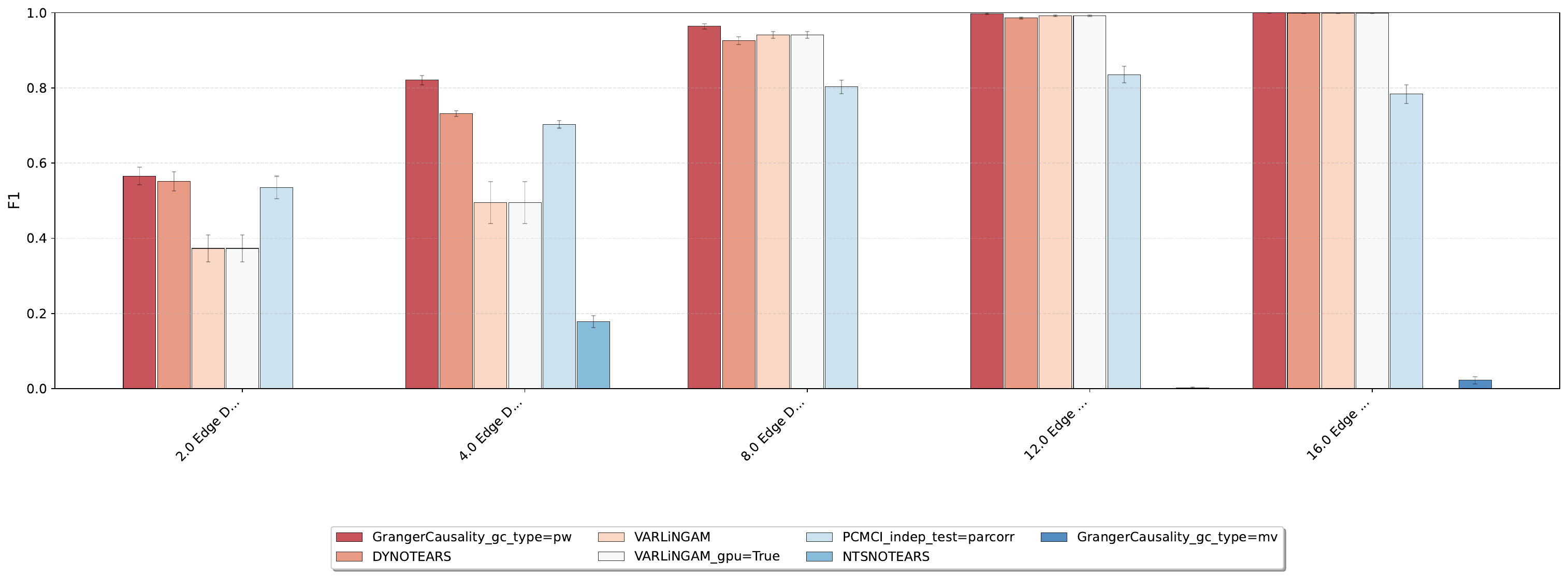}
  \caption{Performance vs. edge density for time-series causal discovery algorithms}
  \label{fig:perf_vs_edge_ts}
\end{figure}

\paragraph{Overall Performance Trends}
For time-series datasets, we observe that pairwise Granger causality consistently achieves strong performance across a variety of experimental settings.
As shown in Figure~\ref{fig:perf_vs_vars_ts}, all algorithms—except VARLiNGAM—fail to complete execution within the imposed time limit, highlighting VARLiNGAM’s exceptional scalability for very large datasets. DYNOTEARS, due to its formulation as a constrained optimization framework, demonstrates notable robustness to varying sample sizes, as illustrated in Figure~\ref{fig:perf_vs_samples_ts}. For the remaining algorithms, performance generally improves with increasing sample size, reflecting their reliance on statistical estimation techniques. Figure~\ref{fig:perf_vs_lag_ts} reveals the impact of time lag on model performance. Despite the increased graph complexity associated with higher lag values, both DYNOTEARS and VARLiNGAM exhibit improved performance, suggesting that higher lag estimates may be beneficial for capturing temporal dependencies more accurately.  Consistent with trends observed in the tabular data experiments, the constraint-based structure of PCMCI renders it sensitive to changes in graph sparsity. As shown in Figure~\ref{fig:perf_vs_edge_ts}, DYNOTEARS and VARLiNGAM show greater resilience to variations in edge density.

\begin{figure}[htbp]
  \centering
  \includegraphics[width=\textwidth, height=0.4\textwidth, keepaspectratio]{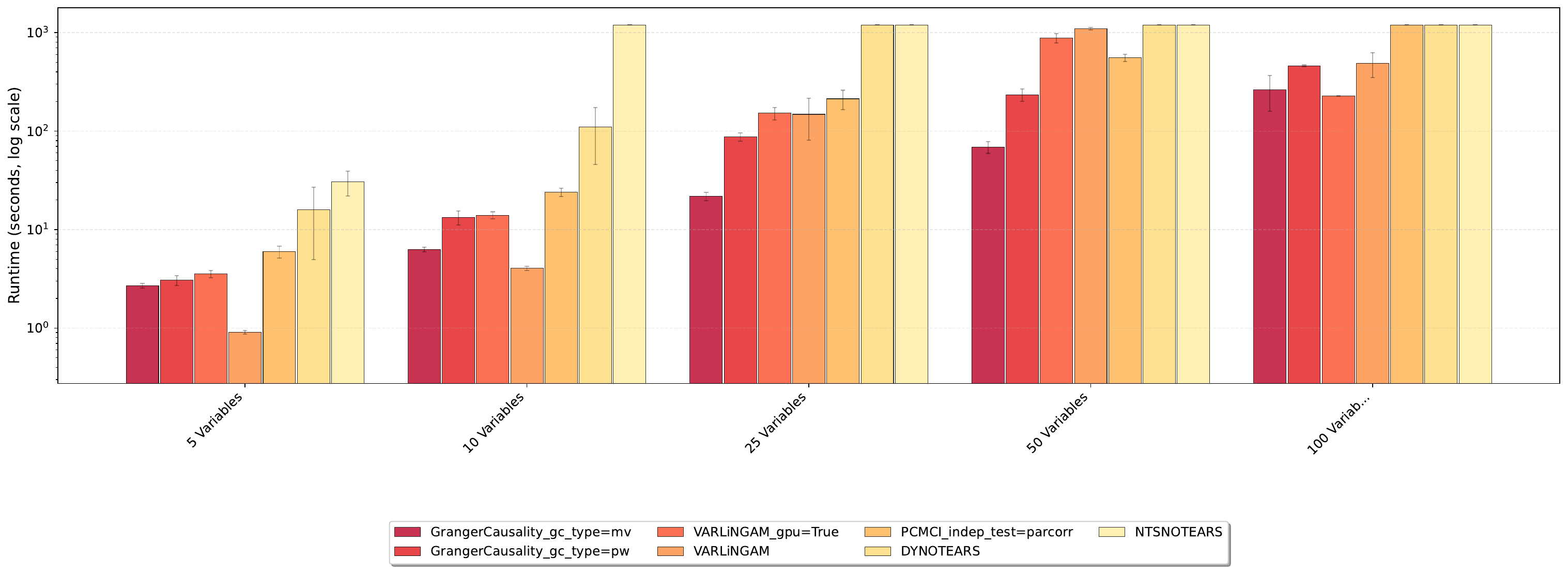}
  \caption{Runtime vs. Number of variables for time-series causal discovery algorithms. The runtime is plotted in the logarithmic scale, and the y-axis indicates the regular scale.}
  \label{fig:runtime_vs_vars_ts}
\end{figure}

\begin{figure}[htbp]
  \centering
  \includegraphics[width=\textwidth, height=0.4\textwidth, keepaspectratio]{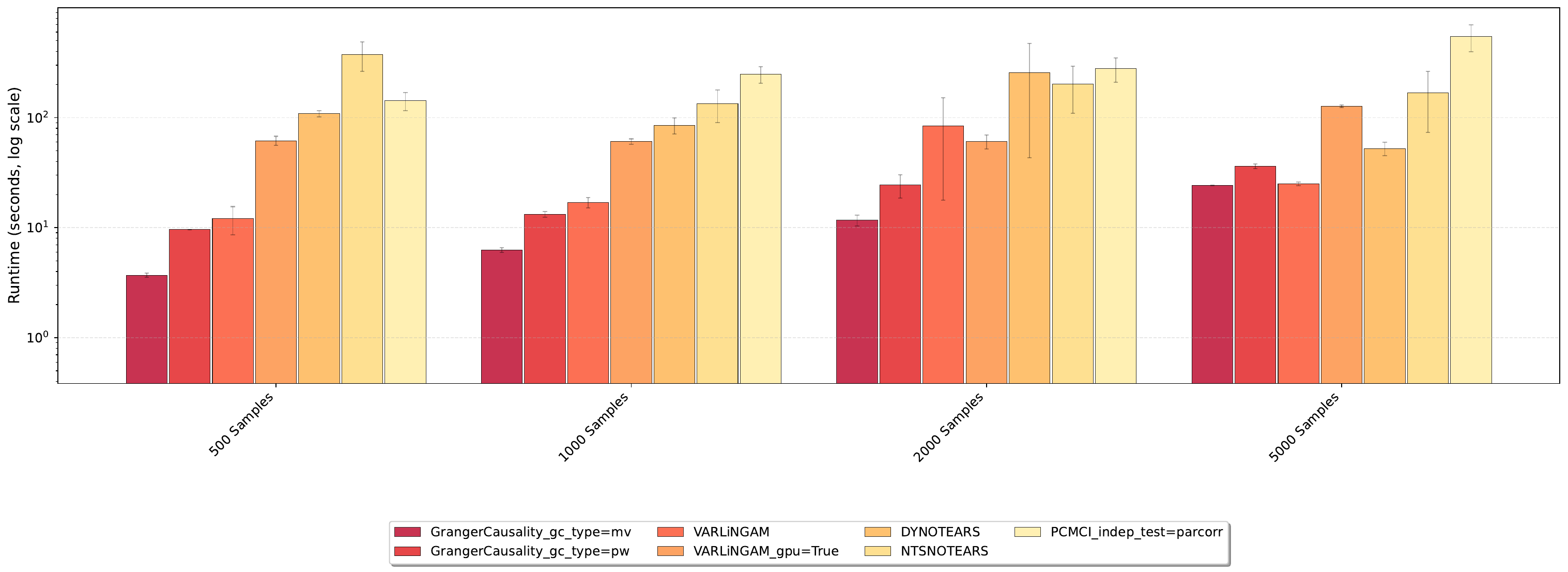}
  \caption{Runtime vs. sample size for time-series causal discovery algorithms. The runtime is plotted in the logarithmic scale, and the y-axis indicates the regular scale.}
  \label{fig:runtime_vs_samples_ts}
\end{figure}

\begin{figure}[htbp]
  \centering
  \includegraphics[width=\textwidth, height=0.4\textwidth, keepaspectratio]{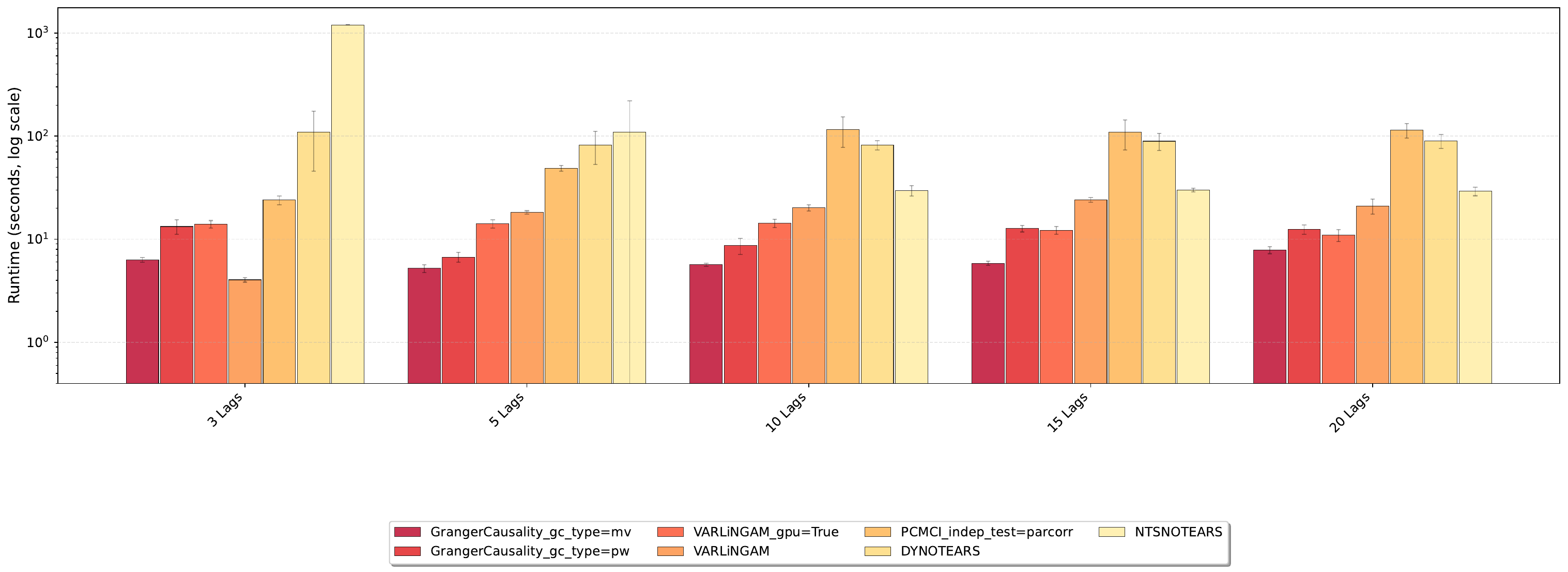}
  \caption{Runtime vs. time lag for time-series causal discovery algorithms. The runtime is plotted in the logarithmic scale, and the y-axis indicates the regular scale.}
  \label{fig:runtime_vs_lag_ts}
\end{figure}

\begin{figure}[htbp]
  \centering
  \includegraphics[width=\textwidth, height=0.4\textwidth, keepaspectratio]{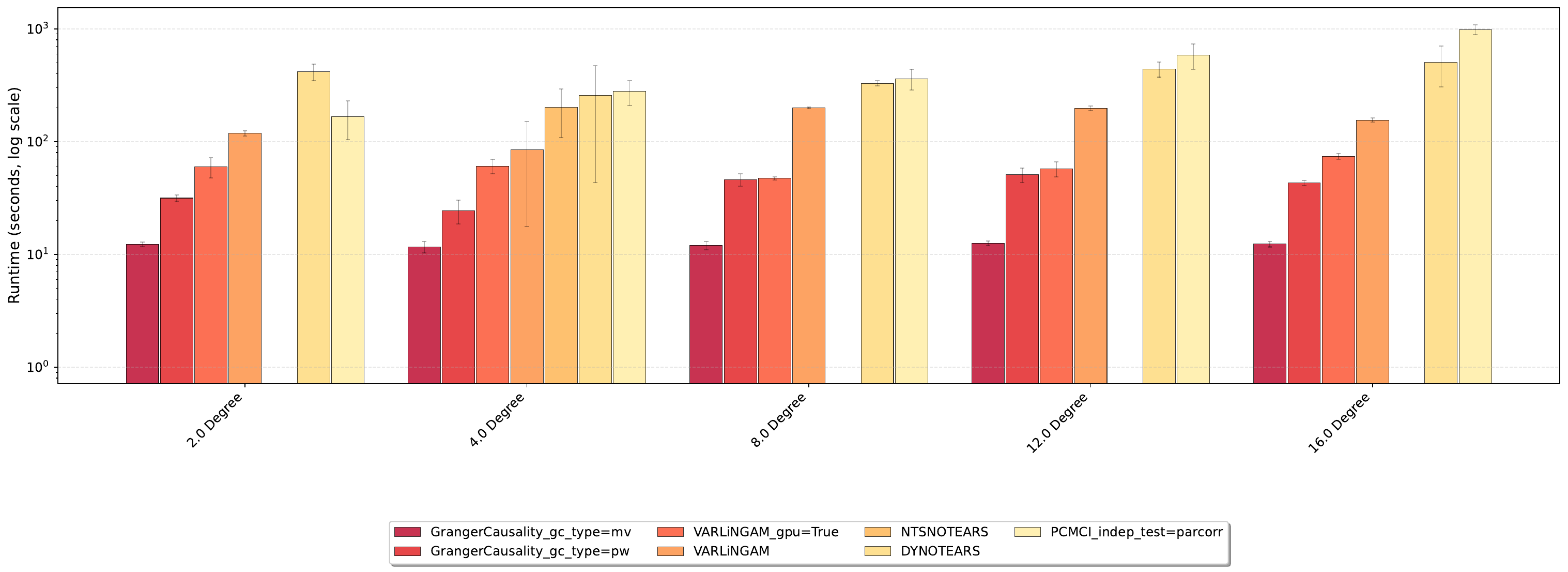}
  \caption{Runtime vs. lagged edge density for time-series causal discovery algorithms. The runtime is plotted in the logarithmic scale, and the y-axis indicates the regular scale.}
  \label{fig:runtime_vs_edge_ts}
\end{figure}

\paragraph{Efficiency Considerations}
As illustrated in Figures ~\ref{fig:runtime_vs_vars_ts}, \ref{fig:runtime_vs_samples_ts}, \ref{fig:runtime_vs_lag_ts} and \ref{fig:runtime_vs_edge_ts}  we observe an exponential increase in runtime with growing system complexity. The GPU-accelerated implementation of VARLiNGAM offers a substantial improvement in execution speed over its traditional counterpart, making it a practical choice for datasets with a large number of variables. PCMCI, as a constraint-based method, exhibits the most pronounced sensitivity to increasing sample size and increasing edge density, with runtime escalating in proportion to the complexity of the independence tests. Furthermore, we find that the number of variables has a more significant impact on runtime than the length of the time lag. Among the evaluated methods, DYNOTEARS and NTSNOTEARS demonstrate notable robustness to variations in edge density within the lagged graph, making them more suitable for complex temporal structures.

\paragraph{Algorithm Selection Recommendations}
For time-series data exhibiting predominantly linear causal relations, DYNOTEARS emerges as a strong candidate, offering both high performance and computational efficiency, which collectively make it a favorable choice over alternative methods.  In our experiments, we primarily employed partial correlation as the independence test for PCMCI. While alternative methods—such as robust partial correlation and Gaussian process regression—yield marginal performance improvements in smaller-scale settings, they consistently fail to meet the imposed time constraints. As a result, these methods may be more appropriate for applications involving smaller datasets where accuracy is prioritized over computational efficiency. For datasets exhibiting non-Gaussian noise, VARLiNGAM is the preferred method due to its ability to exploit non-Gaussianity for causal identification. Moreover, irrespective of the underlying noise distribution, the GPU-accelerated version of VARLiNGAM remains a robust and scalable option, particularly well-suited for datasets with a large number of variables. Although pairwise Granger causality demonstrates high accuracy and computational efficiency, it is important to acknowledge its fundamental drawbacks, which limit its applicability in real-world scenarios. While it can serve as a useful preliminary tool for gaining initial insights into the data structure, we recommend employing more advanced methods—such as those discussed above—for a more robust and interpretable causal analysis.
Ultimately, practitioners should carefully consider their specific data characteristics, computational resources, and domain requirements when selecting the most suitable causal discovery algorithm.



\subsection{Effectiveness of Automatic Causal Discovery}
To comprehensively evaluate the effectiveness of Causal-Copilot's automated pipeline, we designed a multi-category experimental framework, that tests its ability to handle diverse causal discovery scenarios. Our experimental design is motivated by the need to assess how well the system performs across the full spectrum of challenges encountered in real-world causal analysis - from basic scenarios to complex data quality issues and domain-specific applications. We conduct a series of experiments using synthetic datasets with known ground-truth causal structures. For tabular data, we generate data based on key parameters including the number of variables, graph density, relationship types (linear and non-linear), and data quality challenges as shown in Table~\ref{tab:exp_tab}. For time series data, we generate multivariate time series with parameters controlling dimensionality, sample size, lag structure, and noise type. Additionally, we created compound scenarios that combine multiple challenging characteristics simultaneously to simulate complex real-world cases - such as the 'Clinical Data Scenario', 'Financial Data Scenario', and 'IoT Sensor Network' (see Appendix \ref{app:pre_exp_res} for setup details).

\begin{table*}[ht]
  \centering
  \caption{Comprehensive F1 Score Comparison Across All Scenarios (Mean ± Std). $\ddagger$ indicates settings include both linear case and non-linear case, while $\dagger$ indicates settings with purely linear relationships. N/A denotes algorithms that failed to complete due to computational constraints.}
  \label{tab:exp_tab}
  \resizebox{\textwidth}{!}{
  \begin{tabular}{lllcccccc}
  \toprule
  \textbf{Category} & \textbf{Subcategory} & \textbf{Setting} & \textbf{Causal-Copilot} & \textbf{GPT-4o} & \textbf{PC} & \textbf{FCI} & \textbf{GES} & \textbf{DirectLiNGAM} \\
  \midrule
  \multirow{11}{*}{Basic Scenarios} & Default Settings & Normal (p=10, n=1000) $\ddagger$ & 0.900 ± 0.120 & 0.850 ± 0.160 & \textbf{0.920 ± 0.050} & 0.910 ± 0.060 & 0.920 ± 0.090 & 0.220 ± 0.220 \\
  \cmidrule(lr){2-9}
   & \multirow{2}{*}{Graph Density} & Dense (p=0.5) $\ddagger$ & \textbf{0.780 ± 0.170} & 0.450 ± 0.120 & 0.410 ± 0.110 & 0.450 ± 0.110 & 0.400 ± 0.150 & 0.260 ± 0.120 \\
   &   & Sparse (p=0.1) $\ddagger$ & 0.820 ± 0.260 & 0.830 ± 0.280 & 0.810 ± 0.270 & \textbf{0.840 ± 0.240} & 0.790 ± 0.270 & 0.140 ± 0.070 \\
  \cmidrule(lr){2-9}
   & \multirow{3}{*}{Node Count} & Extreme Large (p=500) $\ddagger$ & \textbf{0.860 ± 0.130} & N/A & N/A & N/A & N/A & N/A \\
   &   & Super Large (p=100) $\ddagger$ & \textbf{0.910 ± 0.080} & N/A & 0.680 ± 0.170 & 0.740 ± 0.120 & N/A & 0.240 ± 0.110 \\
   &   & Large (p=50) $\ddagger$ & \textbf{0.950 ± 0.060} & 0.790 ± 0.180 & 0.700 ± 0.140 & 0.790 ± 0.120 & 0.560 ± 0.460 & 0.230 ± 0.110 \\
  \cmidrule(lr){2-9}
   & \multirow{2}{*}{Sample Size} & Extra Large (n=10000) $\ddagger$ & \textbf{0.970 ± 0.050} & 0.760 ± 0.230 & 0.810 ± 0.180 & 0.820 ± 0.100 & 0.870 ± 0.220 & 0.210 ± 0.100 \\
   &   & Large (n=5000) $\ddagger$ & \textbf{0.950 ± 0.070} & 0.770 ± 0.270 & 0.800 ± 0.150 & 0.830 ± 0.120 & 0.800 ± 0.240 & 0.320 ± 0.160 \\
  \cmidrule(lr){2-9}
   & Large Scale & Extreme Large Node and Sample (p=1000, n=10000) $\ddagger$ & \textbf{0.870 ± 0.140} & N/A & N/A & N/A & N/A & N/A \\
  \cmidrule(lr){2-9}
   & Noise Type & Non-Gaussian $\ddagger$ & \textbf{0.980 ± 0.040} & 0.820 ± 0.190 & 0.840 ± 0.170 & 0.850 ± 0.200 & 0.860 ± 0.270 & 0.570 ± 0.470 \\
  \cmidrule(lr){2-9}
   & Mixed Data Types & Discrete (ratio=0.2) $\dagger$ & 0.890 ± 0.140 & N/A & 0.820 ± 0.190 & 0.830 ± 0.110 & \textbf{0.920 ± 0.090} & 0.380 ± 0.040 \\
  \midrule
  \multirow{3}{*}{Data Quality Challenges} & \multirow{3}{*}{Data Quality} & Heterogeneous Domains $\ddagger$ & \textbf{0.780 ± 0.100} & 0.690 ± 0.090 & 0.510 ± 0.210 & 0.620 ± 0.190 & 0.400 ± 0.250 & 0.230 ± 0.090 \\
   &   & Measurement Error $\ddagger$ & \textbf{0.890 ± 0.160} & 0.750 ± 0.300 & 0.690 ± 0.310 & 0.800 ± 0.190 & 0.790 ± 0.250 & 0.280 ± 0.130 \\
   &   & Missing Data $\ddagger$ & \textbf{0.770 ± 0.170} & 0.690 ± 0.210 & 0.640 ± 0.160 & 0.720 ± 0.160 & 0.720 ± 0.140 & 0.450 ± 0.100 \\
  \midrule
  \multirow{3}{*}{Compound Scenarios} & \multirow{3}{*}{Simulated Real-world Scenarios} & Clinical Data Scenario $\dagger$ & \textbf{0.690 ± 0.080} & 0.040 ± 0.040 & 0.530 ± 0.070 & 0.610 ± 0.040 & 0.490 ± 0.120 & 0.220 ± 0.100 \\
   &   & Financial Data Scenario $\ddagger$ & \textbf{0.650 ± 0.130} & N/A & 0.260 ± 0.030 & 0.300 ± 0.030 & N/A & 0.180 ± 0.030 \\
   &   & Social Network Scenario $\ddagger$ & \textbf{0.450 ± 0.080} & N/A & N/A & N/A & N/A & N/A \\
  \bottomrule
  \end{tabular}
  }
\end{table*}
We evaluate Causal-Copilot against both traditional causal discovery algorithms (PC, FCI, GES, and DirectLiNGAM) and a non-contextualized LLM baseline (GPT-4o). The GPT-4o baseline receives the same user queries as Causal-Copilot and is expected to select appropriate algorithms and hyperparameters from the same range, but lacks preliminary statistical diagnosis capabilities and causal expert knowledge. As shown in Table~\ref{tab:exp_tab}, Causal-Copilot demonstrates consistent performance advantages across multiple experimental conditions.

In basic scenarios with standard settings, all methods except DirectLiNGAM perform comparably well. However, as complexity increases, Causal-Copilot's advantages become apparent. It significantly outperforms other methods in dense graphs and extremely large networks (up to 1000 nodes) where most traditional algorithms and setups chosen by GPT-4o fail to complete due to computational constraints. For data quality challenges, Causal-Copilot shows remarkable robustness to heterogeneous domains, measurement errors, and missing data, consistently achieving higher F1 scores than both GPT-4o and traditional algorithms. GPT-4o predominantly selects FCI and PC algorithms with KCI, lacking the ability to adapt to specific application scenarios. Unlike Causal-Copilot, GPT-4o cannot leverage statistical preprocessing capabilities, domain-specific knowledge, or contextual understanding of the data's unique characteristics. Most notably, in compound scenarios that simulate real-world applications, Causal-Copilot maintains superior performance across all settings. In the Clinical Data Scenario (combining discrete variables, measurement errors, missing values, and multi-domain effects across 25 nodes), Causal-Copilot substantially outperforms GPT-4o, which struggles significantly with this complex setting. The Financial Data Scenario (50 nodes with sparse connections across 3 domains) presents challenges with its scale and complexity, where Causal-Copilot succeeds while GPT-4o fails to complete. Similarly, in the Social Network Scenario with 1000 nodes and relatively dense connection patterns (average degree 6), only Causal-Copilot produces meaningful results. These results demonstrate that Causal-Copilot's algorithm and hyperparameter selection strategy, effectively adapts to each tabular scenario's unique characteristics.

For time-series data, as shown in Table \ref{tab:eval_ts}, we compare the performance of Causal-Copilot with state-of-the-art algorithms such as PCMCI, DYNOTEARS, VARLiNGAM, NTSNOTEARS and a non-contextualized LLM baseline (GPT-4o). We can observe that for a standard setting, (variables =20, time lag =5),  Causal-Copilot performs competitively with PCMCI (F1=0.673 vs F1=0.695), trailing behind DYNOTEARS (F1=0.733). In high dimensional data with very large variables (p=50 and p=100), Causal-Copilot shows superior scalability (F1=0.182 for p=1000 vs next best 0.121), outperforming all other methods. This is especially significant as it reinforces the efficacy of our algorithm selection process. Causal-Copilot can adapt to both short and long time-lag structures, achieving the highest scores (F1=0.850 for l=20 vs. next best 0.838) in both settings. Although VARLiNGAM performs well with very large samples (F1=0.759 for n=5000), Causal-Copilot remains competitive (F1=0.649) and shows strong generalization across both large and moderately sized datasets. Causal-Copilot shows robustness to the different types of noise in the data, performing significantly better than the other approaches (F1=0.828 vs F1=0.714 for non-Gaussian noise) under both non-Gaussian and Gaussian noise distributions. Similar to the tabular data, GPT-4o performs reasonably well on time-series data. It essentially chooses an algorithm without any prior statistical information of the dataset, thus resulting in an almost comparable performance to most of the algorithms. Notably, GPT-4o predominantly favors PCMCI as the default algorithm selection choice, making it less useful in specialized scenarios such as non-Gaussian noise or datasets with high complexity. We can observe that it still fails for an extremely high dimensional dataset. Causal-Copilot offers strong generalization, adaptive hyperparameter tuning, advanced statistical analysis, and model selection capabilities, allowing it to succeed in a broad range of scenarios without manual intervention. It especially shines in large-scale, high-lag, and noisy environments, where traditional methods often falter. These results validate the effectiveness of Causal-Copilot as a flexible and intelligent assistant for causal discovery.

\begin{table*}[htbp]
  \centering
  \caption{Comprehensive F1 Score Comparison across all scenarios for time series algorithms (Mean ± Std). The data has linear causal relations and is stationary. N/A denotes algorithms that failed to complete execution due to computational constraints. }
  \label{tab:eval_ts}
  \resizebox{\textwidth}{!}{
  \begin{tabular}{lllcccccc}
  \toprule
  \textbf{Category} & \textbf{Subcategory} & \textbf{Setting} & \textbf{Causal-Copilot} & \textbf{GPT-4o} & \textbf{PCMCI} & \textbf{DYNOTEARS} & \textbf{VARLiNGAM} & \textbf{NTSNOTEARS} \\
  \midrule
  \multirow{11}{*}{Basic Scenarios} & Default Settings & Normal (p=20, l=5) & 0.673 ± 0.018 & 0.655 ± 0.031 & 0.695 ± 0.017 & \textbf{0.733 ± 0.007} & 0.498 ± 0.052  & 0.173 ± 0.018 \\
  \cmidrule(lr){2-9}
  & \multirow{3}{*}{Node Count} & Very Large (p=100, l=3) & \textbf{0.182 ± 0.004} & N/A& N/A & N/A & 0.121 ± 0.007 & N/A \\
  & & Large (p=50, l=3) & 0.284 ± 0.012 & 0.223 ± 0.006 & 0.286 ± 0.015  & N/A & 0.177 ± 0.021 & N/A \\
  & & Small (p=5, l=3) & \textbf{0.978 ± 0.003} & 0.917 ± 0.023 & 0.916 ± 0.047& 0.974 ± 0.001 &  0.965 ± 0.013& 0.807 ± 0.044 \\
  \cmidrule(lr){2-9}
  & \multirow{2}{*}{Time Lag} & Large (l=20)& \textbf{0.850 ± 0.031} & 0.738 ± 0.018 &0.838 ± 0.027 & 0.767 ± 0.010 & 0.773 ± 0.149 & 0.239 ± 0.054 \\
  & & Small (l=3) & \textbf{0.869 ± 0.056} & 0.638 ± 0.011 & 0.704 ± 0.023 & 0.713 ± 0.012 & 0.763 ± 0.014 & 0.461 ± 0.023\\
  \cmidrule(lr){2-9}
  & \multirow{2}{*}{Sample Size} & Extra Large (n=5000) & 0.668 ± 0.003 & 0.715 ± 0.017& 0.728 ± 0.017 & 0.722 ± 0.017 & \textbf{0.759 ± 0.027} & 0.167 ± 0.016\\
  & & Large (n=2000) & 0.623 ± 0.041 & 0.682± 0.020& 0.703 ± 0.010 & 0.732 ± 0.008 & \textbf{0.795 ± 0.055} & 0.178 ± 0.016 \\
  \cmidrule(lr){2-9}
  & \multirow{2}{*}{Noise} & Non-Gaussian & \textbf{0.828 ± 0.163} & 0.679 ± 0.241 & 0.657 ± 0.204 & 0.327 ± 0.201 & 0.714 ± 0.251 & 0.243 ± 0.141 \\
  & & Gaussian & \textbf{0.888 ± 0.060} & 0.651 ± 0.221 & 0.655 ± 0.177 & 0.563 ± 0.308 & 0.690 ± 0.206 & 0.419 ± 0.281\\
  \bottomrule
  \end{tabular}
  }
  \end{table*}

\subsection{Case Study}
\begin{figure*}
  \centering
  \includegraphics[scale=0.47]{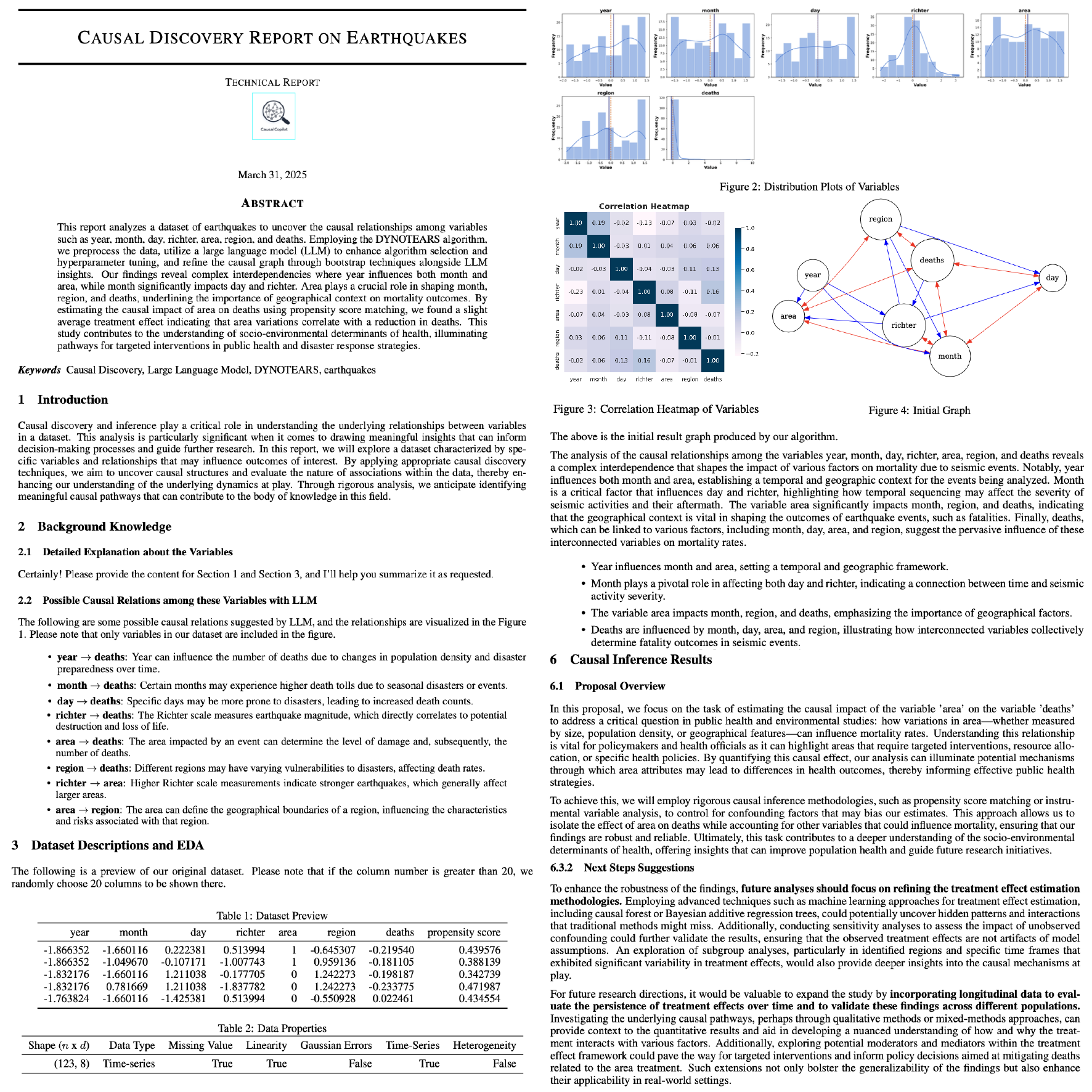}
  \caption{An example report generated by our Causal-Copilot. The left part demonstrates the dataset description, and the right part shows the visualization of the statistics and the causal graph, with the causal analysis results.}
  \label{casestudy}
\end{figure*}

To demonstrate the practical utility of Causal-Copilot, we present a case study based on a sample dataset available on the Causal-Copilot website. As illustrated in Figure~\ref{casestudy}, the system generates a comprehensive PDF report that includes user-friendly statistical summaries, the discovered causal graph, and a step-by-step walkthrough of the causal analysis process. 
With minimal domain expertise required from the user, the case study underscores Causal-Copilot’s capacity to deliver actionable insights on real-world data.

\section{Conclusion}
In this paper, we presented Causal-Copilot, an LLM-based autonomous agent that bridges the gap between advanced causal analysis methods and real-world applications, by enabling the automatic selecting, configuring, and executing of causal analysis algorithms. 
By leveraging LLMs with a causality-related knowledge memory for decision making, Causal-Copilot is capable of managing the causal analysis process using over 20 cutting-edge methods, spanning from causal discovery, causal inference, and effect estimation algorithms.
Our experimental results indicate that this automated approach outperforms baseline methods, indicating the effectiveness of our LLM-based agent for automatic algorithm selection. 

We believe this work constitutes a crucial step in making advanced causal analysis methods accessible to researchers and practitioners across diverse fields. Future directions include further expansion of the algorithm library, considering domain-specific knowledge, enhancing the ability to handle high-dimensional or real-time data streams, and strengthening the adaptability to broader real-world applications.


\acks{We would like to thank Aseem Dandgaval for his assistance in the integration of causal inference algorithms. We also appreciate the help from Wenqin Liu, Chris Zhao and Felix Wu in improving the visualization and user interface. We are deeply grateful to our colleagues in the natural sciences who have shared their interests and challenges in uncovering causal relationships in their domains, which provided the inspiration and motivation for this work.}


\vskip 0.2in
\bibliography{main}

\newpage
\appendix

\section{Acknowledgment of Tools and Libraries}
Causal-Copilot builds upon multiple open-source libraries and frameworks that have significantly contributed to the field of causal analysis. We acknowledge and express our gratitude to the developers and maintainers of these tools. 

For causal discovery, we develop the data simulator based on NOTEARS's data generation process \citep{zheng2018dags}. We leverage comprehensive packages including Causal-learn \citep{zheng2024causal}, Gcastle \citep{zhang2021gcastle}, and CausalNex \citep{Beaumont_CausalNex_2021} which provides diverse causal discovery algorithms. We also benefit from specialized implementations such as FGES and XGES for score-based learning, AcceleratedLiNGAM \citep{ramsey2017million, nazaret2025extremely, akinwande2024acceleratedlingam} for GPU-accelerated linear non-Gaussian methods, GPU-CMIknn  and GPUCSL for GPU-accelerated skeleton discovery \citep{hagedorn2022gpu, zarebavani2019cupc}, pyCausalFS \citep{yu2020causality} for markov-blanket based feature selection, NTS-NOTEARS \citep{zheng2018dags} for the non-linear time-series structure learning approach and Tigramite \citep{runge2019detecting} for constraint-based time series causal discovery. For causal inference, we integrate DoWhy \citep{dowhy}, which implements a four-step methodology (model, identify, estimate, refute) for causal effect estimation, and EconML \citep{econml}, a toolkit for applying machine learning to econometrics with a focus on heterogeneous treatment effects.

Our work stands on the shoulders of these excellent tools, and we have aimed to extend their capabilities by integrating them into a unified, LLM-guided framework that makes causal analysis more accessible to non-experts.

\newpage
\section{Prompt in Causal Copilot}
We list our used prompts for the algorithm selection and reranking in the follows, where the underlined words are the placeholders for corresponding contents provided in the execution process.

\definecolor{lightgreen}{RGB}{145, 204, 117}

\begin{promptbox}[Prompt for Algorithm Selection]{lightgreen}
User original query (TOP PRIORITY): \underline{[USER QUERY]}

The computation has to be finished in the runtime of \underline{[WAIT TIME]} minutes. \\

\textbf{CRITICAL USER PRIORITY DIRECTIVE}

1. User query overrides ALL other considerations

2. Extract expertise, constraints, and requirements from user query FIRST

3. Prioritize fulfilling user's specific needs over general algorithm metrics

4. User-provided domain knowledge supersedes general best practices

5. EVERY recommendation MUST be directly traceable to the user's requirements \\

For the dataset \underline{[TABLE NAME]} that have the following variables: \underline{[COLUMNS]}

And the following statistics: \underline{[STATISTICS DESC]}

And the relevant domain knowledge:
\underline{[DOMAIN KNOWLEDGE]} \\

All candidate algorithms, their descriptions and tags: \underline{[ALGO CONTEXT]} \\

\textbf{CRITICAL SELECTION REQUIREMENT}

When ALL data properties match between algorithms, you MUST strictly adhere to the performance rating hierarchy (Robust $>$ Strong $>$ Moderate $>$ Limited $>$ Poor) from the tagging information. Never select a lower-rated algorithm over a higher-rated one with matching properties. Reflect if the choosen algorithm is the BEST on performance.

I need you to carefully analyze and select the most suitable causal discovery algorithms (up to \underline{[TOP K]}) through a comprehensive multi-step reasoning and decision process as follow:\\

\textbf{Primary Analysis: Data and Requirements Assessment}

1. **User Goal Analysis**:

   - What is the primary causal question the user is trying to answer?
   
   - Is the focus on prediction, explanation, or intervention?
   
   - What degree of interpretability is required?

   - What additional expert knowledge from user that would be relevant for causal structure learning? (e.g., domain-specific data properties)

2. **Data Characteristics Analysis**:

   - Sample size (n): Is it sufficient for statistical power? (small: $<$500, medium: 500-5000, large: $>$5000)
   
   - Variable count (p): How many variables need to be considered? Consider these thresholds:
   
     * Small scale ($<$25 variables): Most algorithms perform well
     
     * Medium scale (25-50 variables): Requires "High" or better scalability rating
     
     * Large scale (50-110 variables): Requires "Very High" or better scalability rating
     
     * Very large scale ($>$110 variables): Requires "Extreme" scalability rating

$\cdots$

3. **Resource Constraints**:

   - Computational resources: GPU availability, memory limitations, time constraints
   
   - Output format requirements: Is a DAG, CPDAG, or PAG preferred or required?\\

\textbf{REQUIRED: Extensive Reasoning Process}

You MUST provide comprehensive reasoning at each step, explicitly connecting dataset characteristics to algorithm selection decisions. Include detailed analysis of why certain algorithms are superior for THIS SPECIFIC dataset while others are unsuitable. \\

\textbf{CRITICAL DIVERSITY INSTRUCTION}

FOCUS EXCLUSIVELY ON THE CURRENT DATASET CHARACTERISTICS. PRIORITIZE ALGORITHMIC DIVERSITY by selecting algorithms from different methodological families (e.g., score-based, constraint-based, continous-optimization-based $\cdots$ ) when multiple algorithms are equally compatible with the requirements.

Your final response should include the complete reasoning process, for each algorithm, include justification, description, and selected algorithm in a JSON object.
\end{promptbox}

\begin{promptbox}[Prompt for Algorithm Reranking]{lightgreen}
You are an expert causal discovery algorithm advisor. Your task is to analyze dataset \underline{[TABLE NAME]} and recommend the most appropriate algorithm from the candidates.\\

\textbf{HIGHEST PRIORITY INSTRUCTION}

1. The user's query MUST be your primary consideration - all recommendations MUST directly address their specific needs

2. Extract and apply ALL domain knowledge and requirements from the user's query

3. Your algorithm selection MUST prioritize fulfilling the exact requirements stated by the user

4. When user expertise contradicts general best practices, ALWAYS favor the user's domain knowledge\\

User original query (TOP PRIORITY): \underline{[USER QUERY]}

The computation has to be finished in the runtime of \underline{[WAIT TIME]} minutes.

\underline{[ACCEPT CPDAG]}

Dataset Profile

- Characteristics: \underline{[STATISTICS INFO]}

- Domain context: \underline{[KNOWLEDGE INFO]}

- Hardware capabilities: \underline{[CUDA WARNING]}

- Columns: \underline{[COLUMNS]}\\

\textbf{CRITICAL INSTRUCTION}

FOCUS EXCLUSIVELY ON THE CURRENT DATASET CHARACTERISTICS. Do not consider hypothetical scenarios or algorithm capabilities that aren't relevant to this specific dataset. Analyze only the actual properties of this dataset (sample size, dimensionality, distributions, etc.) without discussing how algorithms might perform on different datasets.

\textbf{Selection Process (Follow in sequence)}

1. Dataset-Algorithm Compatibility Analysis
For each algorithm candidate, evaluate compatibility across these dimensions BASED SOLELY ON THE CURRENT DATASET:

- Variable type handling: How well does the algorithm process THIS dataset's {continuous/discrete/mixed} variables?

$\cdots$

2. Critical Assumption Verification

For each algorithm:

- Identify which core assumptions would be violated by THIS SPECIFIC dataset

$\cdots$

3. Computational Feasibility Assessment

For each algorithm:

- Estimate runtime based on THIS dataset's dimensions=\underline{[DIMENSIONS]}, samples=\underline{[SAMPLES]}, and expected graph density

$\cdots$

4. Domain-Specific Considerations (SKIP if you find the dataset is not real-world dataset)

**NOTE: infer if this dataset is a real-world dataset from dataset name and column name, SKIP this part for numerical simulation dataset since it doesn't indicate any domains**

- Does THIS domain require handling feedback loops? If yes, eliminate acyclic-only algorithms.

$\cdots$

5. Performance-Interpretability Balance

- What's more important in THIS domain and for THIS dataset: causal accuracy or interpretability?

$\cdots$\\

\textbf{IMPORTANT NOTE ON ALGORITHM SELECTION}

It is acceptable to recommend algorithms that may be "overqualified" for the dataset. If an algorithm can deliver robust performance and superior accuracy without significantly sacrificing efficiency, it should be considered even if simpler alternatives exist. Prioritize algorithms that will provide the best possible causal discovery results, as long as they meet the computational constraints. The goal is optimal performance, not minimal sufficiency.

The return algorithm name MUST strictly be FROM candidates here: 
\underline{[ALGORITHM CANDIDATES]}

Algorithm Profiles: \underline{[ALGORITHM PROFILES]}

\textbf{Simulation Benchmarking Results}

Each algorithm's benchmarking with several hyperparameter choices. Extract and analyze the following from the benchmarking results: \underline{[ALGORITHM BENCHMARKING RESULTS]}

Based on this comprehensive analysis, provide TWO separate scoring assessments:

1. THEORETICAL ASSESSMENT:

   - Score each algorithm (1-5 scale) based on theoretical properties from algorithm profiles
   
$\cdots$

Output in the following JSON format:

$\cdots$

\end{promptbox}

\begin{promptbox}[Description Profile for An Example Algorithm PC]{lightgreen}

\textbf{Executive Summary}

PC is a constraint-based causal discovery algorithm that excels in accurately reconstructing moderate-sized sparse causal networks with primarily linear relationships, producing interpretable partially directed graphs. It's best suited for exploratory causal analysis with adequate sample sizes where causal sufficiency can be reasonably assumed. With GPU acceleration, it can also efficiently handle much larger networks ($>1000$ nodes).

\textbf{1. Real-World Applications}

- **Best Use Cases**

  - **Genomics**: Effectively identifies gene regulatory networks from expression data. Research by Le et al. demonstrated PC's effectiveness in reconstructing gene regulatory networks from microarray data, particularly with smaller gene sets (like 50 genes).

$\cdots$

- **Limitations**

  - **Large-Scale Networks with CPU**: Performance significantly deteriorates with networks exceeding 100 variables when using CPU-based implementations.
  - **Highly Connected Systems**: Accuracy drops sharply with densely connected causal structures.
  - **Systems with Feedback Loops**: Cannot model cyclic causal relationships due to acyclicity assumption.
  - **Data with Hidden Confounders**: Produces misleading results when unmeasured common causes exist.

\textbf{2. Assumptions}

- **Causal Sufficiency**: Assumes all common causes are measured. When violated, produces spurious edges and incorrect orientations.

$\cdots$

\textbf{3. Data Handling Capabilities}

- **Variable Types**: Suited for continuous data with Fisher-Z test; can handle discrete data with chi-squared test; can also handle mixed data types with non-parametric conditional independence tests.

$\cdots$

\textbf{4. Robustness \& Scalability}

- **Missing Data Tolerance**: Poor tolerance for missing values; performance degrades rapidly with missingness above 5

$\cdots$

\textbf{5. Computational Complexity}

- **Theoretical Time Complexity**: Worst-case $O(n^{(k+2)})$, where n is the number of variables and k is the maximum degree in the true graph.

$\cdots$

\textbf{6. Hyperparameters}

- **Number of Hyperparameters**: Only three key hyperparameters (alpha, independence test, depth), making tuning more manageable than algorithms with numerous parameters.

$\cdots$

\textbf{7. Interpretability}

- **Output Format**: Produces a Completed Partially Directed Acyclic Graph (CPDAG) that clearly distinguishes definite causal edges from undetermined relationships.

$\cdots$

\end{promptbox}

\section{Detailed Simulation Experiment Settings}\label{app:pre_exp_setup}

\subsection{Preliminary benchmarking on tabular causal discovery algorithms}

To systematically evaluate causal discovery algorithms across diverse scenarios, we developed a comprehensive data generation framework that simulates various data characteristics commonly encountered in real-world applications, based on the data generation process in \citep{zheng2018dags}. Our simulator generates synthetic datasets with known ground-truth causal structures, allowing for precise evaluation of algorithm performance.

\subsubsection{Data Generation Process}
Our data generation process begins with creating a directed acyclic graph (DAG) structure using Erdős–Rényi (ER) graph models. For each node in the graph, we simulate data according to structural equation models (SEMs) that follow the causal relationships defined by the DAG.

The simulator supports various data characteristics:
\begin{itemize}
    \item \textbf{Relationship Types}: Both linear and non-linear causal relationships, including polynomial, MLP (Multi-Layer Perceptron), MIM (Multiple Index Model), and GP (Gaussian Process) based functions.
    \item \textbf{Noise Distributions}: Gaussian, exponential, Gumbel, uniform, logistic, and Poisson noise distributions.
    \item \textbf{Variable Types}: Continuous variables and discrete categorical variables with configurable ratios. The discrete variables are simulated by generating logits first from its parent nodes and then sample from the corresponding categorical distribution.
    \item \textbf{Data Quality Issues}: Configurable measurement error (with adjustable error standard deviation and affected column percentage) and missing data rates (with configurable missing value percentage). Measurement errors are implemented by adding Gaussian noise with specified standard deviation to selected columns (measurement error ratio) while missing values are randomly introduced to the whole data samples according to the specified rate and replaced with NaN values.
    \item \textbf{Multi-Domain Data}: Generation of data from multiple domains with domain-specific effects. Domain effects are implemented by applying both linear and nonlinear transformations to selected variables, with effect strength proportional to domain index. For linear relationships, a constant offset is added, while for nonlinear relationships, quadratic transformations are applied to be added.
\end{itemize}

\subsubsection{Benchmarking Dataset Structure}
For our preliminary benchmarking, we systematically varied key data characteristics to evaluate algorithm performance across different scenarios. Our simulation configuration included:

\begin{itemize}
    \item \textbf{Default Configuration}: Our baseline setup included 10 nodes, 1000 samples, edge probability of 0.22 (average degree of 2), linear functions, Gaussian noise with scale 1.0, no discrete variables, no measurement errors, and no missing values.
    \item \textbf{Network Size}: Varying the number of nodes (5, 10, 25, 50, 100, 200, 500, 1000) to assess scalability.
    \item \textbf{Sample Size}: Different sample counts (500, 1000, 2500, 5000, 10000) to evaluate sample efficiency.
    \item \textbf{Graph Density}: Edge probabilities ranging from 0.11 to 0.78 (corresponding to average degree of 1 to 7 when the variable size is 10)to test performance on sparse and dense causal networks.
    \item \textbf{Noise Distributions}: Gaussian and uniform noise to assess robustness to different error distributions.
    \item \textbf{Discrete Variables}: Varying ratios of discrete variables (0\%, 10\%, 20\%) to test handling of mixed data types.
    \item \textbf{Data Quality}: Different rates of measurement error (10\%, 30\%, 50\%) and missing data (10\%, 20\%, 30\%).
    \item \textbf{Multi-Domain Data}: Generating data from single and multiple domains (1, 2, 5) to evaluate the robustness under heterogeneous data.
\end{itemize}

For each configuration, we generated multiple datasets with different random seeds to ensure robust evaluation. For all configurations except for the discrete ratio variation (where we only considered linear relationships), we simulated both linear and non-linear function types to comprehensively evaluate algorithm performance across different functional forms.

\subsection{Evaluation settings of Causal-Copilot on tabular data}

For evaluating Causal-Copilot on tabular data, we utilized the same data generation framework described above to create a diverse set of test scenarios. Our evaluation dataset comprised three categories of increasing complexity:

\begin{itemize}
    \item \textbf{Basic Scenarios}: These included variations in sample sizes (1,000 to 10,000), network sizes (10 to 1,000 nodes), relationship types (linear and non-linear MLP functions), noise distributions (Gaussian and uniform), graph densities (sparse to dense with edge probabilities from 0.11 to 0.56), and mixed variable types (with 20\% discrete variables). These scenarios tested the fundamental capabilities of causal discovery algorithms under relatively clean conditions.
    
    \item \textbf{Data Quality Challenges}: This category introduced specific data quality issues including missing values (30\% missing rate), measurement errors (30\% of columns affected with 0.1 standard deviation error), and heterogeneous domains (data from 3 different domains with domain-specific effects). These scenarios evaluated algorithm robustness to common data imperfections.
    \item \textbf{Compound Challenges}: These complex scenarios simulated real-world applications with multiple simultaneous challenges:
    \begin{itemize}
        \item \textit{Clinical Data Scenario}: 25 nodes with sample size 1000, discrete variables (0.3), measurement errors (0.2), missing values (0.05), and multi-domain effects (2 domains). This simulates typical clinical datasets where patient records contain both continuous measurements and discrete diagnostic codes, with measurement errors from different instruments, missing data from incomplete records, and data collected across different hospital departments.
        
        \item \textit{Financial Data Scenario}: 50 nodes with sample size 5000, sparse connections (average degree 1), minimal measurement errors (0.1 with 0.05 standard deviation) and missing values (0.01), with data from 3 domains. This represents financial time series data where many variables have few direct causal relationships, with high-quality but still imperfect data collection systems typical in regulated financial environments.
        
        \item \textit{Social Network Influence Scenario}: 1000 nodes with sample size 2000, moderate connectivity (average degree 6 compared to 1000 node size), measurement errors (0.1), and missing values (0.1). This simulates social network data where user behaviors influence each other in complex patterns, with measurement errors in behavior tracking and missing data from user inactivity periods.
    \end{itemize}
\end{itemize}

For scenarios with discrete data, we only consider linear function type. For each scenario, except those involving discrete data, we generated datasets with both linear and non-linear relationships to evaluate Causal-Copilot performance across different functional forms.

\subsection{Preliminary benchmarking on time-series causal discovery algorithms}
Following \cite{Beaumont_CausalNex_2021}, we develop a rigorous data generation framework for time series data. Our simulator is tailored towards generating stationary time series with predominantly linear causal relations. Additionally, the simulator outputs a ground truth lagged causal graph and a ground truth summary graph, allowing for a comprehensive evaluation. 
\subsubsection{Data Generation Process}
The generation process begins with constructing a temporal causal graph that includes both intra-time-slice (instantaneous) and inter-time-slice (lagged) edges based on user-specified parameters such as the number of nodes, maximum lag order, edge densities, and relative strengths of intra- and inter-node dependencies. The underlying causal model is defined by linear structural equations, where each variable is expressed as a weighted combination of its causal parents and an additive noise term. The simulator supports various noise distributions, including Gaussian, Exponential, and Uniform, and ensures the resulting system is stationary by keeping the coefficient matrices fixed over time.
The simulator supports a wide range of configurations for generating \textbf{stationary}, \textbf{linear}, multivariate time series with known dynamic causal structures. The key data generation options include:

\begin{itemize}
    \item \textbf{Temporal Dependencies:} 
    The simulator generates both instantaneous (intra-slice) and time-lagged (inter-slice) causal dependencies between variables. The temporal depth is configurable, allowing each variable to be influenced by a fixed number of past time steps.

    \item \textbf{Graph Structures:}
    Causal graphs can be constructed using various topologies for both intra-slice and inter-slice structures, including Erdős–Rényi random graphs, Barabási–Albert scale-free networks, and fully connected graphs. The density of edges in these graphs is controlled through the expected degree of connectivity.

    \item \textbf{Edge Weight Variability:}
    The strengths of causal relationships can be sampled from customizable ranges for both instantaneous and lagged connections. Additionally, a temporal decay mechanism is supported, enabling the influence of more distant lags to diminish according to a power law.

    \item \textbf{Noise Distributions:}
    Observations are generated from a structural equation model with additive noise. The simulator allows selection from several noise distributions, including Gaussian, exponential, and Gumbel. The overall magnitude of the noise is also configurable.

    \item \textbf{Scalability and Sampling Control:}
    The simulator allows flexible specification of the number of variables and the total number of time steps to be generated, supporting a range of dataset sizes suitable for benchmarking.
\end{itemize}

This configurable setup enables the controlled generation of realistic linear time series datasets, which serve as a robust testbed for the evaluation of time series causal discovery algorithms.

\subsubsection{Benchmarking Dataset Structure}
For our preliminary benchmarking, we systematically varied key aspects of the time series generation process to evaluate the performance of causal discovery algorithms under different scenarios. Our experimental configurations included:
\begin{itemize}
    \item \textbf{Default Configuration:} 
    The baseline setup included 10 nodes, a maximum lag of 3, average inter-slice degree of 3.0, average intra-slice degree of 2.0, linear structural equations, and Gaussian noise with unit scale.

    \item \textbf{Network Size:} 
    To assess scalability, we varied the number of variables in the system across a range of values: 5, 10, 25, 50, and 100 nodes. The temporal lag and edge densities were kept fixed.

    \item \textbf{Lag Order:} 
    We explored the impact of temporal dependency length by varying the maximum lag from 3 to 20, including intermediate values 5, 10, and 15. The number of nodes and edge densities were held constant.

    \item \textbf{Graph Density:} 
    To examine the influence of network sparsity and connectivity, we varied the expected number of inter-slice edges per node with average degrees of 2.0, 4.0, 8.0, 12.0, and 16.0. This allows for the evaluation of algorithm robustness across sparse and dense temporal graphs.

    \item \textbf{Sample Size:} 
    We tested the sample efficiency of algorithms by generating datasets with varying time series lengths: 500, 1000, 2000, and 5000 time steps.

    \item \textbf{Intra-Slice Edges Only:} 
    To isolate the effect of instantaneous relationships, we generated datasets with no lagged dependencies and varied the number of nodes (10, 20, 30) and the lag values (3, 10) while setting the inter-slice degree to zero.

    \item \textbf{Inter-Slice Edges Present:} 
    In scenarios where both instantaneous and temporal edges were present, we explored combinations of node counts (10, 20, 30) and lag values (3, 10) with fixed intra- and inter-slice degrees, enabling assessment of algorithms under mixed dependency structures.
\end{itemize}

These configurations provide a broad and systematic landscape for evaluating the scalability, temporal sensitivity, and robustness of time series causal discovery methods under stationary linear dynamics.
\subsection{Evaluation settings of Causal-Copilot on time-series data}
For evaluation, we used our data generation function to create some basic scenarios with varying complexity. 
\begin{itemize}
    \item \textbf{Scenarios:} The evaluation scenarios comprise the following: variations in sample sizes (500 time points to 10,000 time points), number of variables (5 to 100), varying time lags (3 to 20), and testing under the presence of different noise distributions (Gaussian and Exponential). 
\end{itemize}

\section{Evaluation Metrics}\label{app:evaluation_protocol}
We employ a standardized evaluation protocol to assess the performance of causal discovery algorithms outputs different kinds of graphs. Our evaluation focuses on the efficiency and structural accuracy of the discovered causal graphs compared to the known ground truth. For the algorithm efficiency, we cap all the runtime to be at maximum 20 minutes and record runtime cost in seconds. For the graph metrics computation, we follow the procedure below to compute precision, recall, F1 and Structural Hamming Distance (SHD).

\subsection{Tabular Data Evaluation}
For tabular data, we convert all types of causal graphs (e.g., CPDAG, PAG) to DAGs by sampling the best DAG representation from the equivalence class. This conversion ensures a consistent comparison framework across different algorithm outputs. During evaluation, we exclude self-loops by dropping the diagonal elements of the adjacency matrices, as these do not represent causal relationships in our framework.

\subsection{Time-Series Data Evaluation}
For time-series data, we compute evaluation metrics based on the summary graph representation, which captures both contemporaneous and time-lagged relationships. Similar to the tabular case, diagonal elements are excluded from the evaluation to focus on inter-variable causal relationships rather than auto-regressive effects.

\section{Detailed Preliminary Causal Discovery Algorithm Benchmarking Results} \label{app:pre_exp_res}

\begin{figure}[htbp]
  \centering
  \includegraphics[width=0.72\textwidth]{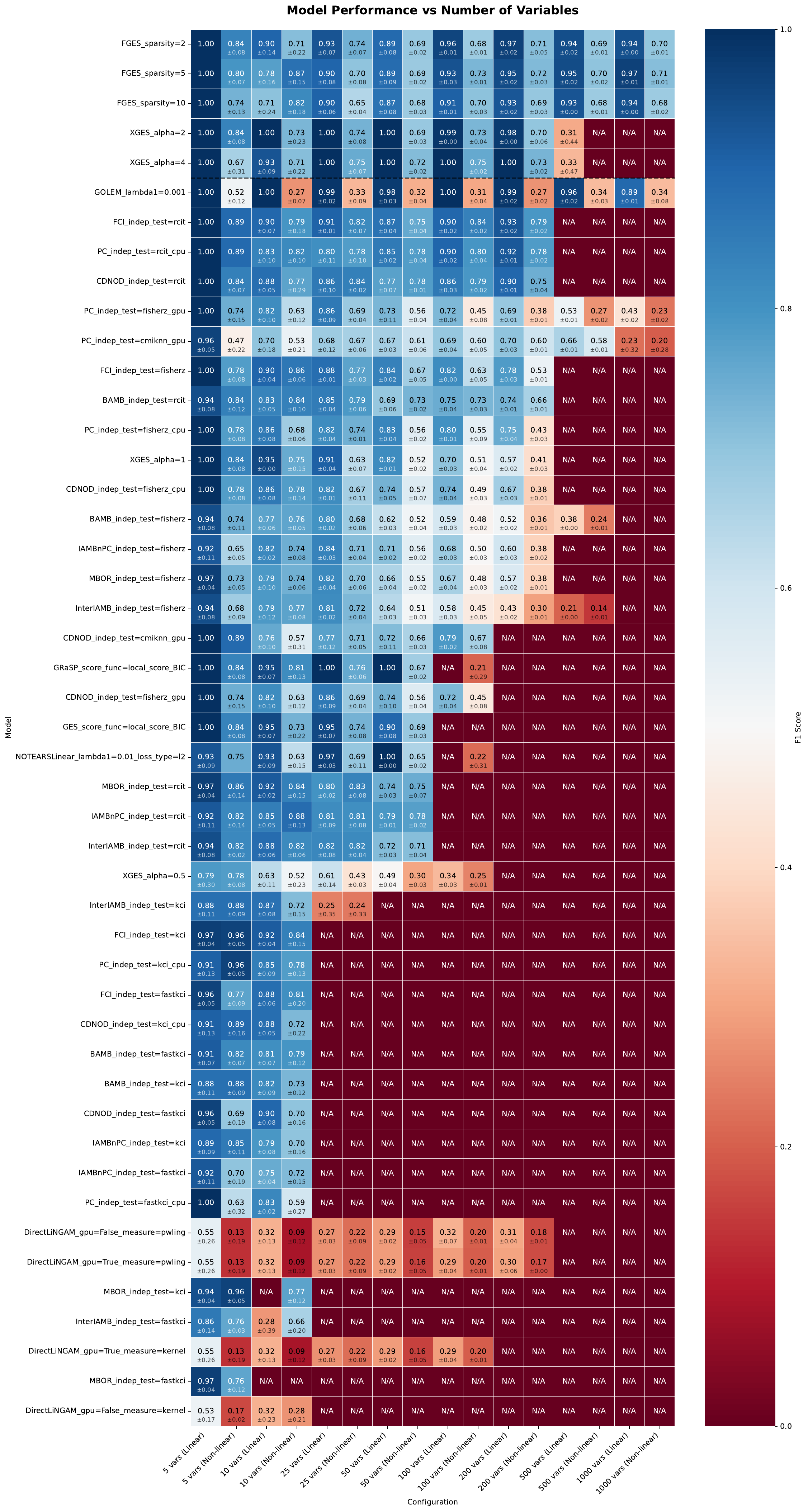}
  \caption{F1 score performance of tabular causal discovery algorithms with varying number of variables. Note that N/A indicates out-of-time in twenty minutes limitation.}
  \label{fig:pre_tab_vars}
\end{figure}

\begin{figure}[htbp]
  \centering
  \includegraphics[width=0.72\textwidth]{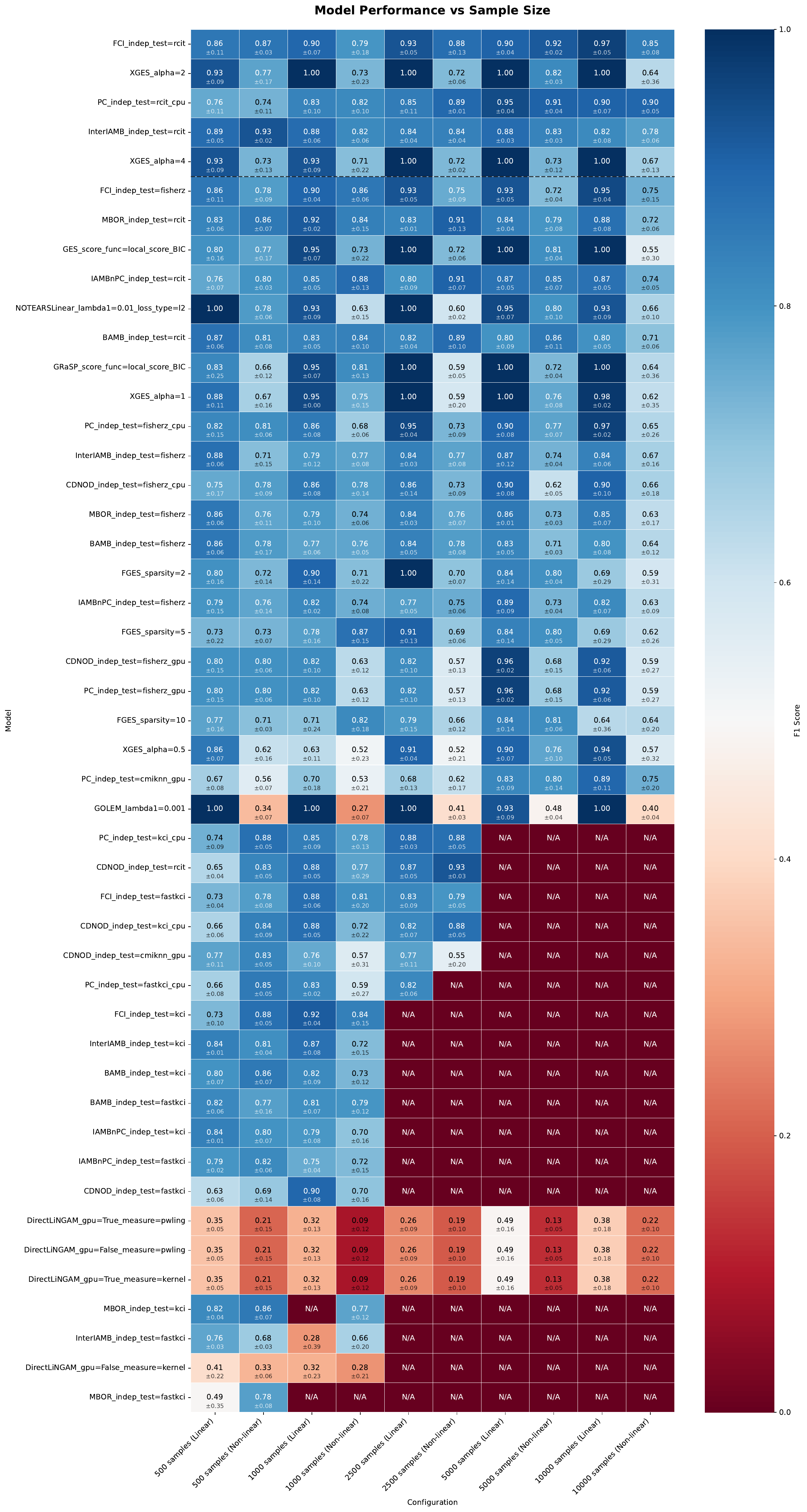}
  \caption{F1 score performance of tabular causal discovery algorithms with varying sample sizes. Note that N/A indicates out-of-time in twenty minutes limitation.}
  \label{fig:pre_tab_samples}
\end{figure}

\begin{figure}[htbp]
  \centering
  \includegraphics[width=0.72\textwidth]{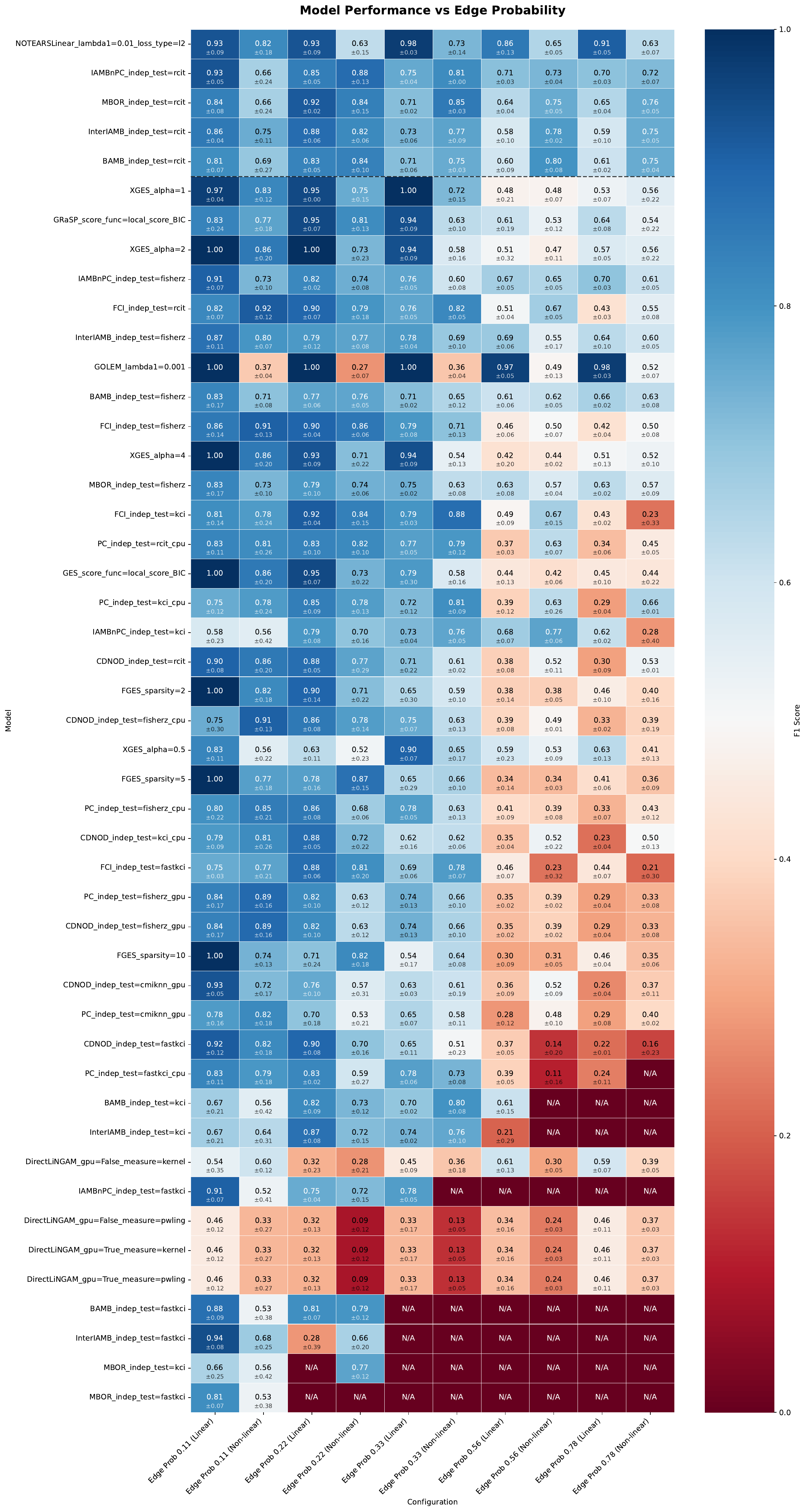}
  \caption{F1 score performance of tabular causal discovery algorithms with varying edge probabilities. Note that N/A indicates out-of-time in twenty minutes limitation.}
  \label{fig:pre_tab_edge_prob}
\end{figure}

\begin{figure}[htbp]
  \centering
  \includegraphics[width=0.72\textwidth]{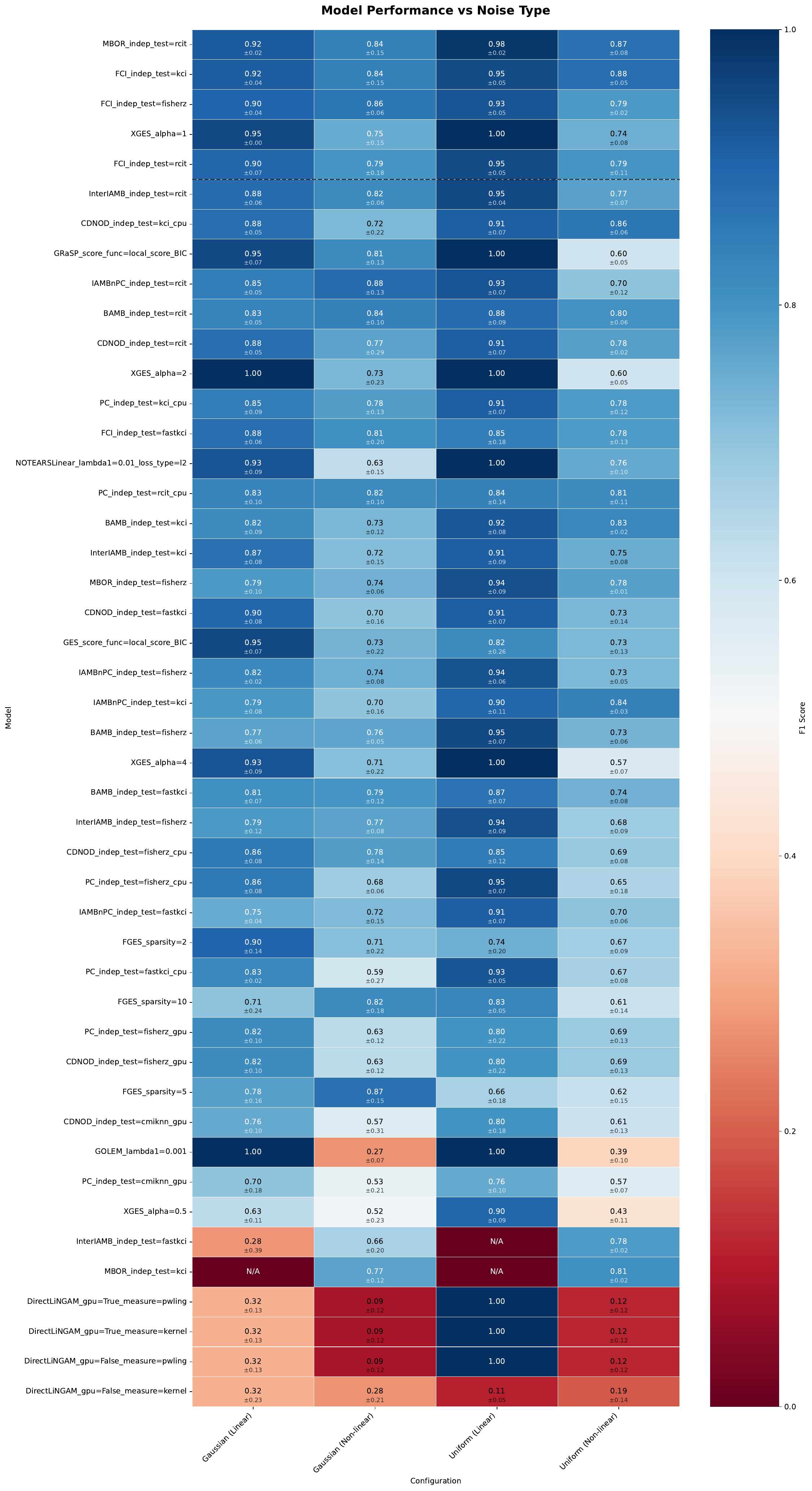}
  \caption{F1 score performance of tabular causal discovery algorithms with different noise types. Note that N/A indicates out-of-time in twenty minutes limitation.}
  \label{fig:pre_tab_noise}
\end{figure}

\begin{figure}[htbp]
  \centering
  \includegraphics[width=0.72\textwidth]{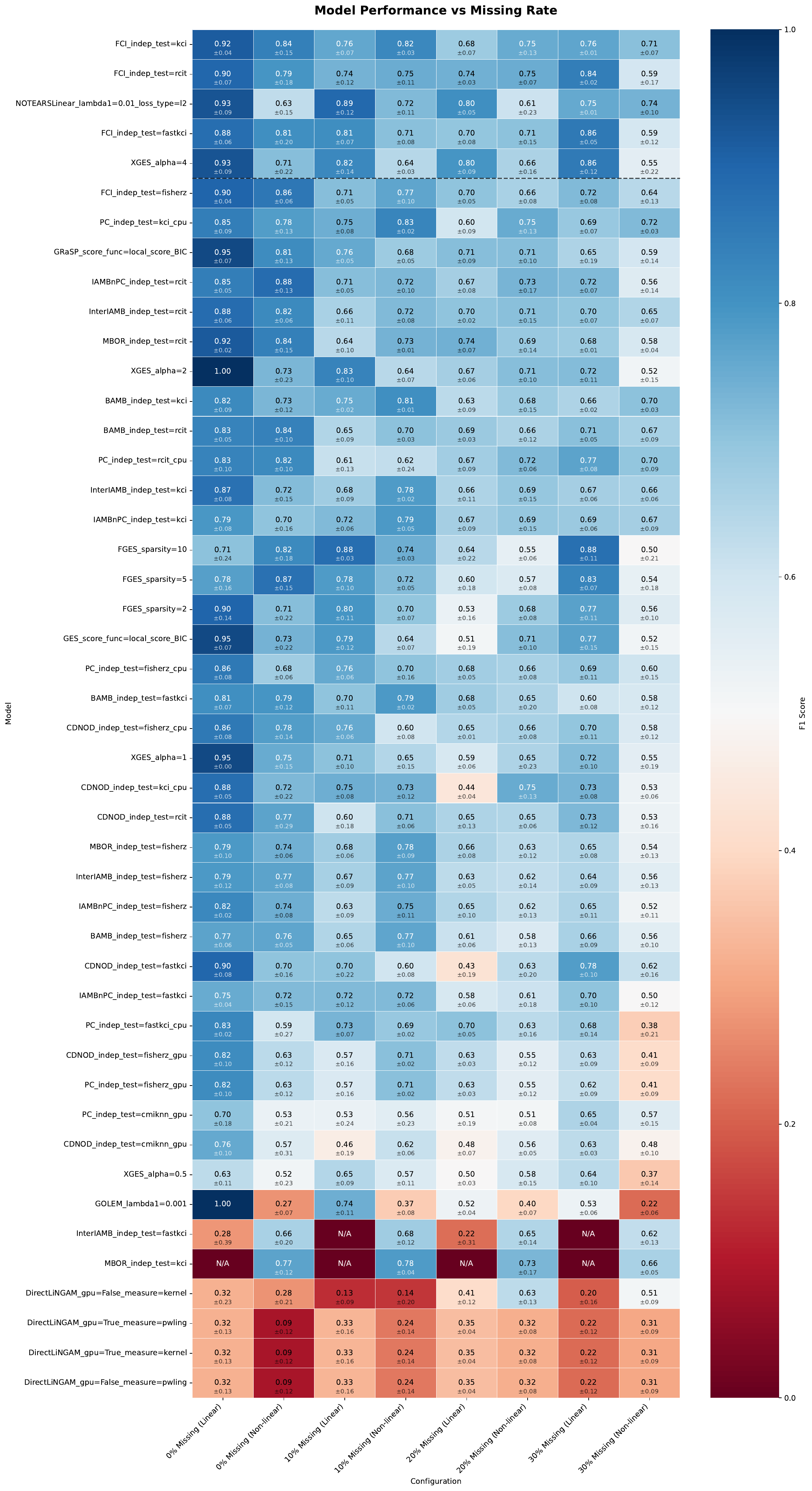}
  \caption{F1 score performance of tabular causal discovery algorithms with varying missing data rates. Note that N/A indicates out-of-time in twenty minutes limitation.}
  \label{fig:pre_tab_missing}
\end{figure}

\begin{figure}[htbp]
  \centering
  \includegraphics[width=0.72\textwidth]{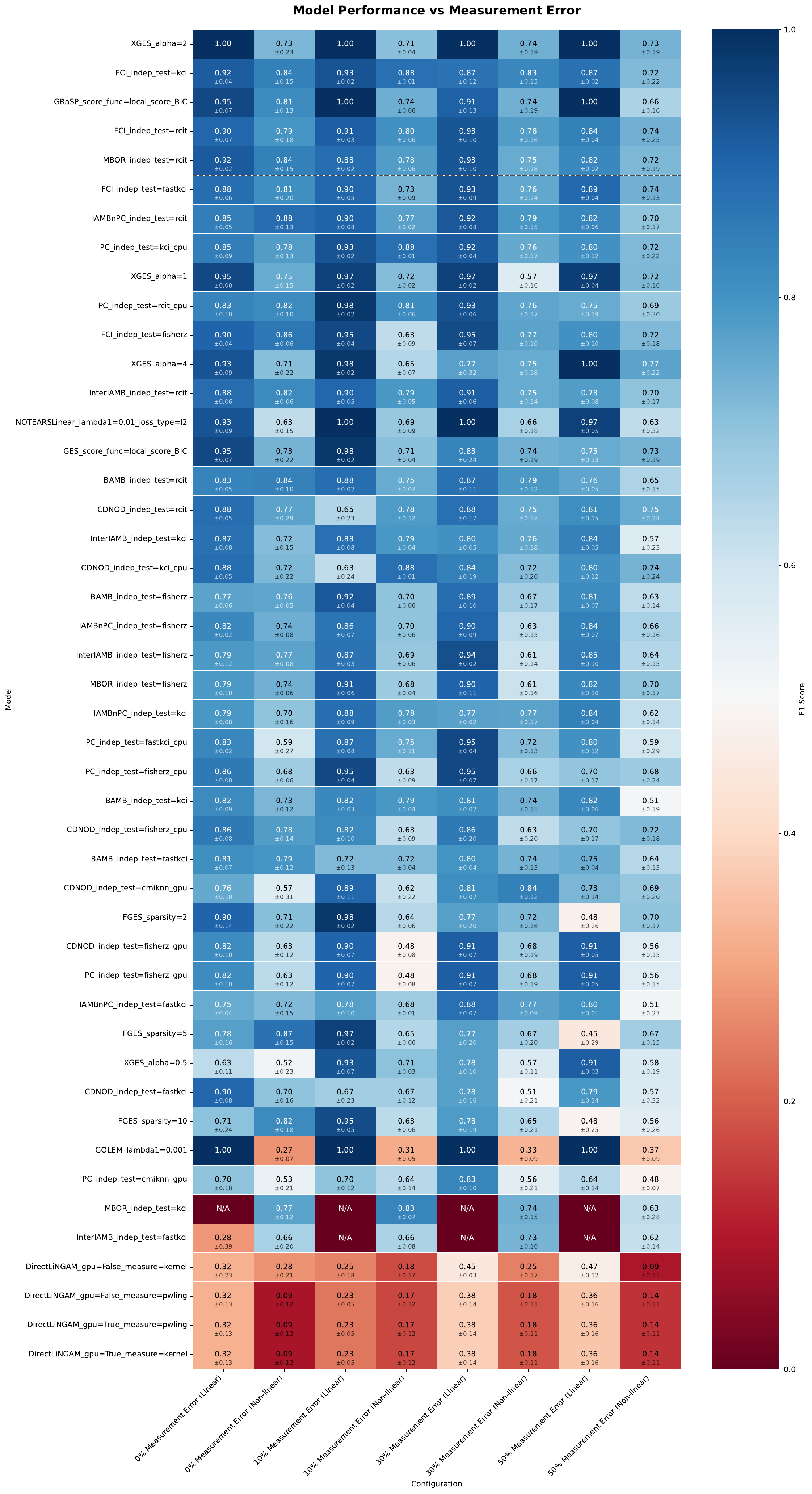}
  \caption{F1 score performance of tabular causal discovery algorithms with varying measurement error. Note that N/A indicates out-of-time in twenty minutes limitation.}
  \label{fig:pre_tab_error}
\end{figure}

\begin{figure}[htbp]
  \centering
  \includegraphics[width=0.72\textwidth]{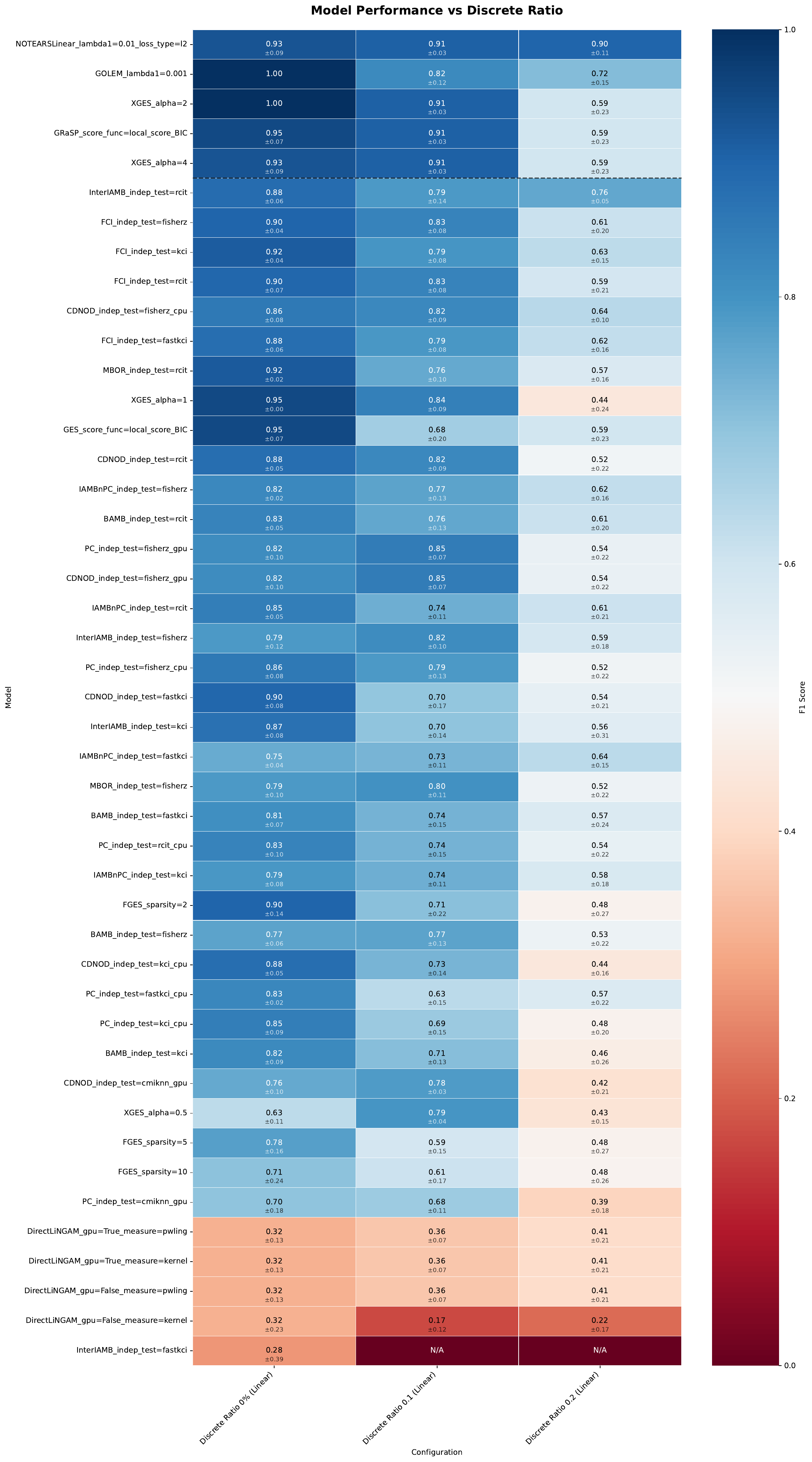}
  \caption{F1 score performance of tabular causal discovery algorithms with varying discrete variable ratios. Note that N/A indicates out-of-time in twenty minutes limitation.}
  \label{fig:pre_tab_discrete}
\end{figure}

\begin{figure}[htbp]
  \centering
  \includegraphics[width=0.72\textwidth]{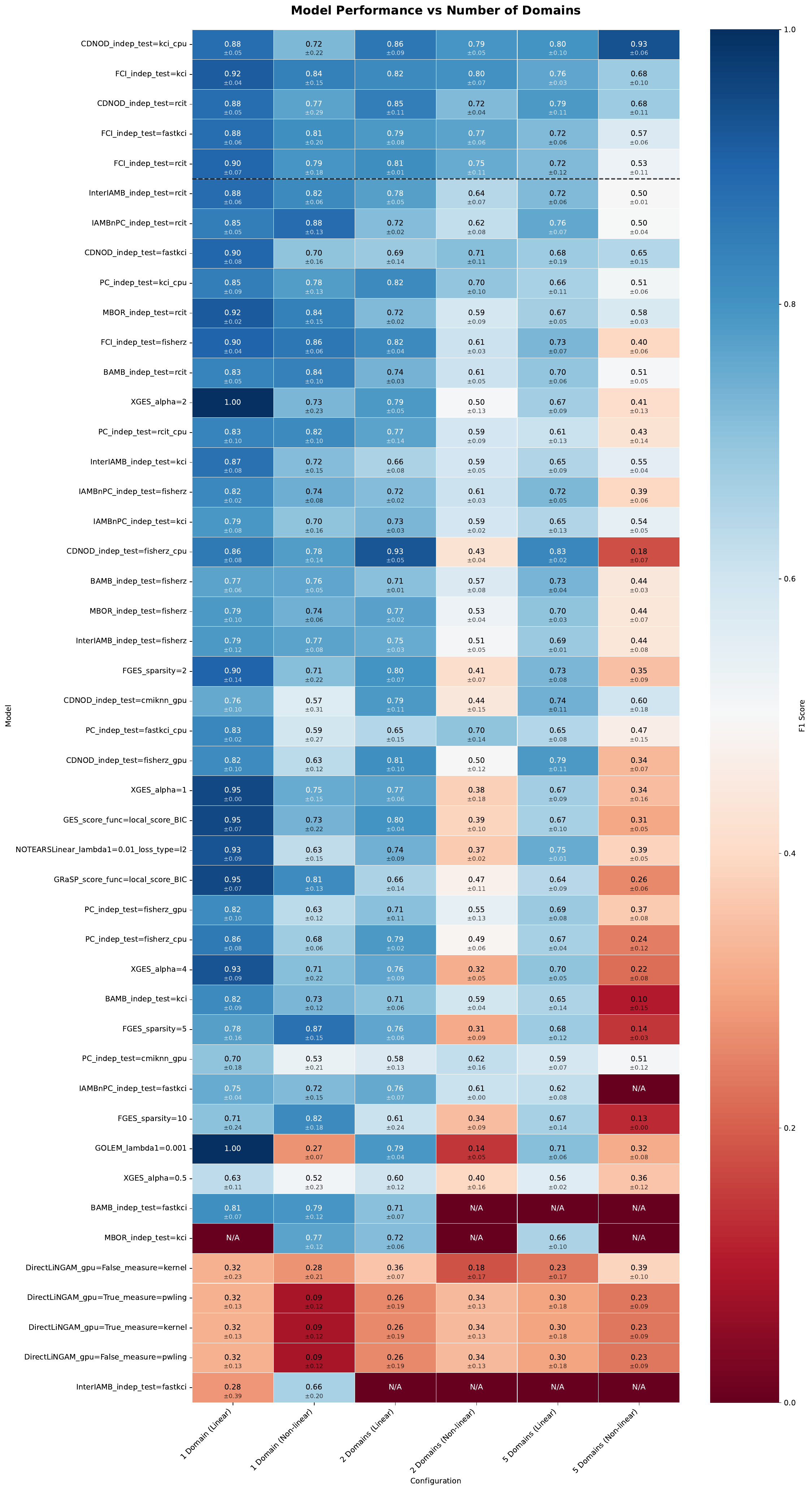}
  \caption{F1 score performance of tabular causal discovery algorithms with varying number of domains. Note that N/A indicates out-of-time in twenty minutes limitation.}
  \label{fig:pre_tab_domains}
\end{figure}

\begin{figure}[htbp]
  \centering
  \includegraphics[width=0.72\textwidth]{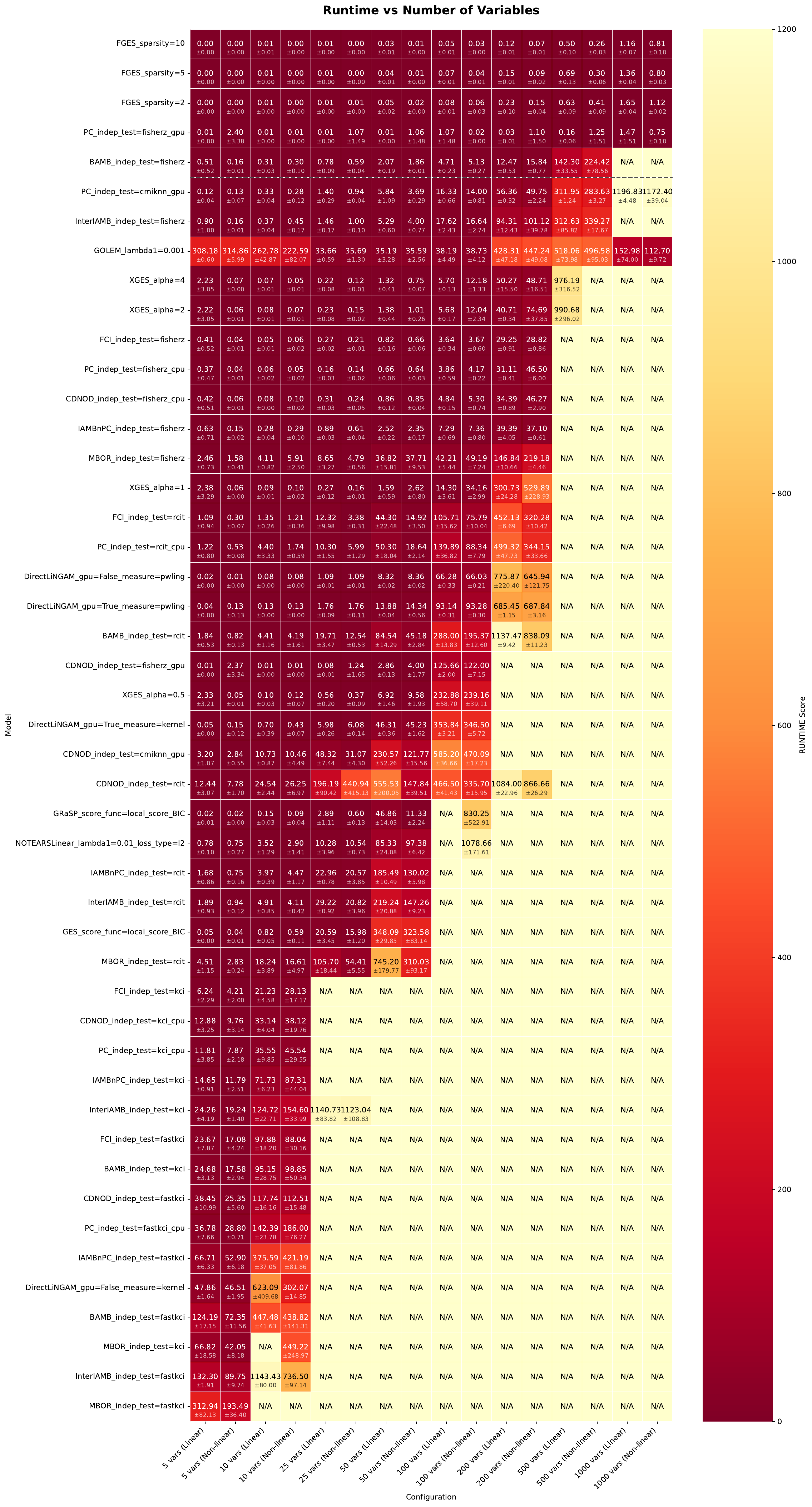}
  \caption{Runtime performance of tabular causal discovery algorithms with varying number of variables. Note that N/A indicates out-of-time in twenty minutes limitation.}
  \label{fig:pre_runtime_vars}
\end{figure}

\begin{figure}[htbp]
  \centering
  \includegraphics[width=0.72\textwidth]{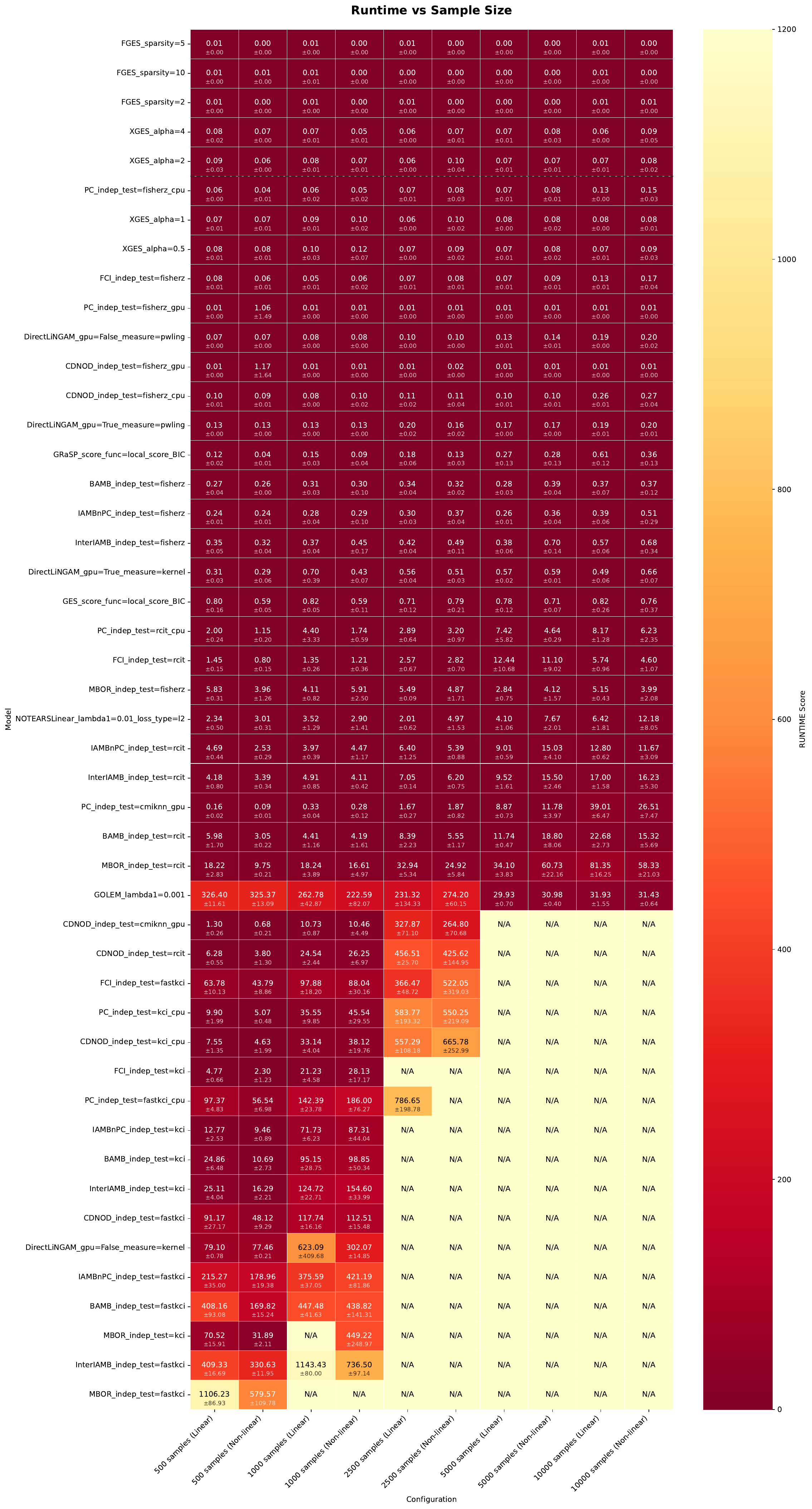}
  \caption{Runtime performance of tabular causal discovery algorithms with varying sample sizes. Note that N/A indicates out-of-time in twenty minutes limitation.}
  \label{fig:pre_runtime_samples}
\end{figure}

\begin{figure}[htbp]
  \centering
  \includegraphics[width=0.72\textwidth]{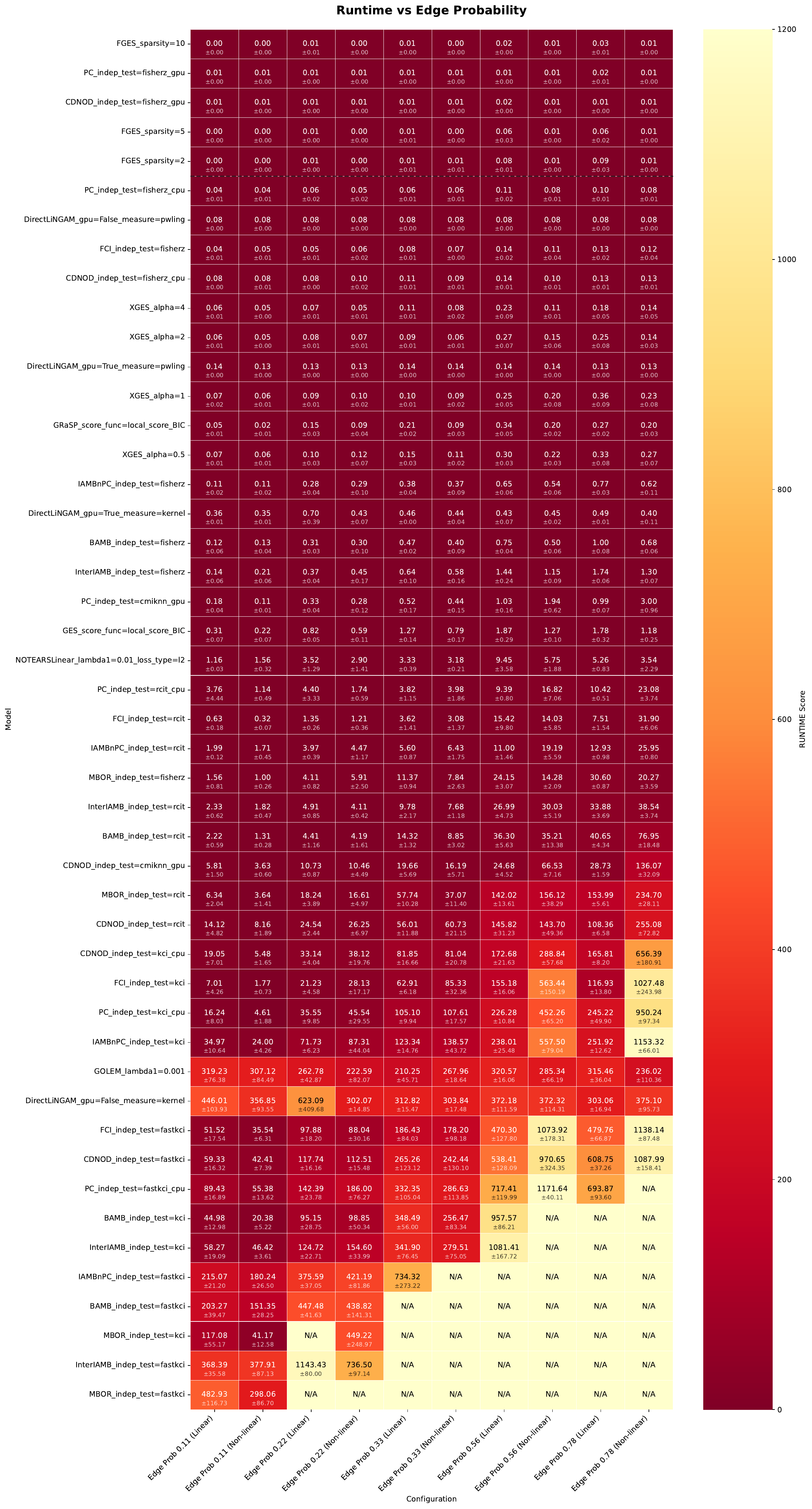}
  \caption{Runtime performance of tabular causal discovery algorithms with varying edge probabilities. Note that N/A indicates out-of-time in twenty minutes limitation.}
  \label{fig:pre_runtime_edge_prob}
\end{figure}

\begin{figure}[htbp]
  \centering
  \includegraphics[width=0.72\textwidth]{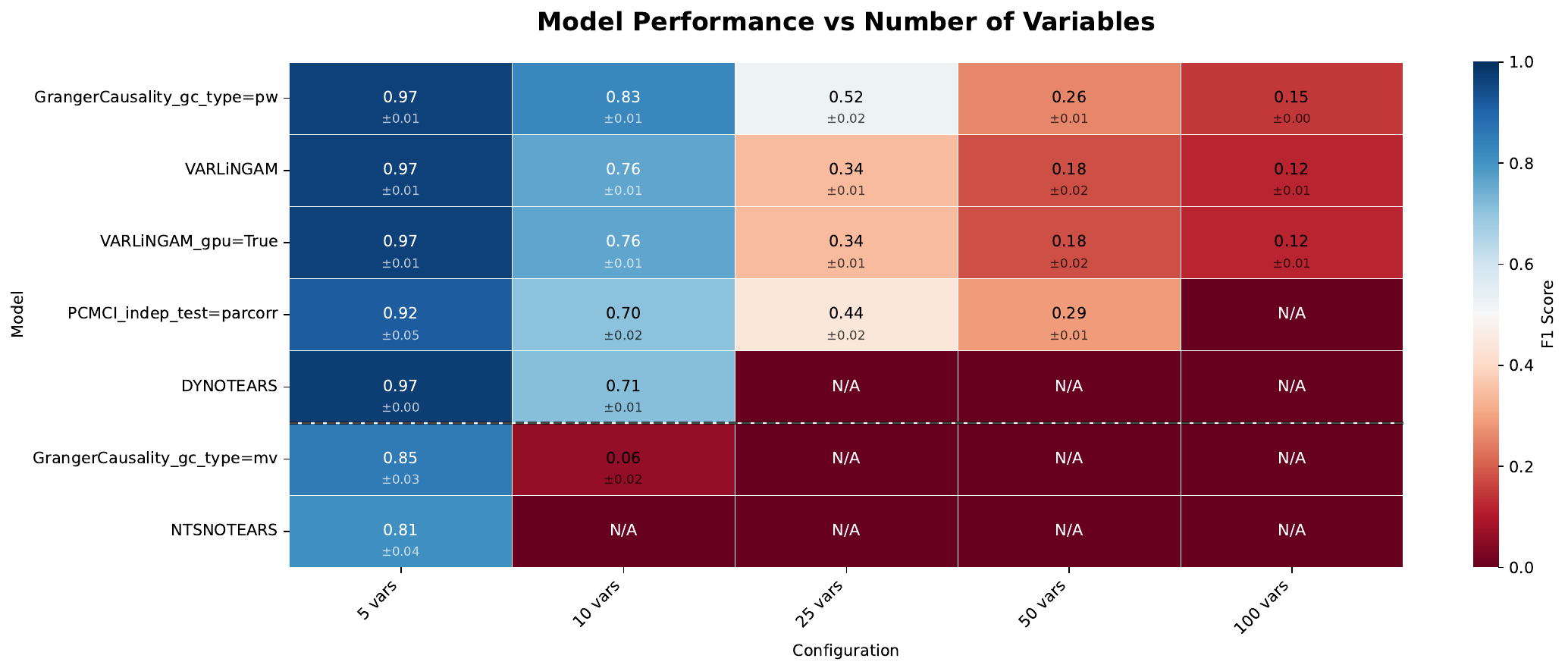}
  \caption{F1 score performance of time-series causal discovery algorithms with varying number of variables}
  \label{fig:ts_f1_vars}
\end{figure}

\begin{figure}[htbp]
  \centering
  \includegraphics[width=0.72\textwidth]{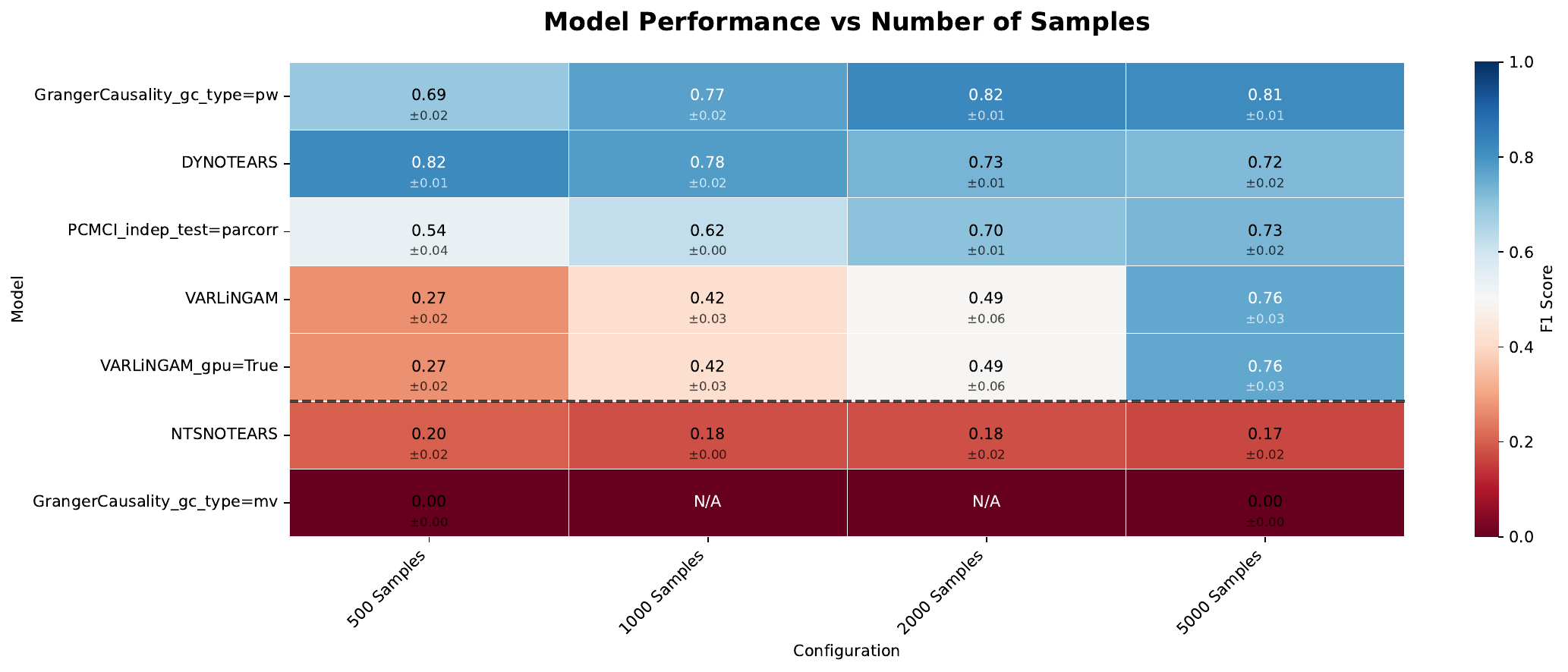}
  \caption{F1 score performance of time-series causal discovery algorithms with varying number of samples}
  \label{fig:ts_f1_samples}
\end{figure}

\begin{figure}[htbp]
  \centering
  \includegraphics[width=0.72\textwidth]{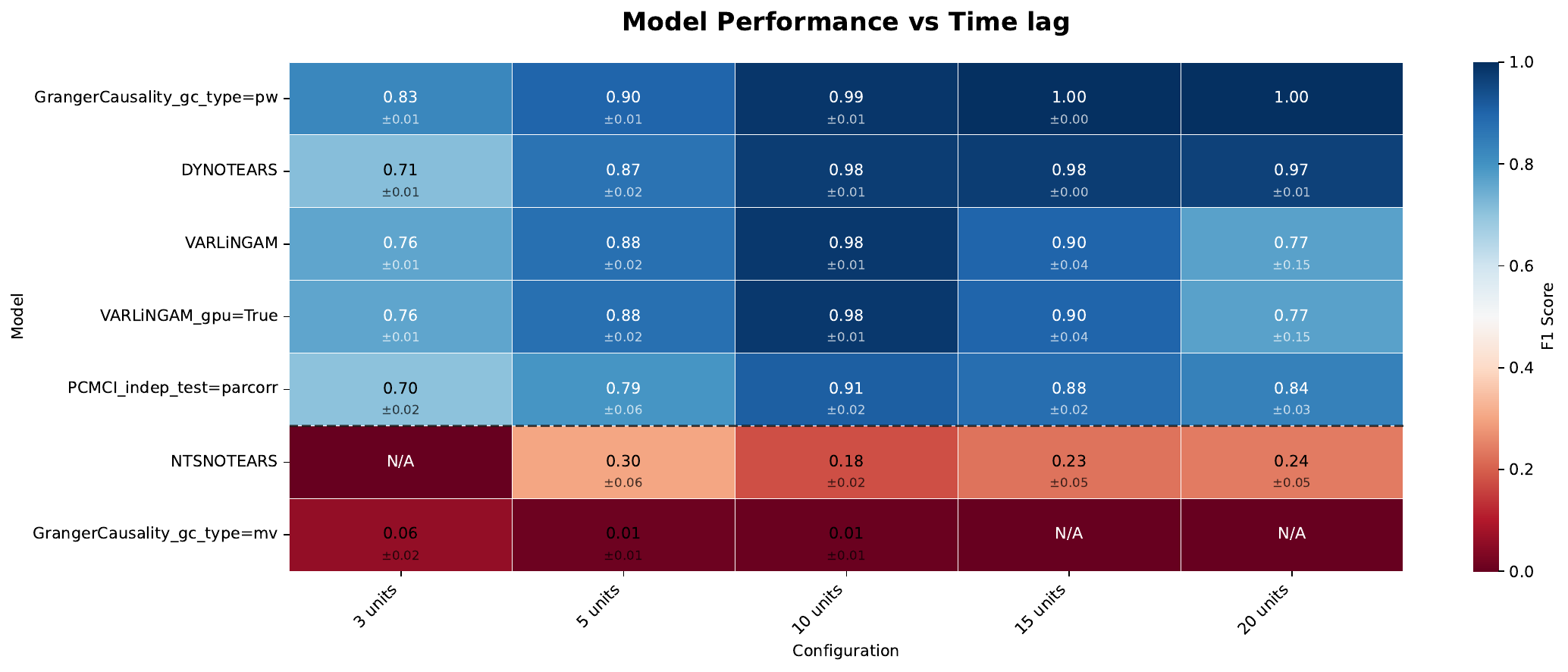}
  \caption{F1 score performance of time-series causal discovery algorithms with different time lags}
  \label{fig:ts_f1_lags}
\end{figure}

\begin{figure}[htbp]
  \centering
  \includegraphics[width=0.72\textwidth]{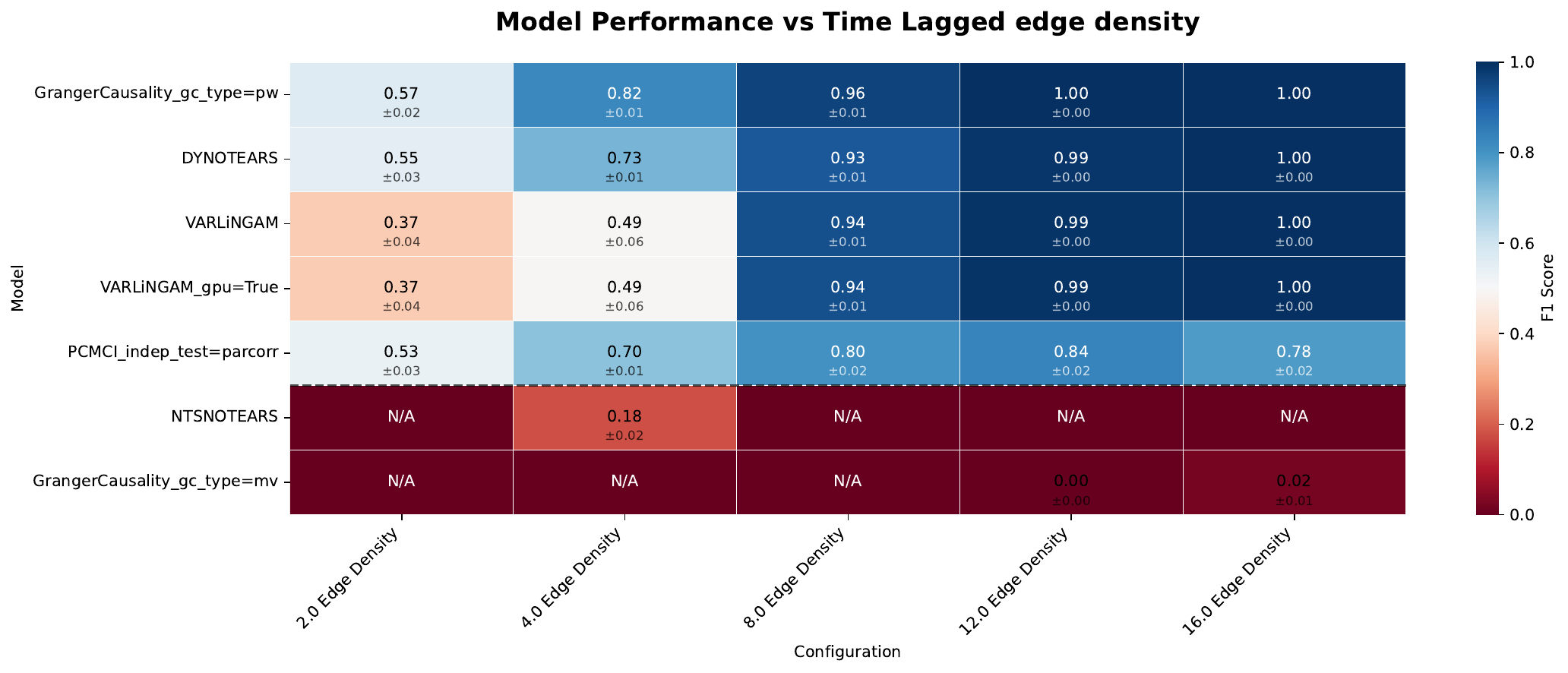}
  \caption{F1 score performance of time-series causal discovery algorithms with different lagged edge densities}
  \label{fig:ts_f1_edges}
\end{figure}

\begin{figure}[htbp]
  \centering
  \includegraphics[width=0.72\textwidth]{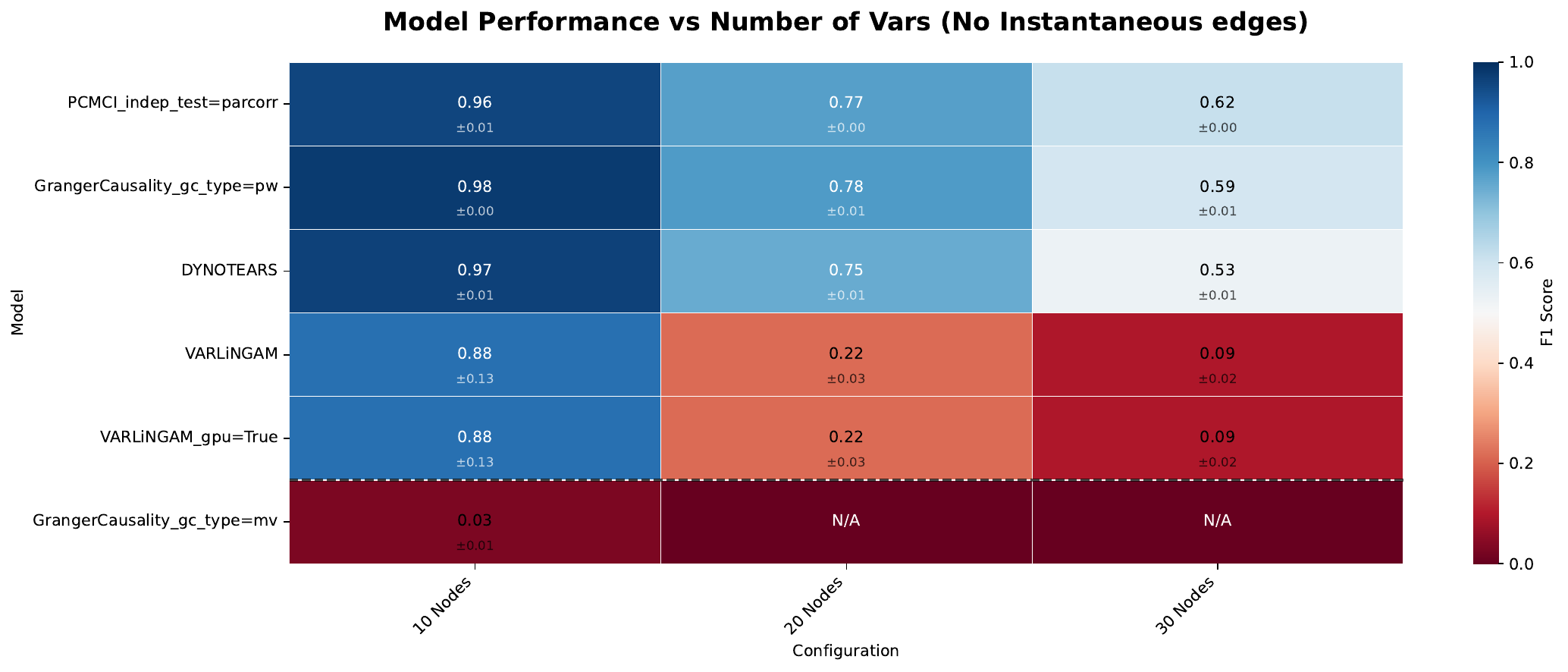}
  \caption{F1 score performance of time-series causal discovery algorithms with purely lagged causal relations, i.e. no instantaneous edges}
  \label{fig:ts_f1_inter_edges}
\end{figure}

\begin{figure}[htbp]
  \centering
  \includegraphics[width=0.72\textwidth]{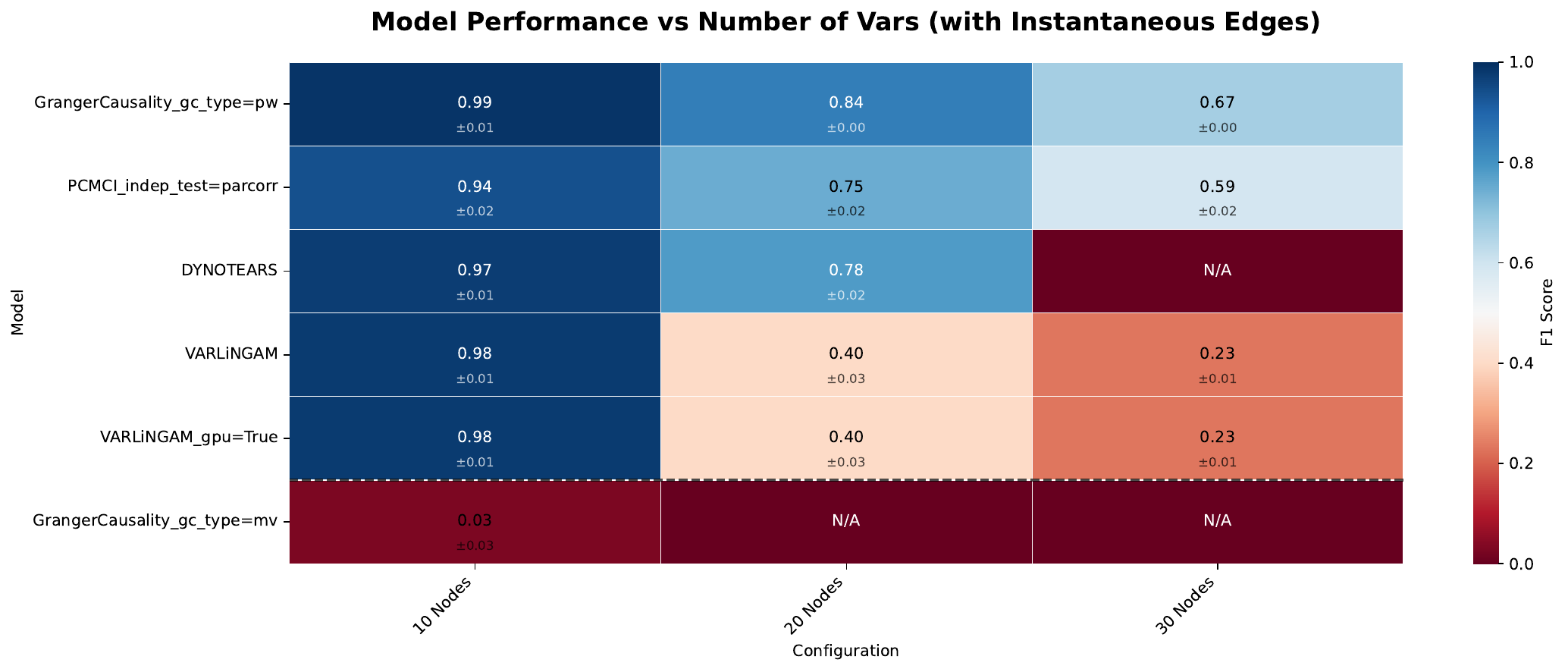}
  \caption{F1 score performance of time-series causal discovery algorithms with both lagged and instantaneous edges}
  \label{fig:ts_f1_intra_edges}
\end{figure}

\begin{figure}[htbp]
  \centering
  \includegraphics[width=0.72\textwidth]{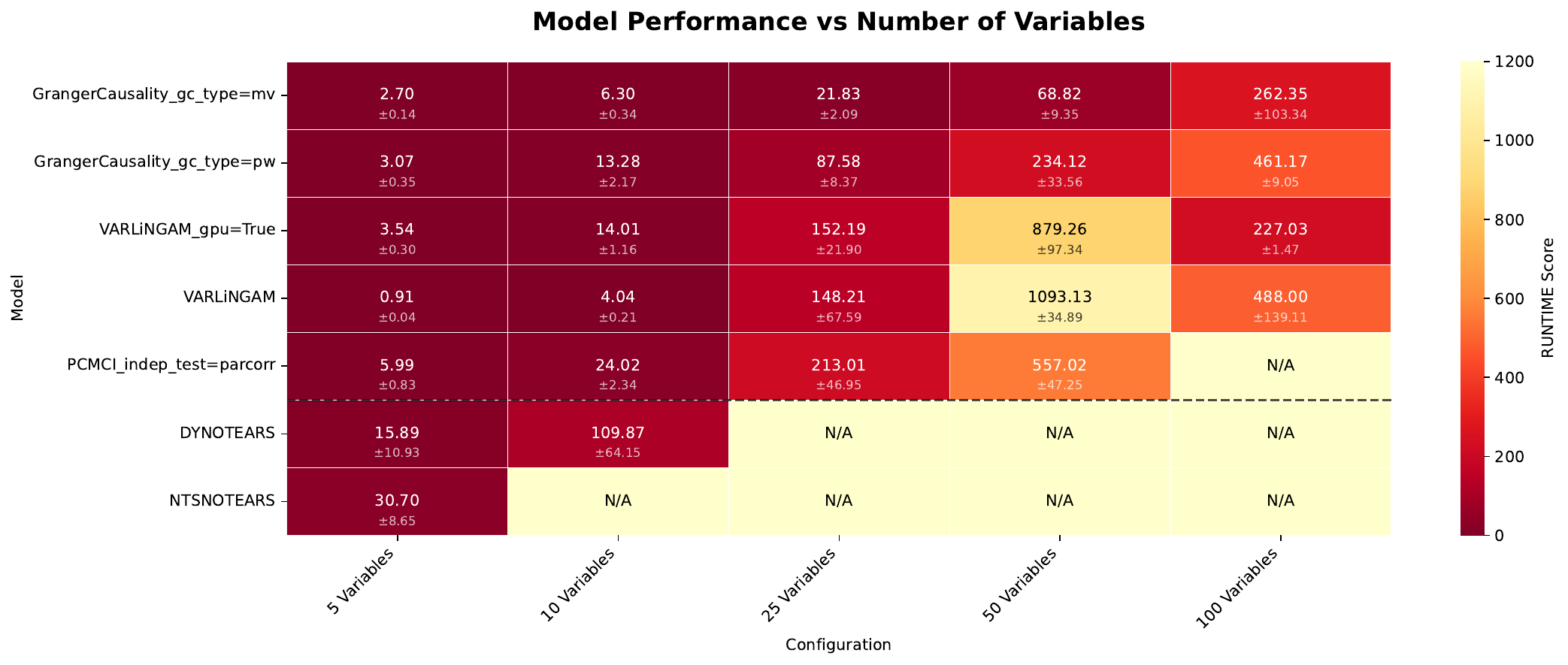}
  \caption{Runtime performance of time-series causal discovery algorithms with varying number of variables}
  \label{fig:ts_rt_vars}
\end{figure}

\begin{figure}[htbp]
  \centering
  \includegraphics[width=0.72\textwidth]{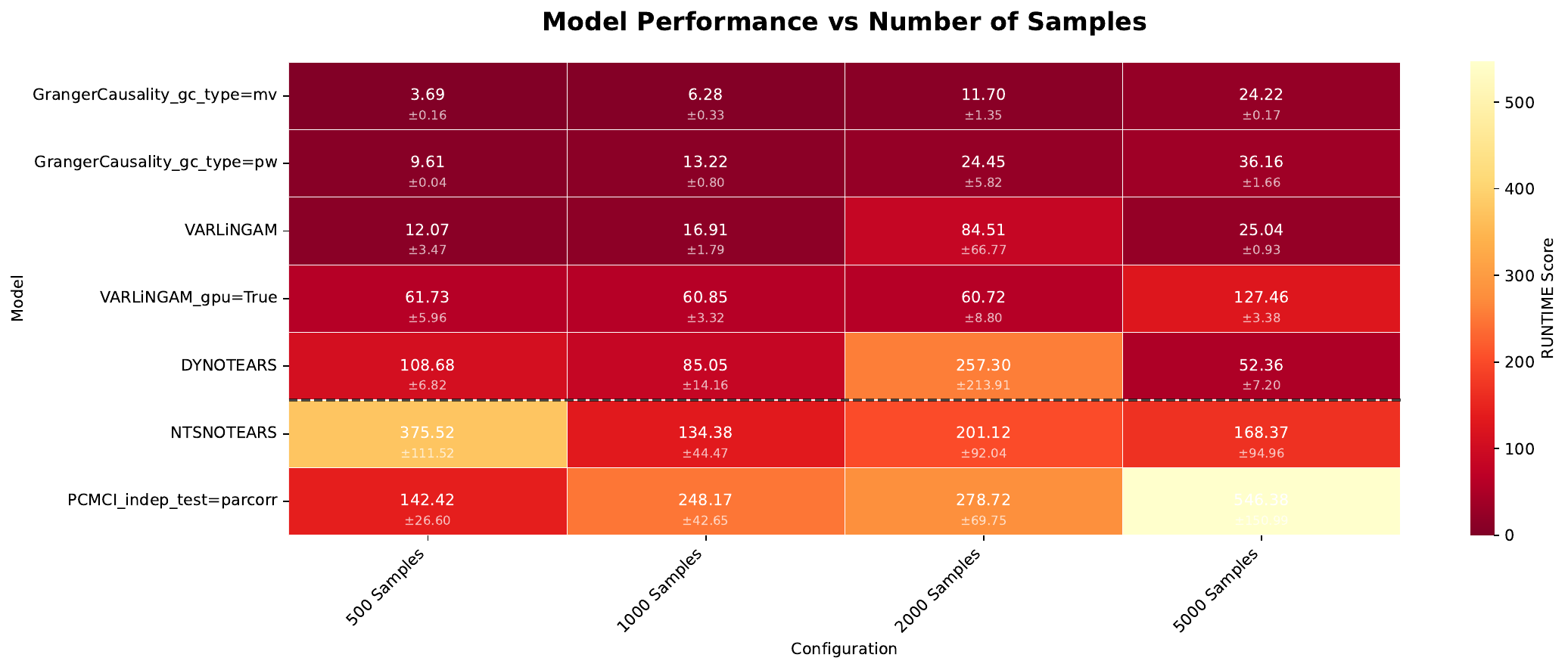}
  \caption{Runtime performance of time-series causal discovery algorithms with varying number of samples}
  \label{fig:ts_rt_samples}
\end{figure}

\begin{figure}[htbp]
  \centering
  \includegraphics[width=0.72\textwidth]{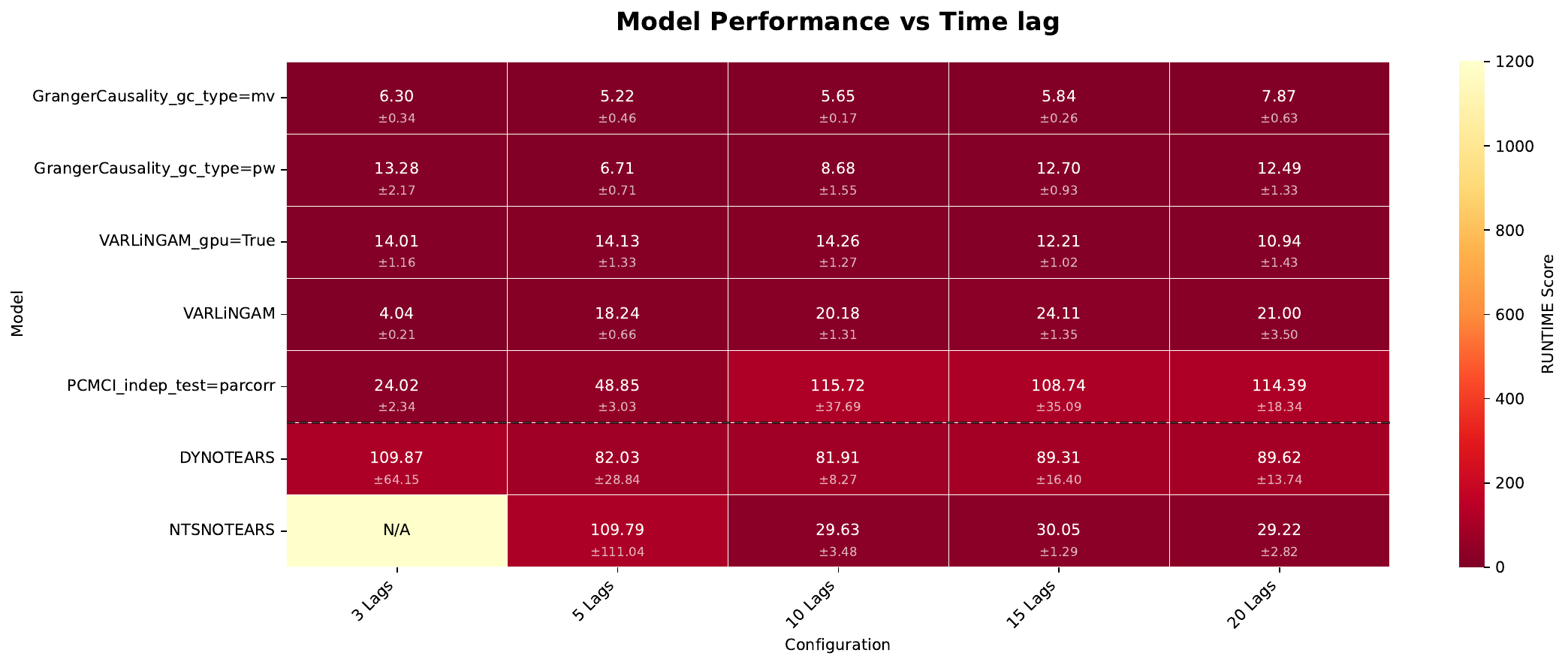}
  \caption{Runtime performance of time-series causal discovery algorithms with varying number of time lags}
  \label{fig:ts_rt_lags}
\end{figure}

\end{document}